\documentclass{article}
\usepackage{arxiv}

\usepackage[utf8]{inputenc} 
\usepackage[T1]{fontenc}    
\usepackage{hyperref}       
\usepackage{url}            
\usepackage{booktabs}       
\usepackage{amsfonts}       
\usepackage{nicefrac}       
\usepackage{microtype}      
\usepackage{graphicx}
\usepackage{enumerate}
\usepackage{animate}
\usepackage{scalerel}
\usepackage{natbib}
\usepackage[title]{appendix}
\usepackage{epstopdf}
\usepackage[flushleft]{threeparttable}
\usepackage{multirow}

\usepackage{amssymb}
\usepackage{booktabs}
\usepackage{mathtools}
\usepackage{makecell}
\usepackage[inline]{enumitem}
\usepackage[amsmath,thmmarks]{ntheorem}
\usepackage{multicol}
\usepackage{tikz} 
\usetikzlibrary{calc,backgrounds}
\usetikzlibrary{shapes.geometric, arrows, arrows.meta}
\usetikzlibrary{positioning}
\usetikzlibrary{decorations.pathreplacing,calligraphy}
\usetikzlibrary{fadings}
\usepackage{bbm}
\usepackage{xcolor}

\usepackage[caption=false,subrefformat=parens,labelformat=parens]{subfig}
\usepackage{caption}
\usepackage{setspace}
\usepackage{algorithmic}
\usepackage{algorithm}
\ifpdf
  \DeclareGraphicsExtensions{.eps,.pdf,.png,.jpg}
\else
  \DeclareGraphicsExtensions{.eps}
\fi

\DeclareMathOperator*{\argmax}{\arg\!\max}

\theoremstyle{plain}

\theoremstyle{nonumberplain}
\theorembodyfont{\normalfont}
\theoremsymbol{\ensuremath{\square}}

\makeatother

\title{Deep Gaussian Process Emulation using Stochastic Imputation}

\author{
  Deyu Ming\thanks{Corresponding author: \texttt{deyu.ming.16@ucl.ac.uk}.} \\
  School of Management\\
  University College London, UK \\
   \And
  Daniel Williamson \\
  College of Engineering, Mathematics and Physical Sciences\\
  University of Exeter, UK \\
   \And
 Serge Guillas \\
  Department of Statistical Science\\
  University College London, UK \\
}

\begin{document}
\maketitle

\begin{abstract}
Deep Gaussian processes (DGPs) provide a rich class of models that can better represent functions with varying regimes or sharp changes, compared to conventional GPs. In this work, we propose a novel inference method for DGPs for computer model emulation. By stochastically imputing the latent layers, our approach transforms a DGP into a linked GP: a novel emulator developed for systems of linked computer models. This transformation permits an efficient DGP training procedure that only involves optimizations of conventional GPs. In addition, predictions from DGP emulators can be made in a fast and analytically tractable manner by naturally utilizing the closed form predictive means and variances of linked GP emulators. We demonstrate the method in a series of synthetic examples and empirical applications, and show that it is a competitive candidate for DGP surrogate inference, combining efficiency that is comparable to doubly stochastic variational inference and uncertainty quantification that is comparable to the fully-Bayesian approach. A \texttt{Python} package \texttt{dgpsi} implementing the method is also produced and available at \url{https://github.com/mingdeyu/DGP}.
\end{abstract}


\section{Introduction}
\label{sec:intro}
Gaussian Processes (GPs) are widely used in Uncertainty Quantification (UQ) applications to emulate computationally expensive computer models for fast model evaluations, reducing computational efforts required for other UQ tasks such as uncertainty propagation, sensitivity analysis, and calibration. The popularity of GP emulators is attributed to their flexibility, native uncertainty incorporation, and analytical tractability for many key properties such as the likelihood function, predictive distribution and associated derivatives. However, many standard and popular kernel functions (e.g., squared exponential and Matérn kernels) that are overwhelmingly used for GP emulation limit the expressiveness of emulators. A number of papers attempt to address this challenge. For example, \citet{paciorek2003nonstationary} introduce a non-stationary kernel to overcome the non-stationary assumption of GPs with standard kernel functions (hereinafter referred to as conventional GPs). Bayesian Treed Gaussian Processes (TGPs), proposed by~\citet{gramacy2008bayesian}, emulate computer models by splitting the input space into several axially-aligned partitions, over which the computer model responses can be better represented by conventional GPs. Other studies such as~\citet{montagna2016computer} and~\citet{volodina2020diagnostics} use augmented kernels and mixtures of conventional kernels respectively to improve the expressiveness of GP emulators. 

Deep Gaussian Processes (DGPs)~\citep{damianou2013deep} model complex Input/Output (I/O) relations, by convolving conventional GPs. Compared to other approaches, DGPs provide a richer class of models with better expressiveness than conventional GPs through a feed-forward hierarchy, mirroring deep neural networks. Although DGPs offer a rich and flexible class of non-stationary models, DGP inference (i.e., training and prediction) has been proven difficult owing to the need to infer the latent layers. Efforts at meeting this challenge from within the machine learning community center around approximate inference. For example, \cite{bui2016deep} use Expectation Propagation (EP) to approximate the analytically intractable objective function so that DGP fitting can be carried out by optimization, e.g., Stochastic Gradient Descent (SGD). Similar to EP, Variational Inference (VI) provides the most popular approach for DGP fitting~\citep{damianou2013deep, wang2016sequential,havasi2018inference}. Doubly Stochastic VI (DSVI)~\citep{salimbeni2017doubly}, which has been shown to outperform EP, is the current state-of-the-art approach and has been utilized by~\citet{radaideh2020surrogate,rajaram2020empirical} for computer model emulation. It is recently also implemented in \texttt{GPflux}~\citep{dutordoir2021gpflux}, an actively maintained open-source library dedicated to DGP. DSVI approximates the exact posterior distribution of the latent variables of a DGP using variational distributions. However, such approximations can be unsatisfactory because the variational distributions can often be poor representations of the true posterior distributions of the latent variables, particularly in the tails. As a result, whilst DSVI offers computational tractability, it can come at the expense of accurate UQ for the latent posteriors, which is essential for computer model emulation.

To address this drawback, \citet{sauer2020active} provide a Fully-Bayesian (FB) inference using elliptical slice sampling~\citep{murray2010elliptical}, that accounts for the various uncertainties in the construction of DGP surrogates. However, computational tractability limits the FB framework implemented in \citet{sauer2020active} to certain minimal DGP specifications (e.g., no more than three-layered DGPs). Additionally, the fully sampling-based inference employed by \citet{sauer2020active} is computationally expensive and thus may not be well-suited to some UQ tasks, such as calibration or sensitivity analysis, that involve computer model emulation.

In this work, we introduce a novel inference, called Stochastic Imputation (SI) that balances the speed embraced by the optimization-based DSVI and accuracy enjoyed by the MCMC-based FB method. It is algorithmically effective and straightforward for DGP surrogate modeling with different hierarchical structures. Unlike other studies that treat DGPs simply as compositions of GPs, we see DGPs through the lenses of linked GPs~\citep{kyzyurova2018coupling,ming2021linked} that enjoy a simple and fast inference procedure. By exploiting the idea that a linked GP can be viewed as a DGP with its hidden layers exposed, our approach is to convert DGPs to linked GPs by stochastically imputing the hidden layers of DGPs. As a result, the training of a DGP becomes equivalent to several simple conventional GP optimization problems, and DGP predictions can be made analytically by naturally utilizing the closed form predictive mean and variance of linked GP under various kernel functions. It is worth noting that EP also implements DGP predictions in an analytical manner. However, linked GP provides closed form DGP predictions with a wider range of kernel choices and more general hierarchies, allowing more flexible DGP specifications and structural engineering (e.g., the input-connected structure that we demonstrate in Section~\ref{sec:heston}) for computer model emulation.

Our aim is to present a novel inference approach to DGP emulation of computer models and to compare it in terms of speed and adequacy of UQ to the variational and FB approaches. Performance of DGPs in general in comparison to other non-stationary GP methods has been made elsewhere and is beyond the scope of this work. The paper is organized as follows. In Section~\ref{sec:review},
we review conventional GPs, linked GPs, and DGPs. Our approach for DGP inference is then presented in Section~\ref{sec:inference}, in which we detail the prediction, imputation, and training procedures for DGP. We then compare our approach to DSVI and FB, via a synthetic experiment in Section~\ref{sec:step_fct}, and a real-world example on financial engineering in Section~\ref{sec:heston}. An additional 5-dimensional synthetic problem and an extra real-world application on surrogate modeling of aircraft engine simulator are presented in Section~\ref{sec:5d-case} and~\ref{sec:aircraft} of the supplement.

\section{Review}
\label{sec:review}
\subsection{Gaussian processes}
\label{sec:gp}
Let $\mathbf{X}\in\mathbb{R}^{M\times D}$ represent $M$ sets of $D$-dimensional input to a computer model and $\mathbf{Y}(\mathbf{X})\in\mathbb{R}^{M\times 1}$ be the corresponding $M$ scalar-valued outputs. Then, the GP model assumes that $\mathbf{Y}(\mathbf{X})$ follows a multivariate normal distribution
$
\mathbf{Y}(\mathbf{X})\sim\mathcal{N}(\boldsymbol{\mu}(\mathbf{X}),\,\boldsymbol{\Sigma}(\mathbf{X})),
$
where $\boldsymbol{\mu}(\mathbf{X})\in\mathbb{R}^{M\times 1}$ is the mean vector whose $i$-th element is often specified as a function of $\mathbf{X}_{i*}$, the $i$-th row of $\mathbf{X}$; $\boldsymbol{\Sigma}(\mathbf{X})=\sigma^2\mathbf{R}(\mathbf{X})\in\mathbb{R}^{M\times M}$ is the covariance matrix with $\mathbf{R}(\mathbf{X})$ being the correlation matrix. The $ij$-th element of $\mathbf{R}(\mathbf{X})$ is specified by $
k(\mathbf{X}_{i*},\,\mathbf{X}_{j*})+\eta\mathbbm{1}_{\{\mathbf{X}_{i*}=\mathbf{X}_{j*}\}}$, where $k(\cdot,\cdot)$ is a given kernel function with $\eta$ being the nugget term and $\mathbbm{1}_{\{\cdot\}}$ being the indicator function. In this study we consider Gaussian processes with zero means, i.e., $\boldsymbol{\mu}(\mathbf{X})=\mathbf{0}$ and kernel functions with the multiplicative form:
$
k(\mathbf{X}_{i*},\,\mathbf{X}_{j*})=\prod_{d=1}^D k_d(X_{id},\,X_{jd}),
$
where $k_d(X_{id},\,X_{jd})=k_d(|X_{id}-X_{jd}|)$ is a one-dimensional isotropic kernel function (e.g., squared exponential and Matérn kernels) with range parameter $\gamma_d$, for the $d$-th input dimension.

Assume that the GP parameters $\sigma^2$, $\eta$ and $\boldsymbol{\gamma}=(\gamma_1,\dots,\gamma_D)$ are known. Then, given the realizations of input $\mathbf{x}=(\mathbf{x}^\top_{1*},\dots,\mathbf{x}^\top_{M*})^\top$ and output $\mathbf{y}=(y_1,\dots,y_M)^\top$, the posterior predictive distribution of output $Y_0(\mathbf{x}_0)$ at a new input position $\mathbf{x}_0\in\mathbb{R}^{1\times D}$ follows a normal distribution with mean $\mu_0(\mathbf{x}_0)$ and variance $\sigma^2_0(\mathbf{x}_0)$ given by:
\begin{equation}
\label{eq:postpred}
\mu_0(\mathbf{x}_0)=\mathbf{r}(\mathbf{x}_0)^\top\mathbf{R}(\mathbf{x})^{-1}\mathbf{y}\quad\mathrm{and}\quad
\sigma^2_0(\mathbf{x}_0)=\sigma^2\left(1+\eta-\mathbf{r}(\mathbf{x}_0)^\top\mathbf{R}(\mathbf{x})^{-1}\mathbf{r}(\mathbf{x}_0)\right),
\end{equation}
where $\mathbf{r}(\mathbf{x}_0)=[k(\mathbf{x}_0,\mathbf{x}_{1*}),\dots,k(\mathbf{x}_0,\mathbf{x}_{M*})]^\top$. The parameters $\sigma^2$, $\eta$ and $\boldsymbol{\gamma}$ are typically estimated e.g., using maximum likelihood or maximum a posteriori~\citep{rasmussen2005gaussian}, though some studies use sampling to propagate their uncertainty. In the remainder of the study, we let $\boldsymbol{\theta}=\{\sigma^2,\eta,\boldsymbol{\gamma}\}$ be the set of GP model parameters and $\widehat{\boldsymbol{\theta}}=\{\widehat{\sigma^2},\widehat{\eta},\widehat{\boldsymbol{\gamma}}\}$ be the corresponding set of estimated model parameters. 

\subsection{Linked Gaussian processes}
\label{sec:linkgp}
\begin{figure}[!ht]
\centering
\scalebox{0.8}{
\begin{tikzpicture}[shorten >=1pt,->,draw=black!50, node distance=4cm]
    \tikzstyle{every pin edge}=[<-,shorten <=1pt]
    \tikzstyle{neuron}=[circle,fill=black!25,minimum size=30.5pt,inner sep=0pt]
    \tikzstyle{layer1}=[neuron, fill=green!50];
    \tikzstyle{layer2}=[neuron, fill=red!50];
    \tikzstyle{layer3}=[neuron, fill=blue!50];
    \tikzstyle{annot} = [text width=4em, text centered]

    \node[layer1, pin=left:$\mathbf{X}$] (I-0) at (0,0) {$\mathcal{GP}^{(1)}_{1}$};
    \node[layer1, pin=left:$\mathbf{X}$] (I-1) at (0,-1.25) {$\mathcal{GP}^{(2)}_{1}$};
    \node[layer1, pin=left:$\mathbf{X}$] (I-2) at (0,-2.95) {$\mathcal{GP}^{(P)}_{1}$};
            \node[layer2, pin={[pin edge={->}]right:$\mathbf{Y}$}] (H-1) at (4cm,-1.25 cm) {$\mathcal{GP}_2$};
            \path [draw] (I-0) -- (H-1) node[font=\small,pos=0.35,fill=white,align=left,sloped] {$\mathbf{W}_{*1}$};
            \path [draw] (I-1) -- (H-1) node[font=\small,pos=0.35,fill=white,align=left,sloped] {$\mathbf{W}_{*2}$};
            \path [draw] (I-2) -- (H-1) node[font=\small,pos=0.35,fill=white,align=left,sloped] {$\mathbf{W}_{*P}$};
            \path (I-1) -- (I-2) node [black, midway, sloped] {$\dots$};
            \path (I-1) -- (I-2) node [black, midway, sloped,transform canvas={xshift=-12.5mm}] {$\dots$};
            \path (I-1) -- (I-2) node [black, midway, sloped,transform canvas={xshift=15mm,yshift=3mm}] {$\dots$};
\end{tikzpicture}}
\caption{The hierarchy of GPs that represents a feed-forward system of two computer models.}
\label{fig:linkgp}
\end{figure}
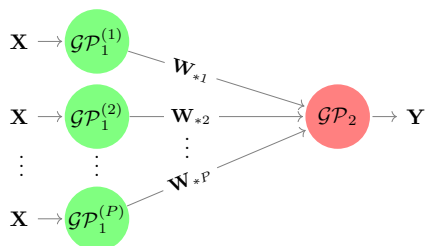

Linked GPs emulate systems of computer models, where each computer model has its own individual GP emulator. Consider a system of two computer models run with $M$ design points, where the first model has $M$ sets of $D$-dimensional input ($\mathbf{X}\in\mathbb{R}^{M\times D}$) and produces $M$ sets of $P$-dimensional output ($\mathbf{W}\in\mathbb{R}^{M\times P}$) that feeds into the second computer model that produces $M$ one-dimensional outputs ($\mathbf{Y}\in\mathbb{R}^{M\times 1}$).  
Let the GP surrogates of the two computer models be $\mathcal{GP}_1$ and $\mathcal{GP}_2$ respectively. Assume that the output $\mathbf{W}$ of the first computer model is conditionally independent across dimensions, i.e., the column vectors $\mathbf{W}_{*p}$ of $\mathbf{W}$ are independent conditional on $\mathbf{X}$. Then, $\mathcal{GP}_1$ is a collection of independent GPs, $\{\mathcal{GP}^{(p)}_{1}\}_{p=1,\dots,P}$. Over the design,  each GP corresponds to a multivariate normal distribution as in Section~\ref{sec:gp} with input $\mathbf{X}$ and output $\mathbf{W}_{*p}$. The hierarchy of GPs that represents the system is shown in Figure~\ref{fig:linkgp}.

Assume that, given inputs $\mathbf{X}=\mathbf{x}$ we observe realisations $\mathbf{w}$ and $\mathbf{y}$ of $\mathbf{W}$ and $\mathbf{Y}$, and that the model parameters involved in $\mathcal{GP}_1$ and $\mathcal{GP}_2$ are known or estimated. Then, the posterior predictive distribution of the global output $Y_0(\mathbf{x}_0)$ at a new global input position $\mathbf{x}_0$ is given by
$Y_0(\mathbf{x}_0)|\mathcal{D}\sim p(y_0|\mathbf{x}_0; \mathbf{y},\mathbf{w},\mathbf{x})$, where $\mathcal{D}=\{\mathbf{Y}=\mathbf{y},\mathbf{W}=\mathbf{w},\mathbf{X}=\mathbf{x}\}$ and $p(y_0|\mathbf{y},\mathbf{w},\mathbf{x})$ is the pdf of $Y_0(\mathbf{x}_0)|\mathcal{D}$. Note that
\begin{align}
\label{eq:linkdensity}
p(y_0|\mathbf{x}_0;\mathbf{y},\mathbf{w},\mathbf{x})=&\int p(y_0|\mathbf{w}_0;\mathbf{y},\mathbf{w},\mathbf{x})p(\mathbf{w}_0|\mathbf{x}_0;\mathbf{y},\mathbf{w},\mathbf{x})\mathrm{d}\mathbf{w}_0\nonumber\\
=&\int p(y_0|\mathbf{w}_0;\mathbf{y},\mathbf{w})\prod_{p=1}^{P}p(w_{0p}|\mathbf{x}_0;\mathbf{w}_{*p},\mathbf{x})\mathrm{d}\mathbf{w}_0,
\end{align}
where $p(y_0|\mathbf{w}_0;\mathbf{y},\mathbf{w})$ and $p(w_{0p}|\mathbf{x}_0;\mathbf{w}_{*p},\mathbf{x})$ are pdf's of the posterior predictive distributions of $\mathcal{GP}_2$ and $\mathcal{GP}^{(p)}_{1}$ respectively; and $\mathbf{w}_0=(w_{01},\dots,w_{0P})$. However, $p(y_0|\mathbf{x}_0;\mathbf{y},\mathbf{w},\mathbf{x})$ is not analytically tractable because the integral in equation~\eqref{eq:linkdensity} does not permit a closed form expression. It can been shown~\citep{titsias2010bayesian,kyzyurova2018coupling,ming2021linked} that, given the GP specifications in Section~\ref{sec:gp}, the mean, $\tilde{\mu}_0(\mathbf{x}_0)$, and variance, $\tilde{\sigma}_0^2(\mathbf{x}_0)$, of $Y_0(\mathbf{x}_0)|\mathcal{D}$ have the following analytical expressions:
\begin{align}
\label{eq:linkgp_mean}
\tilde{\mu}_0(\mathbf{x}_0)=&\mathbf{I(\mathbf{x}_0)}^\top\mathbf{R}(\mathbf{w})^{-1}\mathbf{y},\\
\label{eq:linkgp_var}
\tilde{\sigma}_0^2(\mathbf{x}_0)=&\mathbf{y}^\top\mathbf{R}(\mathbf{w})^{-1}\mathbf{J}(\mathbf{x}_0)\mathbf{R}(\mathbf{w})^{-1}\mathbf{y}-\left(\mathbf{I}(\mathbf{x}_0)^\top\mathbf{R}(\mathbf{w})^{-1}\mathbf{y}\right)^2+\sigma^2\,\left(1+\eta-\mathrm{tr}\left\{\mathbf{R}(\mathbf{w})^{-1}\mathbf{J}(\mathbf{x}_0)\right\}\right),
\end{align}
where 
\begin{itemize}
    \item $\mathbf{I(\mathbf{x}_0)}\in\mathbb{R}^{M\times 1}$ with its $i$-th element $I_i=\prod_{p=1}^P\mathbb{E}\left[k_p(W_{0p}(\mathbf{x}_0),\,w_{ip})\right]$;
    \item $\mathbf{J}(\mathbf{x}_0)\in\mathbb{R}^{M\times M}$ with its $ij$-th element $J_{ij}=\prod_{p=1}^P\mathbb{E}\left[k_p(W_{0p}(\mathbf{x}_0),\,w_{ip})\,k_p(W_{0p}(\mathbf{x}_0),\,w_{jp})\right]$;
\end{itemize}
and the expectations in $\mathbf{I(\mathbf{x}_0)}$ and $\mathbf{J}(\mathbf{x}_0)$ have closed form expressions under the linear kernel, squared exponential kernel, and a class of Matérn kernels~\citep[Proposition 3.4]{ming2021linked}. The linked GP is then defined as a normal approximation $\widehat{p}(y_0|\mathbf{x}_0; \mathbf{y},\mathbf{w},\mathbf{x})$ to $p(y_0|\mathbf{x}_0; \mathbf{y},\mathbf{w},\mathbf{x})$ with its mean and variance given by $\tilde{\mu}_0(\mathbf{x}_0)$ and $\tilde{\sigma}_0^2(\mathbf{x}_0)$.
Moreover, the linked GP can be constructed iteratively to approximate analytically the posterior predictive distribution of global output produced by any feed-forward systems of GPs, and is shown to be a sufficient approximation in terms of minimizing Kullback–Leibler divergence~\citep{ming2021linked}.

\subsection{Deep Gaussian processes} 
The DGP model is a feed-forward composition of conventional GPs and only differs from the linked GP model in that the internal inputs/outputs of GPs are latent. For example, the model hierarchy in Figure~\ref{fig:linkgp} represents a two-layered DGP when the variable $\mathbf{W}$ is latent. The existence of latent variables creates challenges to conduct efficient inference for DGP models. For instance, to train the two-layer DGP in Figure~\ref{fig:linkgp} by the maximum likelihood approach, one needs to optimize the model parameters by maximizing the likelihood function:
\begin{equation}
\label{eq:dgpdensity}
\mathcal{L}=p(\mathbf{y}|\mathbf{x})=\int p(\mathbf{y}|\mathbf{w})\prod_{p=1}^P p(\mathbf{w}_{*p}|\mathbf{x})\mathrm{d}\mathbf{w},
\end{equation}
where $p(\mathbf{y}|\mathbf{w})$ is the multivariate normal pdf of $\mathcal{GP}_2$ and $p(\mathbf{w}_{*p}|\mathbf{x})$ is the multivariate normal pdf of $\mathcal{GP}^{(p)}_{1}$. However, equation~\eqref{eq:dgpdensity} contains an integral with respect to the latent variable $\mathbf{w}$ that is not analytically tractable due to the non-linearity between $\mathbf{y}$ and $\mathbf{w}$, and the number of such intractable integrals increases along with the depth of a DGP.

The inference challenge induced by the latent layers is tackled in the literature by DSVI that uses the variational distribution, a composition of independent (across layers) Gaussian distributions, and thus an Evidence Lower BOund (ELBO) that can be efficiently maximized. In addition to the common concern that the variational approximation may not capture important features of the posterior uncertainty, the maximization of the ELBO can be computationally challenging due to the complexity (e.g., non-convexity and the large amount of model parameters) induced by a network of GPs. Alternatively, \cite{sauer2020active} present a sampling-based FB inference for DGP using MCMC. The FB approach properly quantifies the uncertainties in DGP inference, but does so at the expense of computational efficiency. The approach we describe in the next section aims to blend computational efficiency of VI and accuracy of FB for DGP emulation by combining the linked GP with a sampling approach.

\section{Stochastic Imputation for DGP Inference}
\label{sec:inference}
We view the DGP as an emulator of a feed-forward system of computer models in which, each sub-model is represented by a GP and internal I/O among sub-models are non-observable. Thus, by imputing the hidden layers and exploiting the structural dependence of the internal GP surrogates, we uncover, stochastically, the latent internal I/O from the observed global I/O. As a result, we proceed to make predictions from the DGP using the analytically tractable linked GP. 

\subsection{Model}
\begin{figure}[!ht]
\centering
\scalebox{0.7}{
\begin{tikzpicture}[shorten >=1pt,->,draw=black!50, node distance=4cm]
    \tikzstyle{every pin edge}=[<-,shorten <=1pt]
    \tikzstyle{neuron}=[circle,fill=black!25,minimum size=35pt,inner sep=0pt]
    \tikzstyle{layer1}=[neuron, fill=green!50];
    \tikzstyle{layer2}=[neuron, fill=red!50];
    \tikzstyle{layer3}=[neuron, fill=blue!50];
    \tikzstyle{innerlayer}=[neuron, fill=white];
    \tikzstyle{annot} = [text width=4em, text centered]

    \node[layer1, pin=left:$\mathbf{X}$] (l1-0) at (0,0.2) {$\mathcal{GP}^{(1)}_{1}$};
    \node[layer1, pin=left:$\mathbf{X}$] (l1-1) at (0,-1.25) {$\mathcal{GP}^{(2)}_{1}$};
    \node[layer1, pin=left:$\mathbf{X}$] (l1-2) at (0,-3.15) {$\mathcal{GP}^{(P_1)}_{1}$};
    \node[layer2] (l2-0) at (2.5,0.2) {$\mathcal{GP}^{(1)}_{2}$};
    \node[layer2] (l2-1) at (2.5,-1.25) {$\mathcal{GP}^{(2)}_{2}$};
    \node[layer2] (l2-2) at (2.5,-3.15) {$\mathcal{GP}^{(P_2)}_{2}$};
    \node[innerlayer] (I-0) at (5,0) {\ldots};
    \node[innerlayer] (I-1) at (5,-1.25) {\ldots};
    \node[innerlayer] (I-2) at (5,-2.95) {\ldots};
    \node[layer3,pin={[pin edge={->}]right:$\mathbf{Y}^{(1)}$}] (ln-0) at (7.5,0.2) {$\mathcal{GP}^{(1)}_{L}$};
    \node[layer3, pin={[pin edge={->}]right:$\mathbf{Y}^{(2)}$}] (ln-1) at (7.5,-1.25) {$\mathcal{GP}^{(2)}_{L}$};
    \node[layer3, pin={[pin edge={->}]right:$\mathbf{Y}^{(P_L)}$}] (ln-2) at (7.5,-3.15) {$\mathcal{GP}^{(P_L)}_{L}$};
\path (l1-1) -- (l1-2) node [black, midway, sloped] {$\dots$};
\path (l2-1) -- (l2-2) node [black, midway, sloped] {$\dots$};
\path (ln-1) -- (ln-2) node [black, midway, sloped] {$\dots$};
\path (I-1) -- (I-2) node [black, midway, sloped] {$\dots$};
\path (l1-1) -- (l1-2) node [black, midway, sloped,transform canvas={xshift=-15mm}] {$\dots$};
\path (ln-1) -- (ln-2) node [black, midway, sloped,transform canvas={xshift=15mm}] {$\dots$};
\draw[->] (l1-0) -- (l2-0);
\draw[->] (l1-0) -- (l2-1);
\draw[->] (l1-0) -- (l2-2);
\draw[->] (l1-1) -- (l2-0);
\draw[->] (l1-1) -- (l2-1);
\draw[->] (l1-1) -- (l2-2);
\draw[->] (l1-2) -- (l2-0);
\draw[->] (l1-2) -- (l2-1);
\draw[->] (l1-2) -- (l2-2);
\draw[->,path fading=east] (l2-0) -- (I-0);
\draw[->,path fading=east] (l2-0) -- (I-1);
\draw[->,path fading=east] (l2-0) -- (I-2);
\draw[->,path fading=east] (l2-1) -- (I-0);
\draw[->,path fading=east] (l2-1) -- (I-1);
\draw[->,path fading=east] (l2-1) -- (I-2);
\draw[->,path fading=east] (l2-2) -- (I-0);
\draw[->,path fading=east] (l2-2) -- (I-1);
\draw[->,path fading=east] (l2-2) -- (I-2);
\draw[->,path fading=west] (I-0) -- (ln-0);
\draw[->,path fading=west] (I-0) -- (ln-1);
\draw[->,path fading=west] (I-0) -- (ln-2);
\draw[->,path fading=west] (I-1) -- (ln-0);
\draw[->,path fading=west] (I-1) -- (ln-1);
\draw[->,path fading=west] (I-1) -- (ln-2);
\draw[->,path fading=west] (I-2) -- (ln-0);
\draw[->,path fading=west] (I-2) -- (ln-1);
\draw[->,path fading=west] (I-2) -- (ln-2);
\end{tikzpicture}}
\caption{The generic DGP hierarchy considered to illustrate the Stochastic Imputation (SI).}
\label{fig:dgpmodel}
\end{figure}
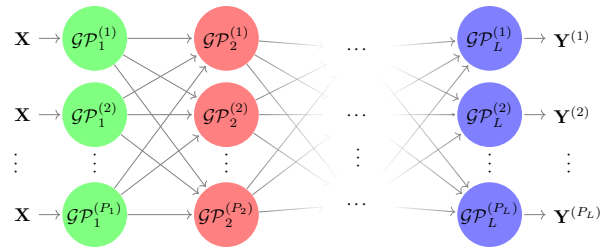

We illustrate our approach by considering the generic $L$-layered DGP hierarchy shown in Figure~\ref{fig:dgpmodel}, where $\mathbf{X}\in\mathbb{R}^{M\times D}$ is the global input and $\{\mathbf{Y}^{(p)}\}_{p=1,\dots,P_L}\in\mathbb{R}^{M\times 1}$ are $P_L$ global outputs. Let $\mathbf{W}^{(p)}_l\in\mathbb{R}^{M\times 1}$ be the output of $\mathcal{GP}^{(p)}_{l}$ for $p=1,\dots,P_l$ and $l=1,\dots,L-1$ and assume that the outputs $\{\mathbf{W}^{(p)}_l\}_{p=1,\dots,P_l}$ from GPs from the $l$-th layer are conditionally independent given the corresponding inputs that are produced by the feeding GPs from the $(l-1)$-th layer. In the rest of the work, we use $\{\mathbf{W}^{(p)}_l\}$ as the shorthand of $\{\mathbf{W}^{(1)}_1,\dots,\mathbf{W}^{(P_1)}_1,\dots,\mathbf{W}^{(1)}_{L-1},\dots,\mathbf{W}^{(P_{L-1})}_{L-1}\}$, and $\{\boldsymbol{\theta}^{(p)}_l\}$ as the set of model parameters of all GPs in the DGP architecture. 

\subsection{Prediction}
\label{sec:prediction}
Assume that the model parameters $\boldsymbol{\theta}^{(p)}_l$ of $\mathcal{GP}^{(p)}_l$ are known and distinct for all $p=1,\dots,P_l$ and $l=1,\dots,L$, and that we have an observation $\mathbf{x}$ and $\mathbf{y}=(\mathbf{y}^{(1)},\dots,\mathbf{y}^{(P_L)})$ of the global input $\mathbf{X}$ and output $\mathbf{Y}=(\mathbf{Y}^{(1)},\dots,\mathbf{Y}^{(P_L)})$. To obtain the posterior predictive distribution of the $p$-th output $Y^{(p)}_0(\mathbf{x}_0)$ at a new input position $\mathbf{x}_0$, the stochastic imputation procedure fills in the latent variables $\{\mathbf{W}^{(p)}_l\}$ by a random realization $\{\mathbf{w}^{(p)}_l\}$ drawn from $p(\{\mathbf{w}^{(p)}_l\}|\mathbf{y},\mathbf{x})$, the posterior distribution of latent variables. We defer the discussion on how to draw realizations from $p(\{\mathbf{w}^{(p)}_l\}|\mathbf{y},\mathbf{x})$ to Section~\ref{sec:latent}. After obtaining $\{\mathbf{w}^{(p)}_l\}$, the posterior predictive distribution $p(y^{(p)}_0|\mathbf{x}_0;\mathbf{y},\mathbf{x})$ of $Y^{(p)}_0(\mathbf{x}_0)$ for all $p=1,\dots,P_L$ can then be approximated by a linked GP with closed form mean and variance. However, a single imputation would neglect the uncertainties of the hidden layers, i.e., the imputation uncertainty is not appropriately assessed. Therefore, one can draw $N$ realizations $\{\mathbf{w}^{(p)}_{l}\}_1,\dots,\{\mathbf{w}^{(p)}_{l}\}_N$ of $\{\mathbf{W}^{(p)}_{l}\}$ from $p(\{\mathbf{w}^{(p)}_l\}|\mathbf{y},\mathbf{x})$, and construct $N$ linked GPs accordingly. Finally, the information contained in these $N$ linked GPs can be combined to describe the posterior predictive distribution of $Y^{(p)}_0(\mathbf{x}_0)$ that properly reflects the uncertainty because of the latent variables.

Note that $p(y^{(p)}_0|\mathbf{x}_0;\mathbf{y},\mathbf{x})$ can be approximated by a mixture of $N$ constructed linked GPs:
\begin{align*}
\label{eq:mixture}
p(y^{(p)}_0|\mathbf{x}_0;\mathbf{y},\mathbf{x})=&\int p(y^{(p)}_0|\mathbf{x}_0;\mathbf{y},\{\mathbf{w}^{(p)}_{l}\},\mathbf{x})\,p(\{\mathbf{w}^{(p)}_{l}\}|\mathbf{y},\mathbf{x})\,\mathrm{d}\{\mathbf{w}^{(p)}_{l}\}\nonumber\\
=&\mathbb{E}_{\{\mathbf{W}^{(p)}_{l}\}|\mathbf{y},\mathbf{x}}\left[p(y^{(p)}_0|\mathbf{x}_0;\mathbf{y},\{\mathbf{W}^{(p)}_{l}\},\mathbf{x})\right]\nonumber\\
\approx&\frac{1}{N}\sum_{i=1}^N p(y^{(p)}_0|\mathbf{x}_0;\mathbf{y},\{\mathbf{w}^{(p)}_{l}\}_i,\mathbf{x})\\
\approx&\frac{1}{N}\sum_{i=1}^N \widehat{p}(y^{(p)}_0|\mathbf{x}_0;\mathbf{y},\{\mathbf{w}^{(p)}_{l}\}_i,\mathbf{x})
\end{align*}
in which $\widehat{p}(y^{(p)}_0|\mathbf{x}_0;\mathbf{y},\{\mathbf{w}^{(p)}_{l}\},\mathbf{x})$ denotes the pdf of the linked GP. Thus, the approximate posterior predictive mean and variance of $Y^{(p)}_0(\mathbf{x}_0)$ can be obtained by:
\begin{equation}
    \label{eq:mean_var}
    \tilde{\mu}^{(p)}_0=\frac{1}{N}\sum_{i=1}^N\tilde{\mu}^{(p)}_{0,i} \quad\mathrm{and}\quad
    (\tilde{\sigma}^{(p)}_0)^2=\frac{1}{N}\sum_{i=1}^N((\tilde{\mu}^{(p)}_{0,i})^2+(\tilde{\sigma}^{(p)}_{0,i})^2)-(\tilde{\mu}^{(p)}_0)^2,
\end{equation}
where $\{\tilde{\mu}^{(p)}_{0,i},(\tilde{\sigma}^{(p)}_{0,i})^2\}_{i=1,\dots,N}$ are closed form means and variances, the expressions of which are given in~\eqref{eq:linkgp_mean} and~\eqref{eq:linkgp_var}, of the $N$ constructed linked GPs. The DGP prediction procedure in SI is given in Algorithm~\ref{alg:prediction}.

\begin{algorithm}[htbp]
\caption{Prediction from the DGP model in Figure~\ref{fig:dgpmodel} using SI}
\label{alg:prediction}
\begin{algorithmic}[1]
\REQUIRE{\begin{enumerate*}[label=(\roman*)]
  \item Observations $\mathbf{x}$ and $\mathbf{y}$;
  \item $\{\mathcal{GP}^{(p)}_l\}$;
  \item a new input location $\mathbf{x}_0$.
\end{enumerate*}}
\ENSURE{Mean and variance of $Y^{(p)}_0(\mathbf{x}_0)$, for $p=1,\dots,P_L$.}
\label{alg:mi}
\STATE{Impute latent variables $\{\mathbf{W}^{(p)}_l\}$ by $N$ realisations $\{\mathbf{w}^{(p)}_l\}_1,\dots,\{\mathbf{w}^{(p)}_l\}_N$ drawn from $p(\{\mathbf{w}^{(p)}_l\}|\mathbf{y},\mathbf{x})$;}
\STATE{Construct $N$ linked GPs accordingly;}
\STATE{Compute $\tilde{\mu}^{(p)}_0$ and $(\tilde{\sigma}^{(p)}_0)^2$ of $Y^{(p)}_0(\mathbf{x}_0)$ using~\eqref{eq:mean_var} for all $p=1,\dots,P_L$.}
\end{algorithmic} 
\end{algorithm}

In a sampling-orientated FB inference the description of the posterior predictive distribution of $Y^{(p)}_0(\mathbf{x}_0)$ would require $N$ realizations of both latent variables and model parameters sampled from their posterior distributions. In addition, to obtain more precise estimates of posterior predictive mean and variance, FB also needs an adequate number of realizations of all latent variables at the prediction locations sampled through the posterior predictive distributions, for each of $N$ sampled latent variables and model parameters. The computational cost for this prediction procedure can be expensive in tasks such as DGP-based optimization and calibration that involve a large amount of DGP predictions at different input positions. Analogously, DSVI implements predictions via sampling and thus is exposed to the same issues of the FB approach. Besides, predictions made from DSVI (as well as other VI-based approaches) lose the interpolation property~\citep{hebbal2021bayesian} that is desired in emulating deterministic computer models. Our method combines the linked GP and an MCMC method, retaining interpolation and achieving closed form predictions (given multiple imputed latent variables) with thorough uncertainty quantification of predictions and imputations.

\subsection{Imputation}
\label{sec:latent}
\begin{figure}[!ht]
\centering
\scalebox{0.65}{
\begin{tikzpicture}[shorten >=1pt,->,draw=black!50, node distance=4cm]
    \tikzstyle{every pin edge}=[<-,shorten <=1pt]
    \tikzstyle{neuron}=[circle,fill=black!25,minimum size=50pt,inner sep=0pt]
    \tikzstyle{layer1}=[neuron, fill=green!50];
    \tikzstyle{layer2}=[neuron, fill=red!50];
    \tikzstyle{layer3}=[neuron, fill=blue!50];
    \tikzstyle{innerlayer}=[neuron, fill=white];
    \tikzstyle{annot} = [text width=2.25em, text centered]
    \tikzset{every pin edge/.style={draw=black}}
    \node[annot] (l1-0) at (-0.5,0) {$\mathbf{W}^{(1)}_{l-1}$};
    \node[annot] (l1-1) at (-0.5,-1.25) {$\mathbf{W}^{(2)}_{l-1}$};
    \node[annot] (l1-2) at (-0.5,-2.95) {$\mathbf{W}^{(P_{l-1})}_{l-1}$};
    \node[innerlayer] (l2-0) at (2.5,0.35) {};
    \node[layer2] (l2-1) at (2.5,-1.25) {$\mathcal{GP}^{(p)}_{l}$};
    \node[innerlayer] (l2-2) at (2.5,-3.3) {};
    \node[layer3,pin={[pin edge={->}]right:$\mathbf{W}^{(1)}_{l+1}$}] (ln-0) at (6.25,0.75) {$\mathcal{GP}^{(1)}_{l+1}$};
    \node[layer3, pin={[pin edge={->}]right:$\mathbf{W}^{(2)}_{l+1}$}] (ln-1) at (6.25,-1.25) {$\mathcal{GP}^{(2)}_{l+1}$};
    \node[layer3, pin={[pin edge={->}]right:$\mathbf{W}^{(P_{l+1})}_{l+1}$}] (ln-2) at (6.25,-3.7) {$\mathcal{GP}^{{(P_{l{+}1})}}_{l+1}$};
\path (l1-1) -- (l1-2) node [black, midway, sloped] {$\dots$};
\path (ln-1) -- (ln-2) node [black, midway, sloped] {$\dots$};
\path (ln-1) -- (ln-2) node [black, midway, sloped,transform canvas={xshift=19mm}] {$\dots$};
\draw[->,black!25,path fading=west] (l2-0) -- (ln-0);
\draw[->,black!25,path fading=west] (l2-0) -- (ln-1);
\draw[->,black!25,path fading=west] (l2-0) -- (ln-2);
\draw[->,black!25,path fading=west] (l2-2) -- (ln-0);
\draw[->,black!25,path fading=west] (l2-2) -- (ln-1);
\draw[->,black!25,path fading=west] (l2-2) -- (ln-2);
\draw[->,black!25,path fading=east] (l1-0) -- (l2-0);
\draw[->,black!25,path fading=east] (l1-0) -- (l2-2);
\draw[->,black!25,path fading=east] (l1-1) -- (l2-0);
\draw[->,black!25,path fading=east] (l1-1) -- (l2-2);
\draw[->,black!25,path fading=east] (l1-2) -- (l2-0);
\draw[->,black!25,path fading=east] (l1-2) -- (l2-2);
\draw[->,black] (l1-0) -- (l2-1);
\draw[->,black] (l1-1) -- (l2-1);
\draw[->,black] (l1-2) -- (l2-1);
\path [draw,black] (l2-1) -- (ln-0) node[font=\small,pos=0.5,fill=white,align=left,sloped] {$\mathbf{W}^{(p)}_{l}$};
\path [draw,black] (l2-1) -- (ln-1) node[font=\small,pos=0.5,fill=white,align=left,sloped] {$\mathbf{W}^{(p)}_{l}$};
\path [draw,black] (l2-1) -- (ln-2) node[font=\small,pos=0.5,fill=white,align=left,sloped] {$\mathbf{W}^{(p)}_{l}$};
\end{tikzpicture}}
\caption{The two-layered elementary DGP model that is targeted by ESS-within-Gibbs to sample a realization of output $\mathbf{W}^{(p)}_{l}$ from $\mathcal{GP}^{(p)}_{l}$ given all other latent variables.}
\label{fig:essingibbs}
\end{figure}
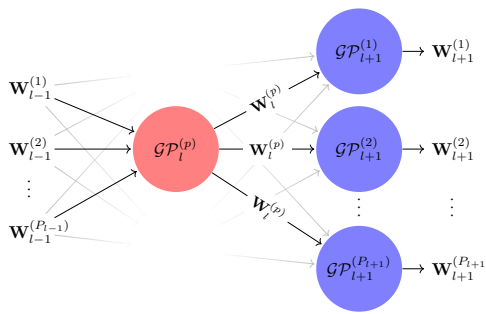

Exact simulation of latent variables $\{\mathbf{W}^{(p)}_l\}$ from $p(\{\mathbf{w}^{(p)}_l\}|\mathbf{y},\mathbf{x})$ is difficult because of the complexity of the posterior distribution induced by the deep hierarchy of GPs. Naive application of MCMC methods (that are poorly mixing and require fine tuning with considerable human intervention) can greatly reduce the efficiency and hinder the automation of DGP inference. Elliptical Slice Sampling (ESS), a rejection-free MCMC technique, has been shown~\citep{sauer2020active} to be a well-suited tuning-free method for latent variable simulations in three-layered DGP models. We thus utilize the ESS within a Gibbs sampler (ESS-within-Gibbs) to impute latent variables for the generic DGP model shown in Figure~\ref{fig:dgpmodel}. In general, the ESS is designed to sample from posterior $\pi(\mathbf{w})$ over the latent variable $\mathbf{w}\in\mathbb{R}^{M\times 1}$ of the form:
\begin{equation}
\label{eq:ess}
\pi(\mathbf{w})\propto\mathcal{L}(\mathbf{w})\mathcal{N}(\mathbf{w};\boldsymbol{\mu},\boldsymbol{\Sigma}),
\end{equation}
where $\mathcal{L}(\mathbf{w})$ is a likelihood function and $\mathcal{N}(\mathbf{w};\boldsymbol{\mu},\boldsymbol{\Sigma})$ is a multivariate normal prior of $\mathbf{w}$ with mean $\boldsymbol{\mu}$ and covariance matrix $\boldsymbol{\Sigma}$. Note that $p(\{\mathbf{w}^{(p)}_l\}|\mathbf{y},\mathbf{x})$ cannot be factorized into the form of~\eqref{eq:ess} and thus ESS cannot be directly applied. However, the conditional posteriors $p(\mathbf{w}^{(p)}_{l}|\{\mathbf{w}^{(p)}_l\}\setminus\mathbf{w}^{(p)}_{l},\mathbf{y},\mathbf{x})$ of the output from $\mathcal{GP}^{(p)}_{l}$ for some $p\in\{1,\dots,P_{l}\}$ and $l\in\{1,\dots,L-1\}$ can be expressed in the form of~\eqref{eq:ess} as follows:
\begin{equation}
\label{eq:elementary_pos}
p(\mathbf{w}^{(p)}_{l}|\{\mathbf{w}^{(p)}_l\}\setminus\mathbf{w}^{(p)}_{l},\mathbf{y},\mathbf{x})\propto \prod_{q=1}^{P_{l+1}}p(\mathbf{w}^{(q)}_{l+1}|\mathbf{w}^{(1)}_{l},\dots,\mathbf{w}^{(p)}_{l},\dots,\mathbf{w}^{(P_{l})}_{l})\,p(\mathbf{w}^{(p)}_{l}|\mathbf{w}^{(1)}_{l-1},\dots,\mathbf{w}^{(P_{l-1})}_{l-1}),
\end{equation}
where all terms are multivariate normal; $p(\mathbf{w}^{(p)}_{1}|\mathbf{w}^{(1)}_{0},\dots,\mathbf{w}^{(P_{0})}_{0})=p(\mathbf{w}^{(p)}_{1}|\mathbf{x})$ when $l=1$ and
\begin{equation*}
\prod_{q=1}^{P_{L}}p(\mathbf{w}^{(q)}_{L}|\mathbf{w}^{(1)}_{L-1},\dots,\mathbf{w}^{(p)}_{L-1},\dots,\mathbf{w}^{(P_{L-1})}_{L-1})=\prod_{q=1}^{P_{L}}p(\mathbf{y}^{(q)}|\mathbf{w}^{(1)}_{L-1},\dots,\mathbf{w}^{(p)}_{L-1},\dots,\mathbf{w}^{(P_{L-1})}_{L-1})
\end{equation*}
when $l=L-1$. Graphically, the Gibbs sampler allows the application of ESS for each latent variable $\mathbf{W}^{(p)}_{l}$ from a two-layered elementary DGP shown in Figure~\ref{fig:essingibbs}. A single-step ESS-within-Gibbs that draws a realization from $p(\{\mathbf{w}^{(p)}_l\}|\mathbf{y},\mathbf{x})$ is given in Algorithm~\ref{alg:ess}, where the algorithm for the ESS update on Line~\ref{alg:one_ess} is given in~\citet[Algorithm 1]{nishihara2014parallel}.

\begin{algorithm}[!ht]
\caption{One-step ESS-within-Gibbs to sample from $p(\{\mathbf{w}^{(p)}_l\}|\mathbf{y},\mathbf{x})$}
\label{alg:ess}
\begin{algorithmic}[1]
\REQUIRE{A current sample $\{\mathbf{w}^{(p)}_l\}_i$ drawn from $p(\{\mathbf{w}^{(p)}_l\}|\mathbf{y},\mathbf{x})$.}
\ENSURE{A new sample $\{\mathbf{w}^{(p)}_l\}_{i+1}$ drawn from $p(\{\mathbf{w}^{(p)}_l\}|\mathbf{y},\mathbf{x})$.}
\FOR{$l=1,\dots,L-1$}
\FOR{$p=1,\dots,P_l$}
\STATE{\label{alg:one_ess}Draw $\mathbf{w}^{(p)}_l$ from $p(\mathbf{w}^{(p)}_{l}|\{\mathbf{w}^{(p)}_l\}\setminus\mathbf{w}^{(p)}_{l},\mathbf{y},\mathbf{x})$ in the form of~\eqref{eq:elementary_pos} via an ESS update;}
\ENDFOR
\ENDFOR
\end{algorithmic} 
\end{algorithm}

\subsection{Training}
We have so far assumed that the model parameters $\boldsymbol{\theta}^{(p)}_l$ of $\mathcal{GP}^{(p)}_l$ are known. In this section, we detail how these parameters are optimized under SI. A naive training for the DGP model in Figure~\ref{fig:dgpmodel} might be to impute the latent variables $\{\mathbf{W}^{(p)}_l\}$ by sampling from the imputer $p(\{\mathbf{w}^{(p)}_l\}|\mathbf{y},\mathbf{x})$ and then to optimize the model parameters $\{\boldsymbol{\theta}^{(p)}_l\}$ following the training procedure for conventional GPs. However, $\{\boldsymbol{\theta}^{(p)}_l\}$ are also required by $p(\{\mathbf{w}^{(p)}_l\}|\mathbf{y},\mathbf{x})$ and thus we should update our imputer with our current best guess (in the sense of the maximum likelihood given the imputed latent variables) of the model parameters. We thus use an iterative training process, called the Stochastic Expectation-Maximization (SEM) algorithm~\citep{celeux1985sem}, that updates model parameters at a given iteration $t\in\{1,\dots,T-1\}$ via the following two steps:
\begin{itemize}
    \item \textbf{\underline{I}mputation-step}: impute the latent variables $\{\mathbf{W}^{(p)}_l\}$ by a single realization $\{\mathbf{w}^{(p)}_l\}$ drawn from the imputer $p(\{\mathbf{w}^{(p)}_l\}|\mathbf{y},\mathbf{x};\{\widehat{\boldsymbol{\theta}}^{(p,t)}_l\})$ given estimates $\{\widehat{\boldsymbol{\theta}}^{(p,t)}_l\}$ of $\{\boldsymbol{\theta}^{(p)}_l\}$;
    \item \textbf{\underline{M}aximization-step}: given the pseudo-complete data $\{\mathbf{y},\{\mathbf{w}^{(p)}_l\},\mathbf{x}\}$, update $\{\widehat{\boldsymbol{\theta}}^{(p,t)}_l\}$ to $\{\widehat{\boldsymbol{\theta}}^{(p,t+1)}_l\}$ by maximizing the likelihood function $\mathcal{L}(\{\boldsymbol{\theta}^{(p)}_l\})=p(\mathbf{y},\{\mathbf{w}^{(p)}_l\}|\mathbf{x};\{\boldsymbol{\theta}^{(p)}_l\})$, which amounts to separate optimization problems of individual GPs; update the imputer to $p(\{\mathbf{w}^{(p)}_l\}|\mathbf{y},\mathbf{x};\{\widehat{\boldsymbol{\theta}}^{(p,t+1)}_l\})$ with the optimized model parameter estimates $\{\widehat{\boldsymbol{\theta}}^{(p,t+1)}_l\}$.  
\end{itemize}

By alternating a stochastic I-step and a deterministic M-step, SEM produces a Markov chain $\{\widehat{\boldsymbol{\theta}}^{(p,1)}_l\},\dots,\{\widehat{\boldsymbol{\theta}}^{(p,T)}_l\}$ that does not converge pointwise but contains points that represent best (i.e., maximum complete-data likelihood) estimates of model parameters given a sequence of plausible values of latent variables~\citep{ip1994stochastic,nielsen2000stochastic,ip2002single}, and one can then establish pointwise estimates $\{\widehat{\boldsymbol{\theta}}^{(p)}_l\}$ of model parameters by averaging the chain after discarding burn-in periods $B$ ~\citep{diebolt1996stochastic}:
\begin{equation}
\label{eq:estimates}
  \widehat{\boldsymbol{\theta}}^{(p)}_l=\frac{1}{T-B}\sum_{t=B+1}^{T}\widehat{\boldsymbol{\theta}}^{(p,t)}_l\quad\forall\,p,l.
\end{equation}
For computational and numerical advantages of SEM over EM and other stochastic EM variants, e.g., Monte Carlo EM~\citep{wei1990monte}, see~\citet{celeux1996stochastic,ip2002single}.

The SEM algorithm forms a key part of our DGP inference because it has properties that make SI competitive for DGP training in comparison to FB and DSVI. FB trains the DGP by applying MCMC methods to both latent variables and model parameters. Although it captures the model uncertainty more thoroughly (in principle, albeit not always in practice due to MCMC issues on sampling model parameters), it has several computational disadvantages in comparison to SI. Firstly, FB needs to store sampled latent variables in addition to sampled model parameters and thus can require a substantial amount of memory if the length of chain is long or the number of elementary GP nodes in the DGP is large. SI, instead, only stores updated model parameter estimates produced over iterations and is therefore more memory-efficient. The MCMC sampling in FB over the model parameters can also be computationally expensive itself (long chains of Gibbs-type draws with multiple evaluations of correlation matrix inversions at each draw). Rather than sampling, SI breaks the training problem of DGP into simpler and faster optimization problems of individual GPs and updates all model parameters in the GPs simultaneously with little human intervention. As SEM can be shown to be a stochastic perturbation of the EM dynamics~\citep{ip2002single}, the training of SI can be expected to stabilize in a comparatively small number of iterations.

DSVI trains the DGP by maximizing the ELBO, which involves a large number of model parameters including kernel hyper-parameters, variational parameters and inducing point locations for each layer. Although optimization of the ELBO is computationally tractable, it embeds a simplified assumption on latent posteriors and thus can underestimate predictive uncertainties. In contrast, SI only involves optimizations of conventional GPs with respect to kernel hyper-parameters using latent posteriors that are exploited thoroughly via ESS. This makes it particularly suitable for surrogate modeling for UQ, where we often have small-to-moderate data that are generated by computationally expensive simulators, and where accurate quantification of posterior uncertainties is essential. 

The pseudo-code for DGP training in SI via SEM is given in Algorithm~\ref{alg:training}. It may be argued that a large $C$ is needed in the I-step of Algorithm~\ref{alg:training} in order to draw a realization from the stationary distribution of the imputer, and thus the training of SI can be computationally expensive to implement. However, since SEM can be seen as an example of the data augmentation method~\citep{celeux1996stochastic}, in practice one does not need a large $C$ for effective inference~\citep{ip1994stochastic,zhang2020improved}. In our experience, $C=10$ is often sufficient to obtain appropriate samples from the imputer. In addition, since in each I-step SEM only requires one realization, it is not essential to conduct convergence assessment of ESS-within-Gibbs, which is not the case for the FB approach.

\begin{algorithm}[!ht]
\caption{Training algorithm for the DGP model in Figure~\ref{fig:dgpmodel} using SI via SEM}
\label{alg:training}
\begin{algorithmic}[1]
\REQUIRE{\begin{enumerate*}[label=(\roman*)]
  \item Observations $\mathbf{x}$ and $\mathbf{y}$;
  \item initial values of model parameters $\{\widehat{\boldsymbol{\theta}}^{(p,1)}_l\}$;
  \item total number of iterations $T$ and burn-in period $B$ for SEM; 
  \item burn-in periods $C$ for ESS. 
\end{enumerate*}}
\ENSURE{Point estimates ${\widehat{\boldsymbol{\theta}}^{(p)}_l}$ of model parameters.}
\FOR{$t=1,\dots,T-1$}
\STATE{\textbf{I-step}: draw a realization $\{\mathbf{w}^{(p)}_l\}$ from the imputer $p(\{\mathbf{w}^{(p)}_l\}|\mathbf{y},\mathbf{x};\{\widehat{\boldsymbol{\theta}}^{(p,t)}_l\})$ by evaluating $C$ steps of ESS-within-Gibbs in Algorithm~\ref{alg:ess};}
\STATE{\textbf{M-step}: update model parameters by solving individual GP training problems: ${\widehat{\boldsymbol{\theta}}^{(p,t+1)}_l}=\argmax\log p(\mathbf{w}^{(p)}_{l}|\mathbf{w}^{(1)}_{l-1},\dots,\mathbf{w}^{(P_{l-1})}_{l-1};\boldsymbol{\theta}^{(p)}_l)$ for all $p,l$.}
\ENDFOR
\STATE{Compute point estimates ${\widehat{\boldsymbol{\theta}}^{(p)}_l}$ of model parameters by equation~\eqref{eq:estimates}.}
\end{algorithmic} 
\end{algorithm}

To deliver complete inference for the DGP model in Figure~\ref{fig:dgpmodel}, one starts from Algorithm~\ref{alg:training} to obtain estimates of model parameters for all individual GPs. Given the trained DGP (i.e., a network of trained individual GPs $\{\mathcal{GP}^{(p)}_l\}$), one can then proceed to make predictions at new input locations using Algorithm~\ref{alg:prediction}, in which the multiple imputation step on Line~\ref{alg:mi} is achieved by invoking  Algorithm~\ref{alg:ess} multiple ($N$) times.

\section{Step Function}
\label{sec:step_fct}
Consider a synthetic computer model with a step-wise functional form:
\begin{equation*}
    f(x)=\begin{cases}
    1,\quad 0.5\leq x<1\\
    -1,\quad 0\leq x<0.5
    \end{cases}
\end{equation*}
with input domain $[0,1]$. In this experiment, we consider a three-layered DGP, where each layer contains only one GP (i.e., $P_1=P_2=P_3=1$). Different inference approaches are compared by first measuring the predictive accuracy of the trained DGP in terms of the Normalized Root Mean Squared Error of Predictions (NRMSEPs):
\begin{equation*}
        \mathrm{NRMSEP}=\frac{\sqrt{\frac{1}{n}\sum_{i=1}^{n}(f(x_{0i})-\tilde{\mu}_{0i})^2}}{\max\{f(x_{0i})_{i=1,\dots,n}\}-\min\{f(x_{0i})_{i=1,\dots,n}\}},
\end{equation*}
where $f(x_{0i})$ and $\tilde{\mu}_{0i}$ denotes respectively the true output of the computer model and mean prediction from the trained DGP evaluated at the testing input position $x_{0i}$ for $i=1,\dots,n$. We then check if the uncertainty quantified by the trained DGP provides sensible indications of input region (i.e., the discontinuity at $x=0.5$) that is deemed important by examining the produced predictive standard deviation $\tilde{\sigma}_{0i}$ for $i=1,\dots,n$.      

\subsection{Implementation}
Ten equally spaced design points were chosen over the input domain $[0,1]$, whose corresponding output are computed by evaluating the synthetic computer model. We select $n=200$ testing points whose inputs are equally spaced over the input domain. For FB, we use the \texttt{R} package \texttt{deepgp} (available at \url{ https://CRAN.R-project.org/package=deepgp}) by setting the total number of MCMC simulations to $10000$ and the burn-in period to $8000$ with thinning by half. These are default settings used in exercises of~\citet{sauer2020active}. DSVI is implemented using the \texttt{Python} library \texttt{GPflux} (available at \url{https://github.com/secondmind-labs/GPflux}). To ensure a fair comparison to FB and SI, we switch off the sparse approximation of DSVI by setting the number of inducing points to be same as the number of training data points (i.e., $10$). The ELBO is maximized using the \texttt{Adam} optimizer~\citep{kingma2015adam} with the learning rate of $0.01$ and $1000$ iterations. These are the standard settings in \texttt{GPflux} for ELBO optimization. With regard to SI, we implemented it using our \texttt{Python} package \texttt{dgpsi}. The total number of SEM iterations, $T$, is set to $500$ with the first $75\%$ (i.e., $375$) of total iterations being the burn-in period $B$. The warm-up period $C$ for the ESS in the I-step of SEM is set to $10$. $50$ imputations are conducted to make predictions at the testing input positions. These are default settings in \texttt{dgpsi}. Since the default DSVI implementation mimics the effects of the input-connected structure introduced in~\citet{duvenaud2014avoiding} through the use of linear mean functions~\citep{salimbeni2017doubly}, we also explore the benefit of the input connection (IC) to SI by explicitly augmenting the input of GPs (in all layers except for those in the first layer) with the global input $\mathbf{x}$. The SI with the input connection is referred to as SI-IC hereinafter. For all approaches, we use squared exponential kernels. The nugget term (the likelihood variance in the case of DSVI) is set to a small value ($\sim 10^{-6}$) for interpolation. Unless otherwise stated, we use the same setup for different inference methods in the remainder of this study. Although the objective of the study is to introduce SI by comparing it with other inference approaches rather than comparing DGP to other GP models, in this and all remaining examples we also report results given by a conventional GP, following~\citet{salimbeni2017doubly,sauer2020active}, because the conventional GP can be seen as a one-layered DGP and is still the most widely used model for emulation. The conventional GP is trained by the \texttt{R} package \texttt{RobustGaSP}~\citep{gu2018robustgasp} .

\subsection{Results}
\begin{figure}[htbp]
\centering 
\subfloat[GP]{\label{fig:gp}\includegraphics[width=0.2\linewidth]{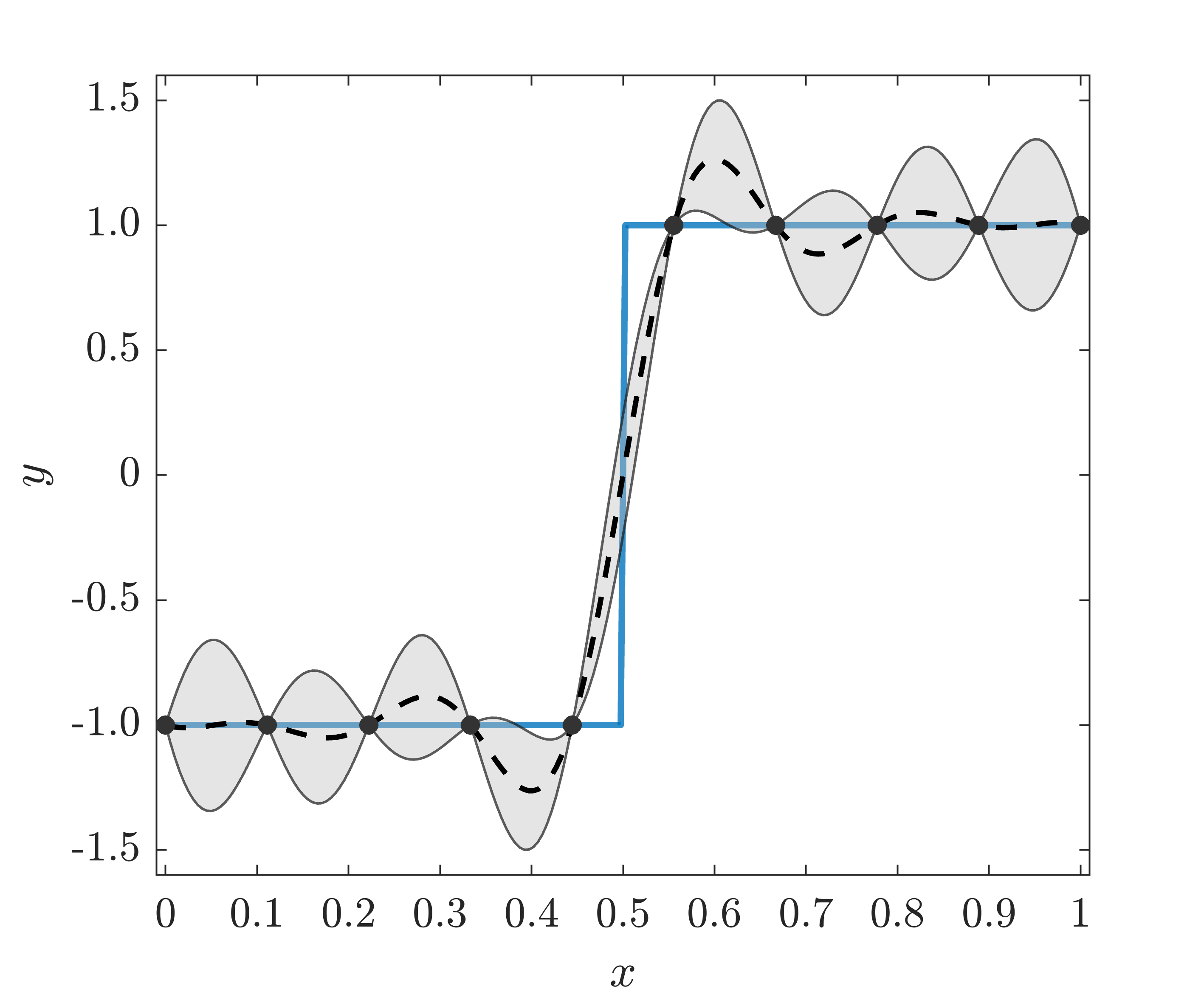}}
\subfloat[DGP (FB)]{\label{fig:dgp_sexp_fb}\includegraphics[width=0.2\linewidth]{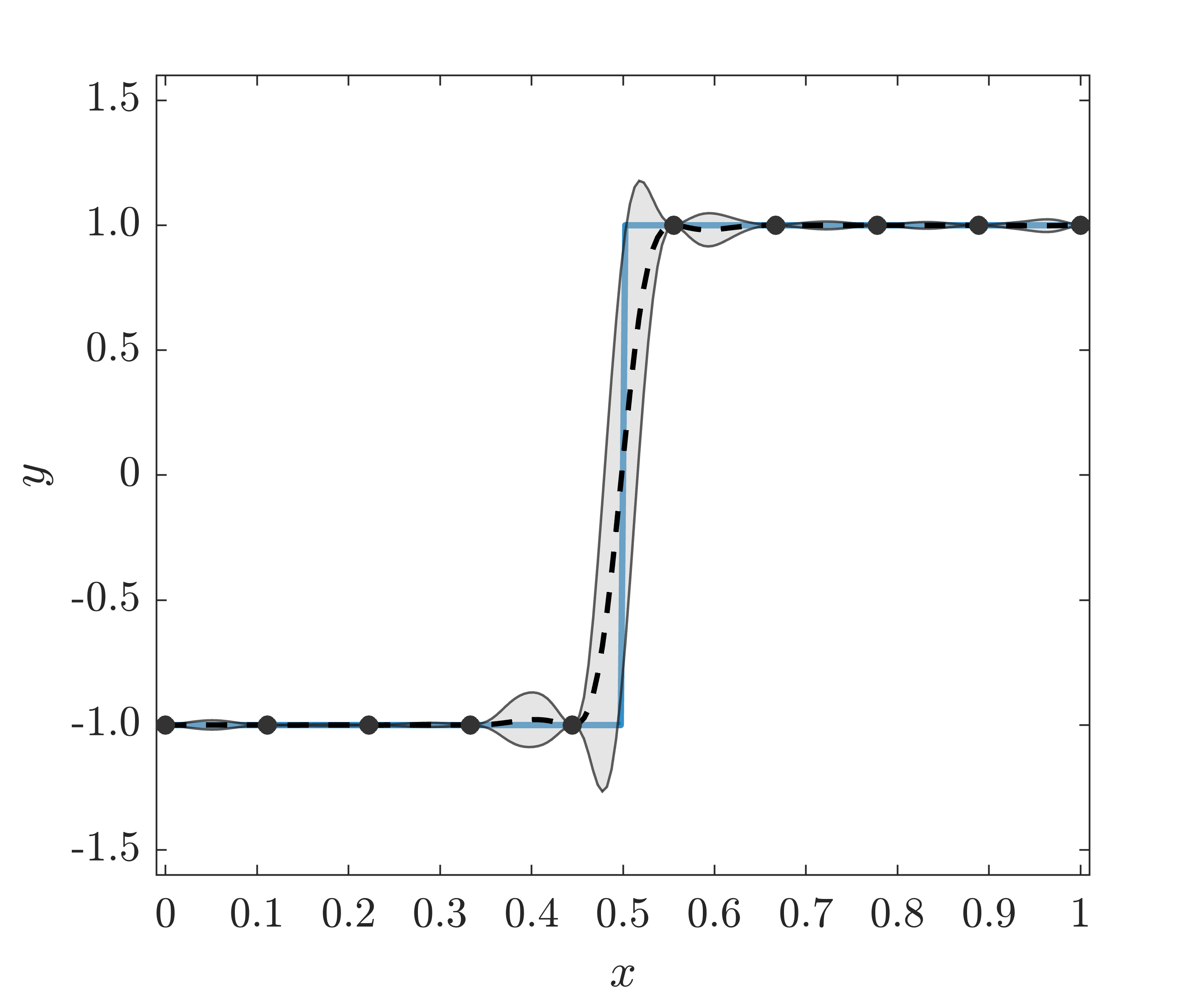}}
\subfloat[DGP (DSVI)]{\label{fig:dgp_sexp_vi}\includegraphics[width=0.2\linewidth]{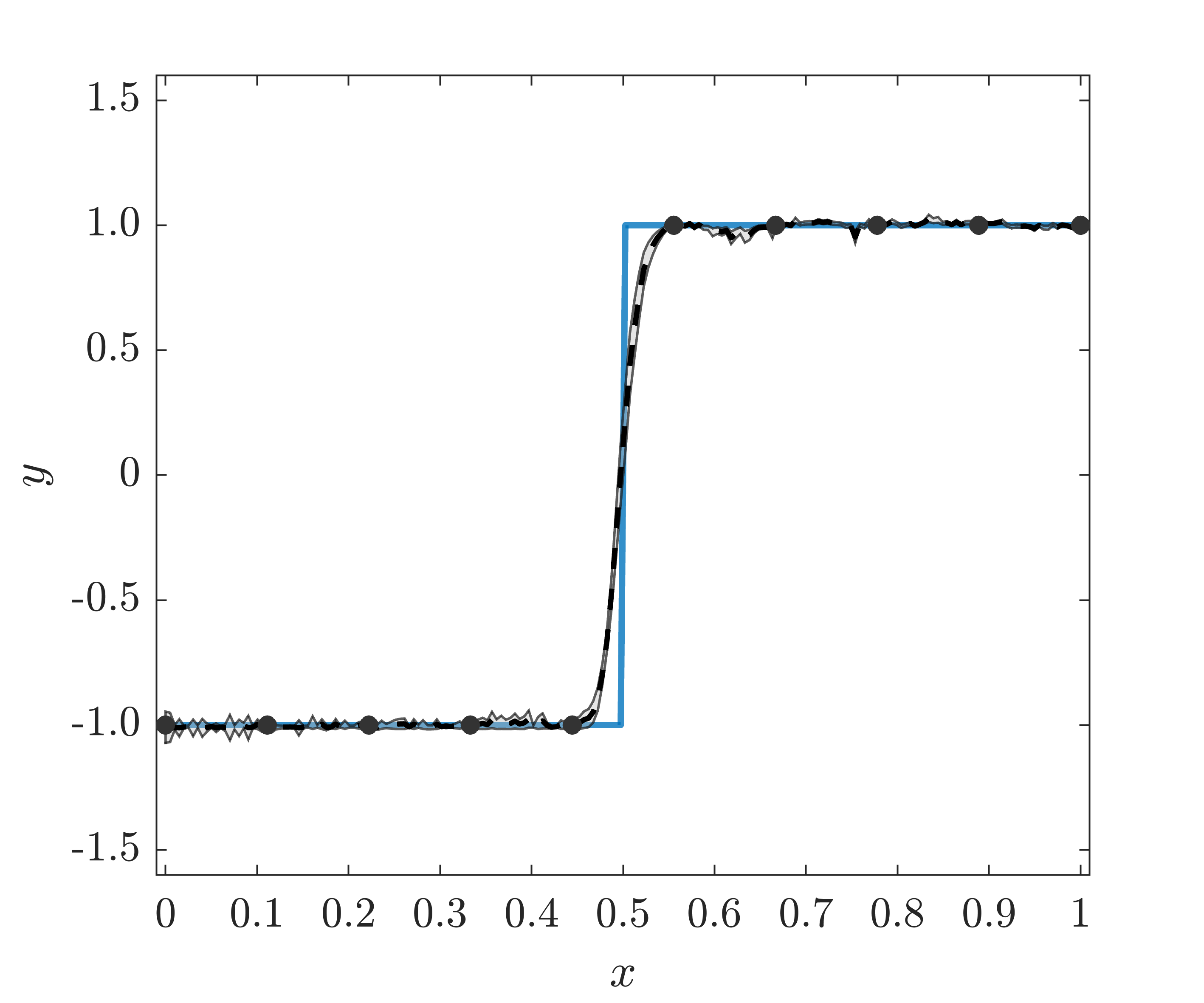}}
\subfloat[DGP (SI)]{\label{fig:dgp_sexp_is}\includegraphics[width=0.2\linewidth]{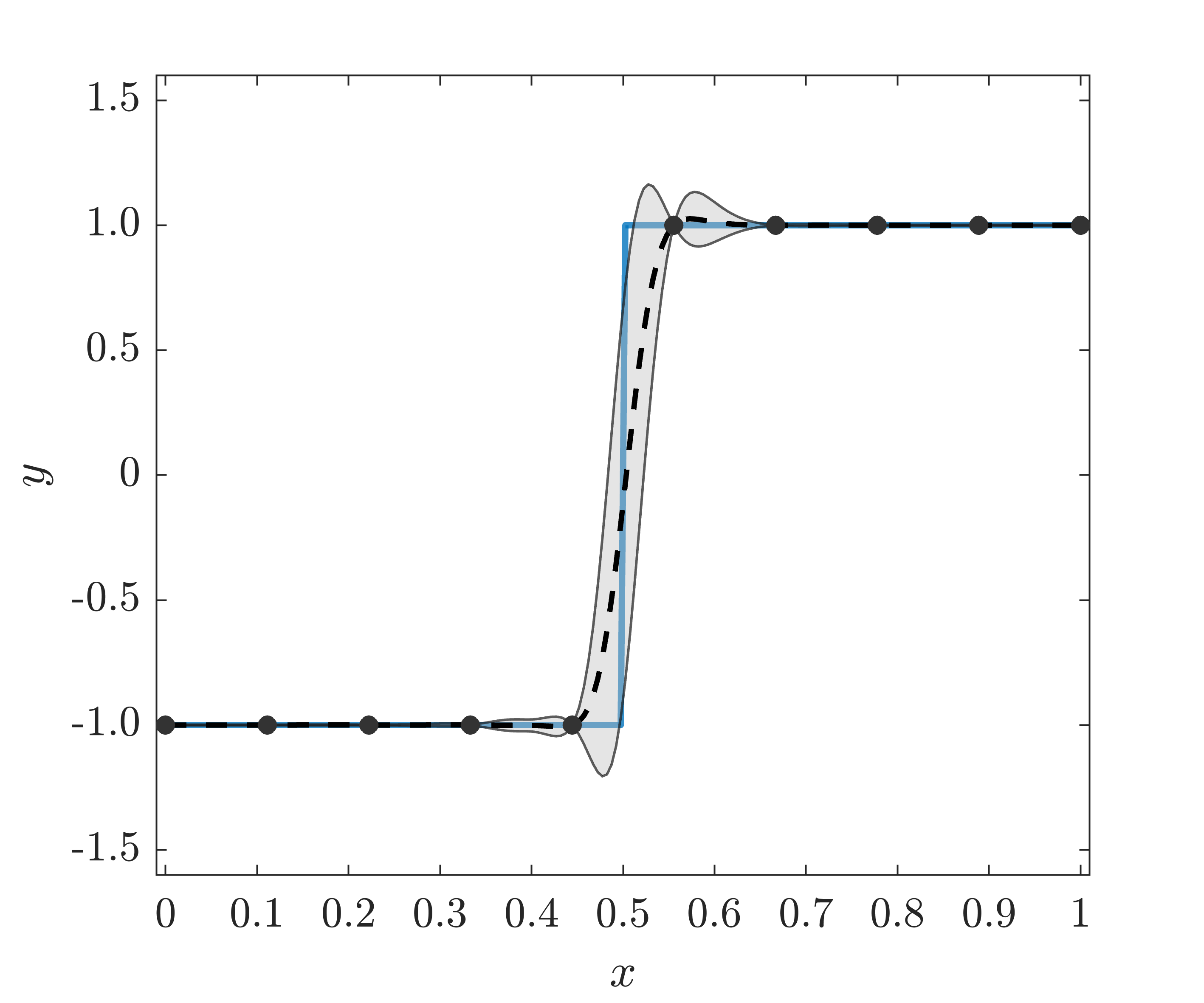}}
\subfloat[DGP (SI-IC)]{\label{fig:dgp_sexp_is_ic}\includegraphics[width=0.2\linewidth]{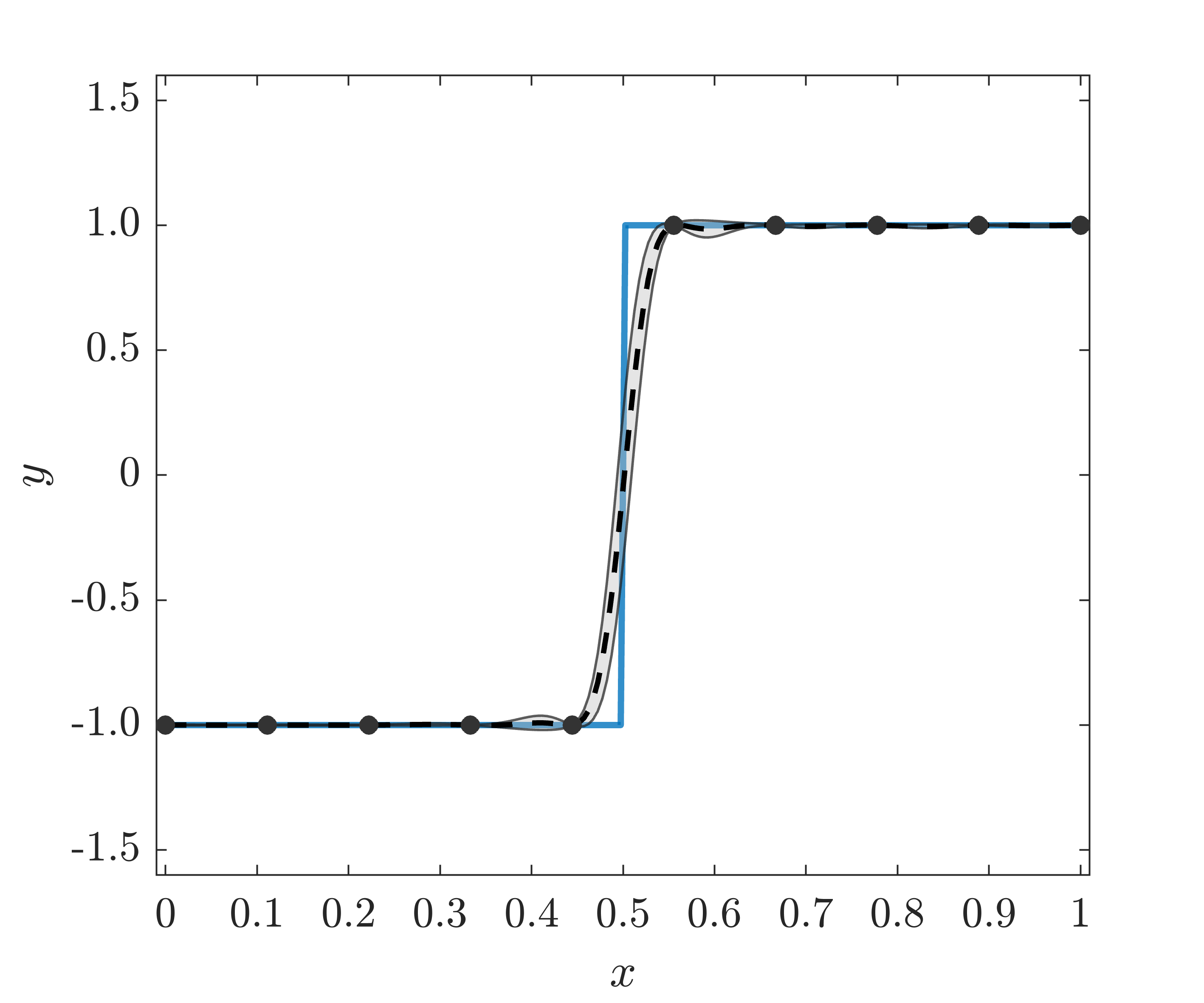}} 
\caption{DGP emulators of the step function (the solid line) trained by different inference methods. The dashed line is the mean prediction; the shaded area is the predictive interval (i.e., two predictive standard deviations above and below the predictive mean); the filled circles are training points.}
\label{fig:compare}
\end{figure}

\begin{figure}[!ht]
\centering 
\subfloat[GP]{\label{fig:sd_gp}\includegraphics[width=0.2\linewidth]{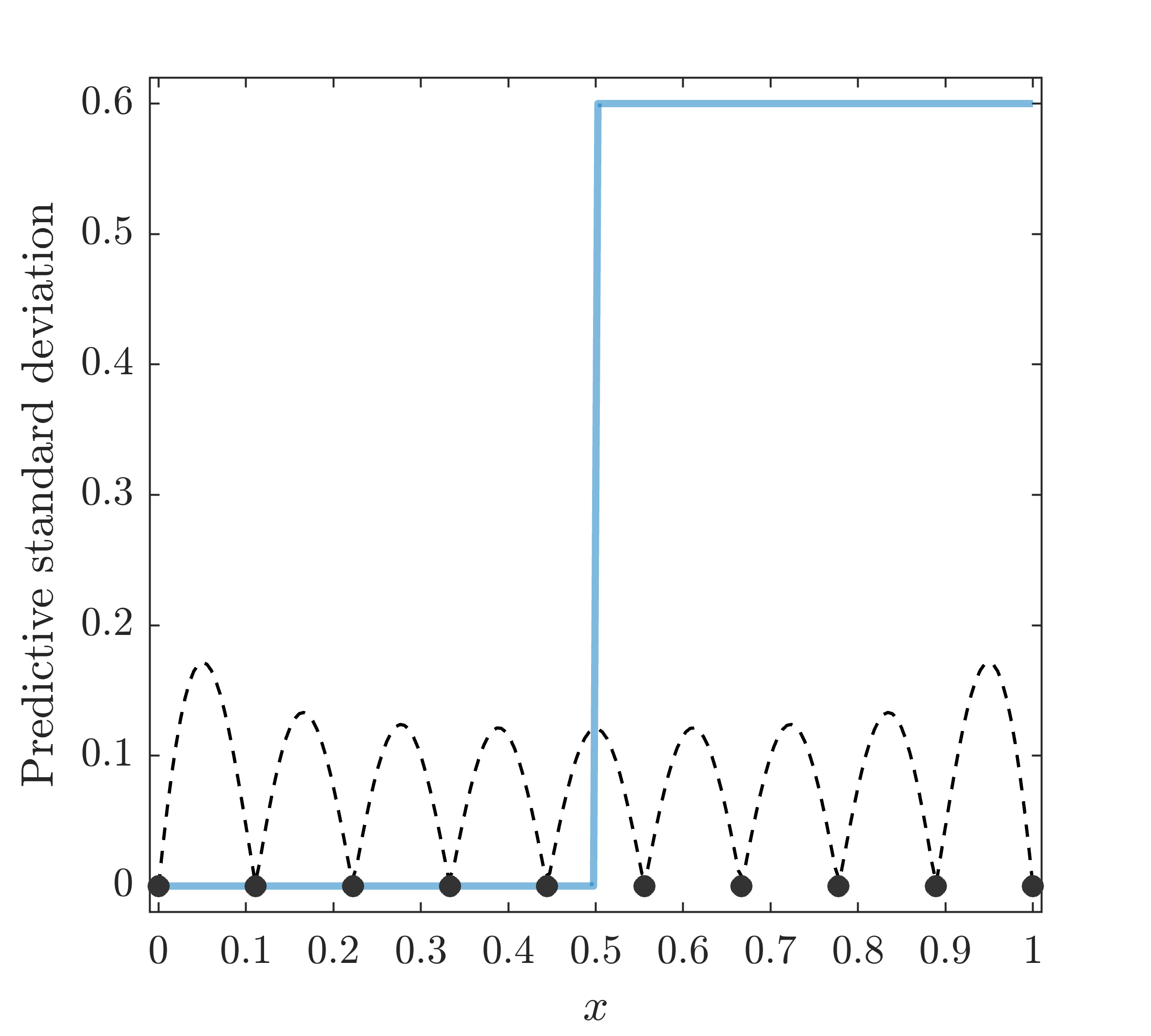}} 
\subfloat[FB]{\label{fig:sd_sexp_fb}\includegraphics[width=0.2\linewidth]{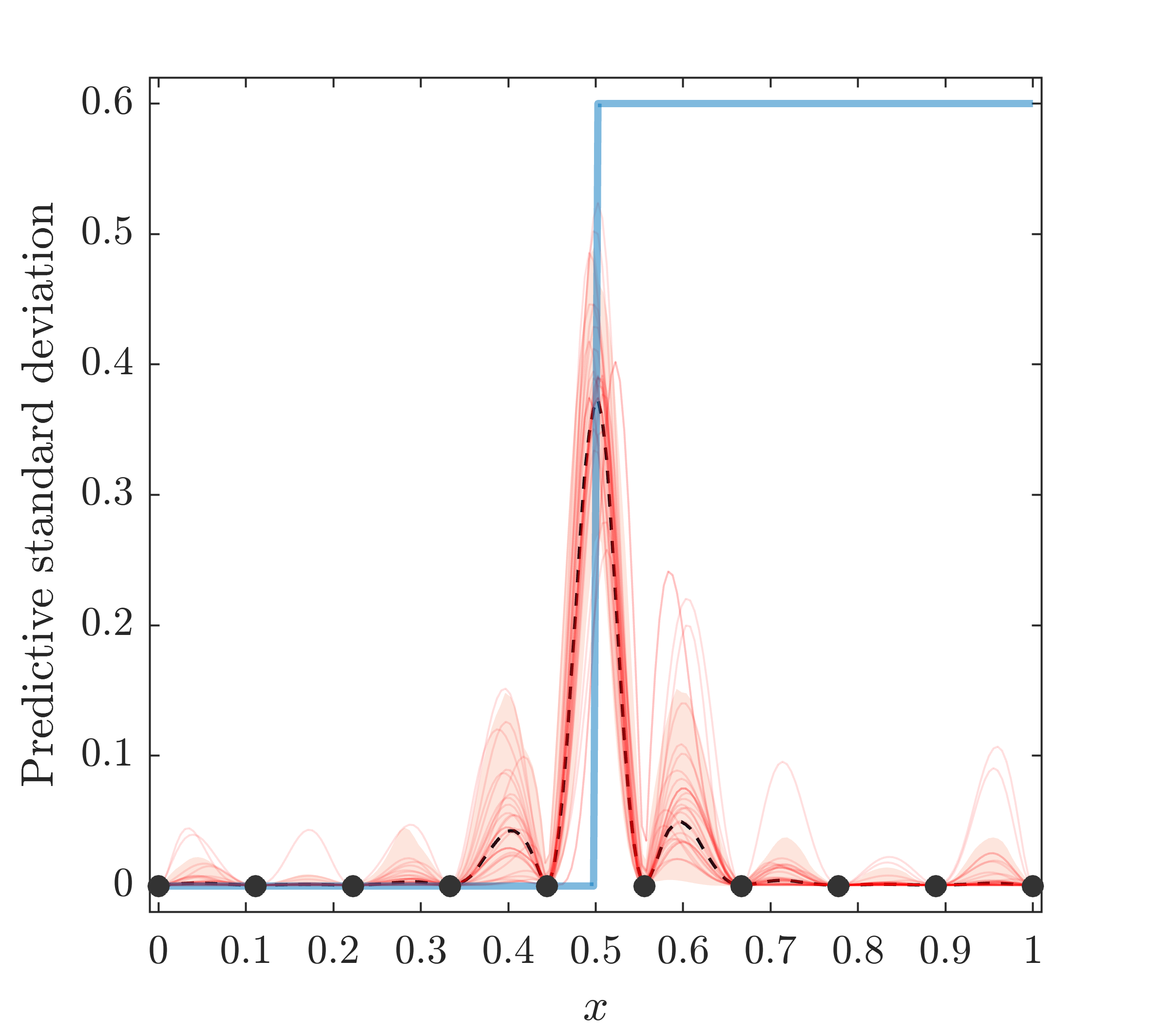}}
\subfloat[DSVI]{\label{fig:sd_sexp_vi}\includegraphics[width=0.2\linewidth]{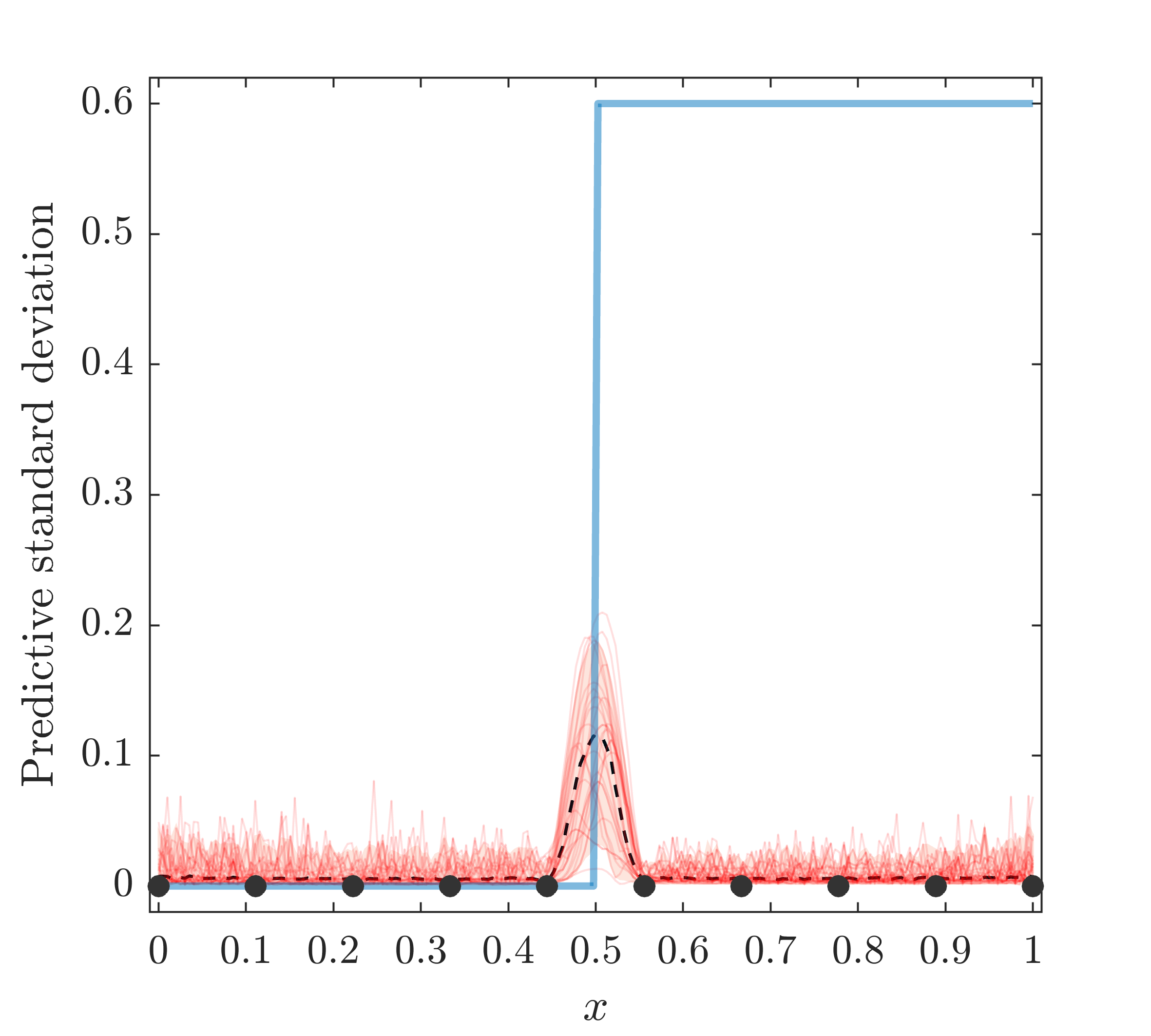}}
\subfloat[SI]{\label{fig:sd_sexp_is}\includegraphics[width=0.2\linewidth]{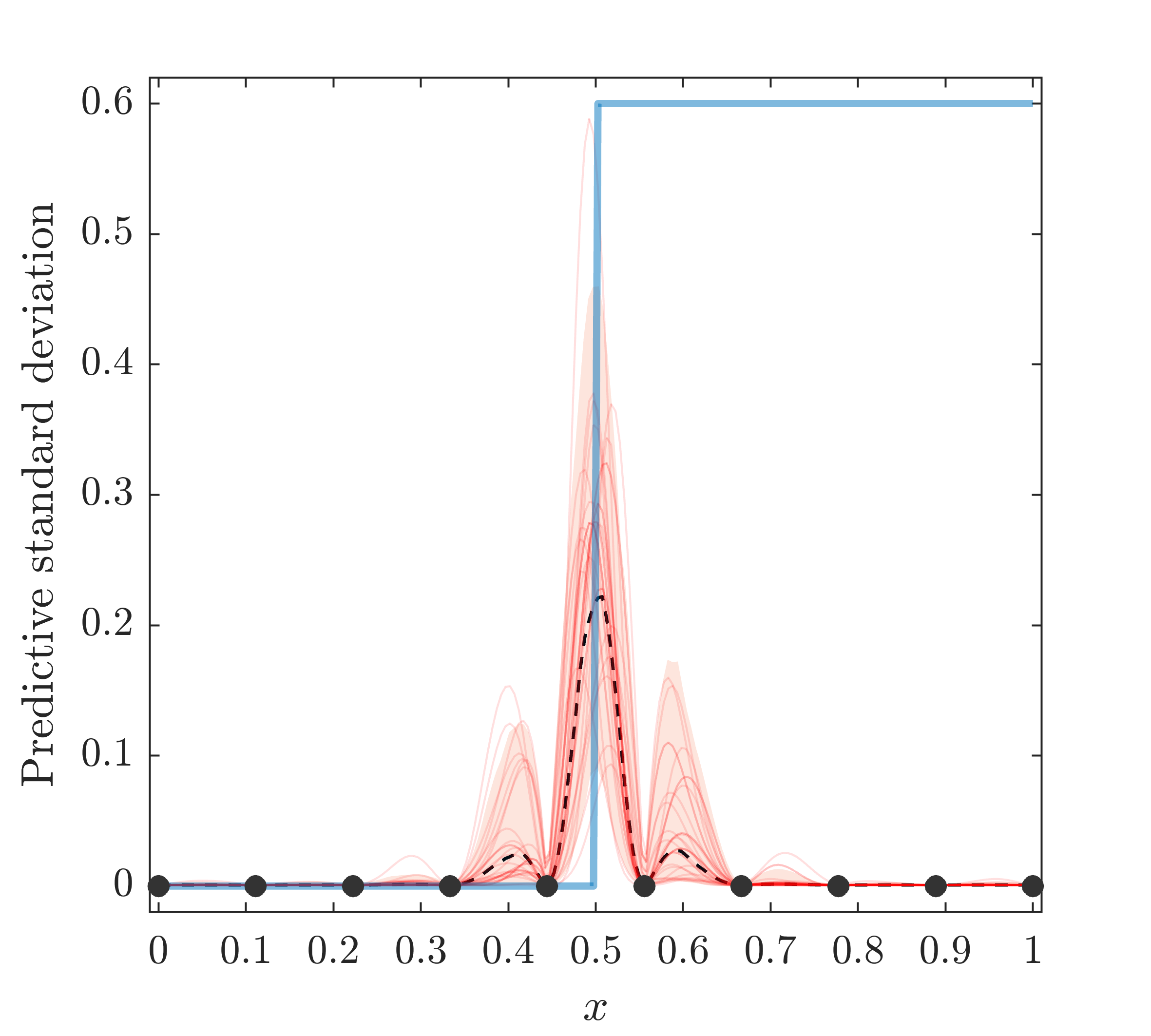}}
\subfloat[SI-IC]{\label{fig:sd_sexp_is_ic}\includegraphics[width=0.2\linewidth]{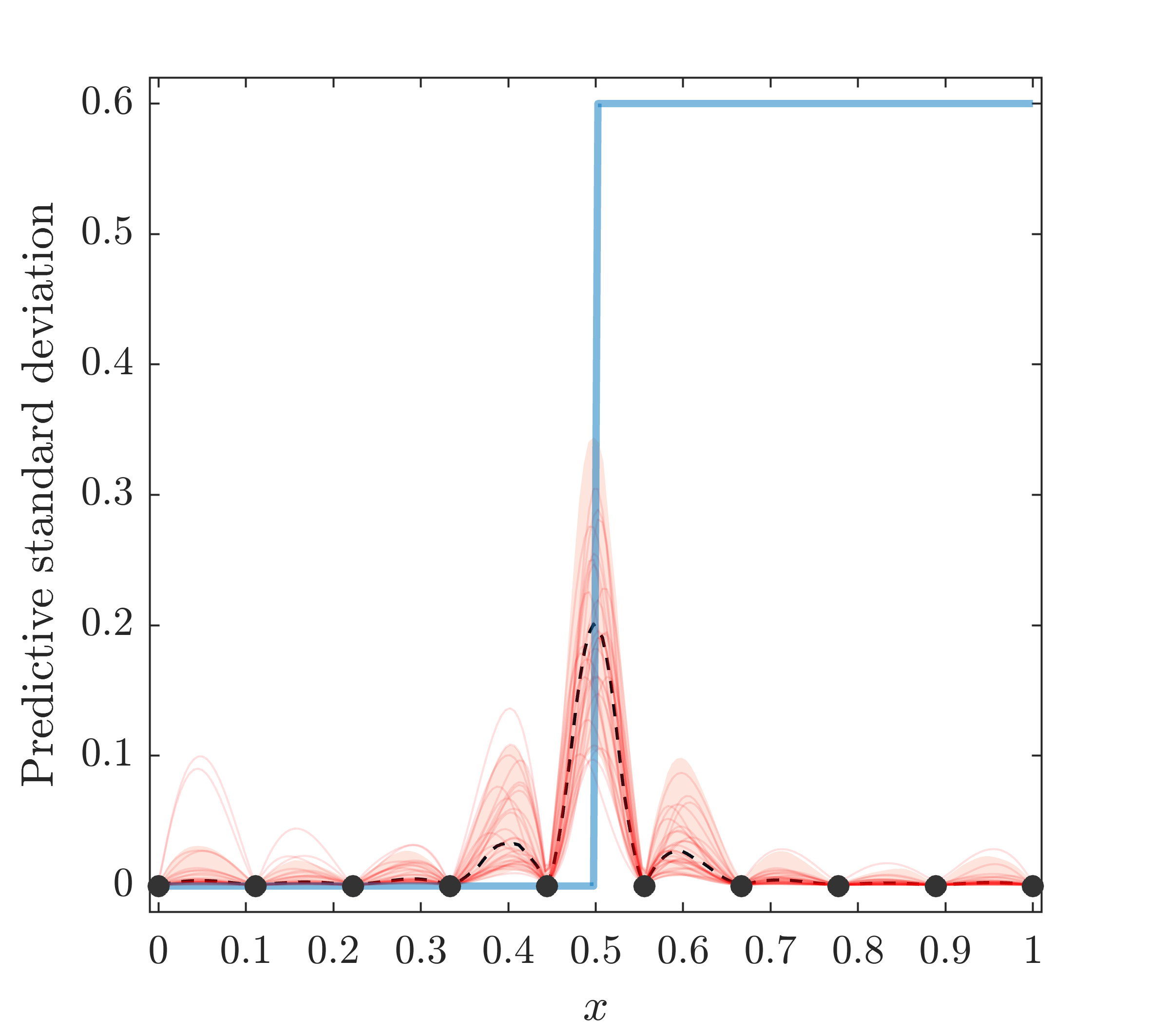}}
\caption{Predictive standard deviations of GP and DGP emulators over the input domain. The shaded area in~\emph{(b)} to \emph{(e)} represents the interval between the $5$-th and $95$-th percentiles (with the dash line highlighting the $50$-th percentile) of $100$ predictive standard deviations produced by the corresponding $100$ repeatedly trained DGPs; $30$ out of $100$ predictive standard deviations are randomly selected and drawn as the solid lines in~\emph{(b)} to \emph{(e)}. The underlying true step function and  training input locations (shown as filled circles) are projected into all sub-figures.}
\label{fig:sd}
\end{figure}

It is apparent from Figure~\ref{fig:compare} that, regardless of the inference method, the DGP model outperforms the GP model in emulating the underlying step function. The DGPs trained by FB, SI, and SI-IC provide better mean predictions than that trained by the DSVI. Both SI and FB quantify larger uncertainties than DSVI and SI-IC around the discontinuity of the step function. As addressed in Section~\ref{sec:prediction}, the DGP emulator trained by DSVI loses the interpolation property as the predictive uncertainties do not reduce to zero at some training data points. To examine the variability of such observations on predictive uncertainties under the randomness (due to latent simulations) involved in different methods, we repeat each inference approach (except for the conventional GP) $100$ times and summarize predictive standard deviations across different trials in Figure~\ref{fig:sd}. It is clear from Figure~\ref{fig:sd} that FB, SI, and SI-IC produce DGP emulators with better uncertainty quantification of the underlying step function than DSVI does because they highlight locations where abrupt functional transitions present with sufficiently higher predictive standard deviations. 

\begin{figure}[!ht]
\centering 
\subfloat[NRMSEP]{\label{fig:nrmse}\includegraphics[width=0.45\linewidth]{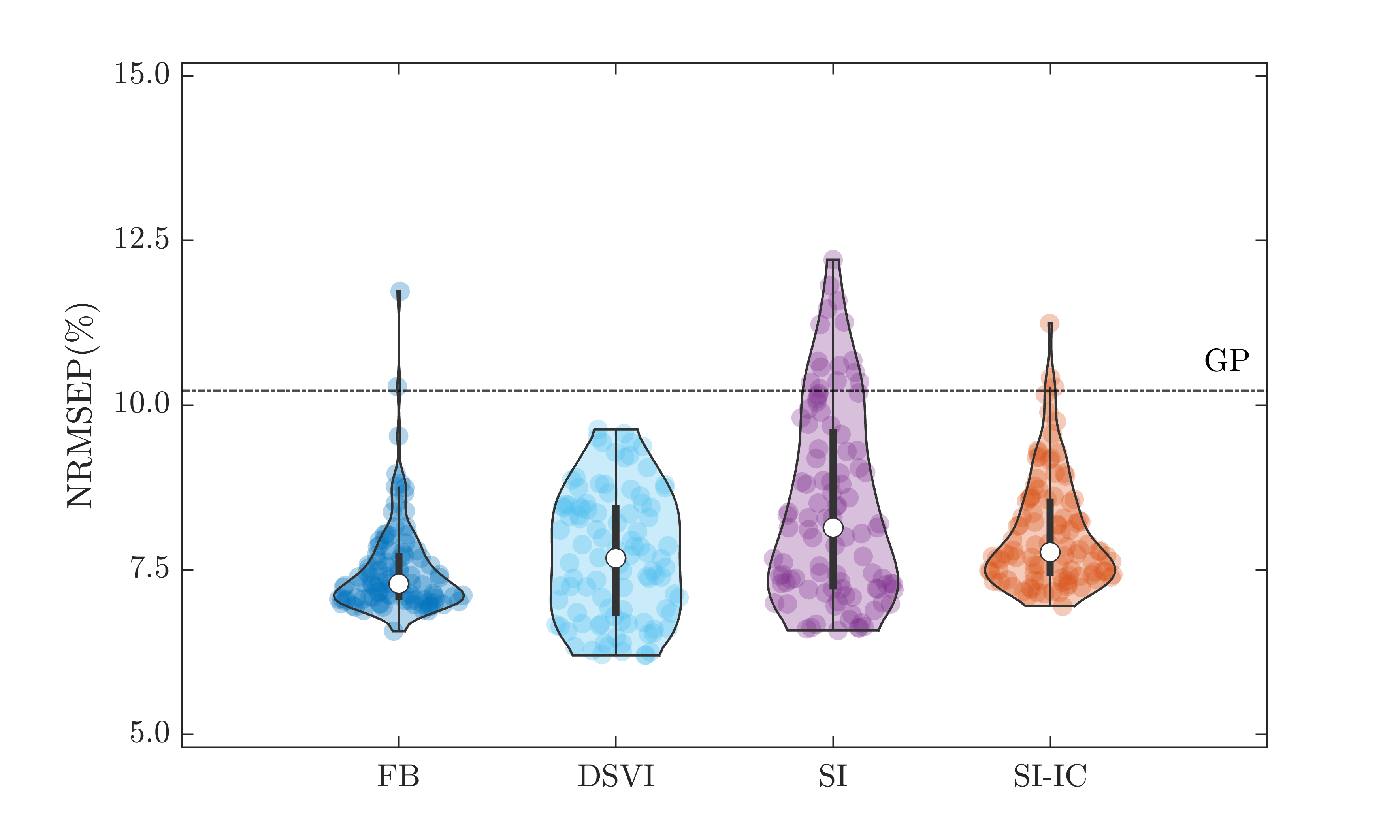}}
\subfloat[Computation time]{\label{fig:time}\includegraphics[width=0.45\linewidth]{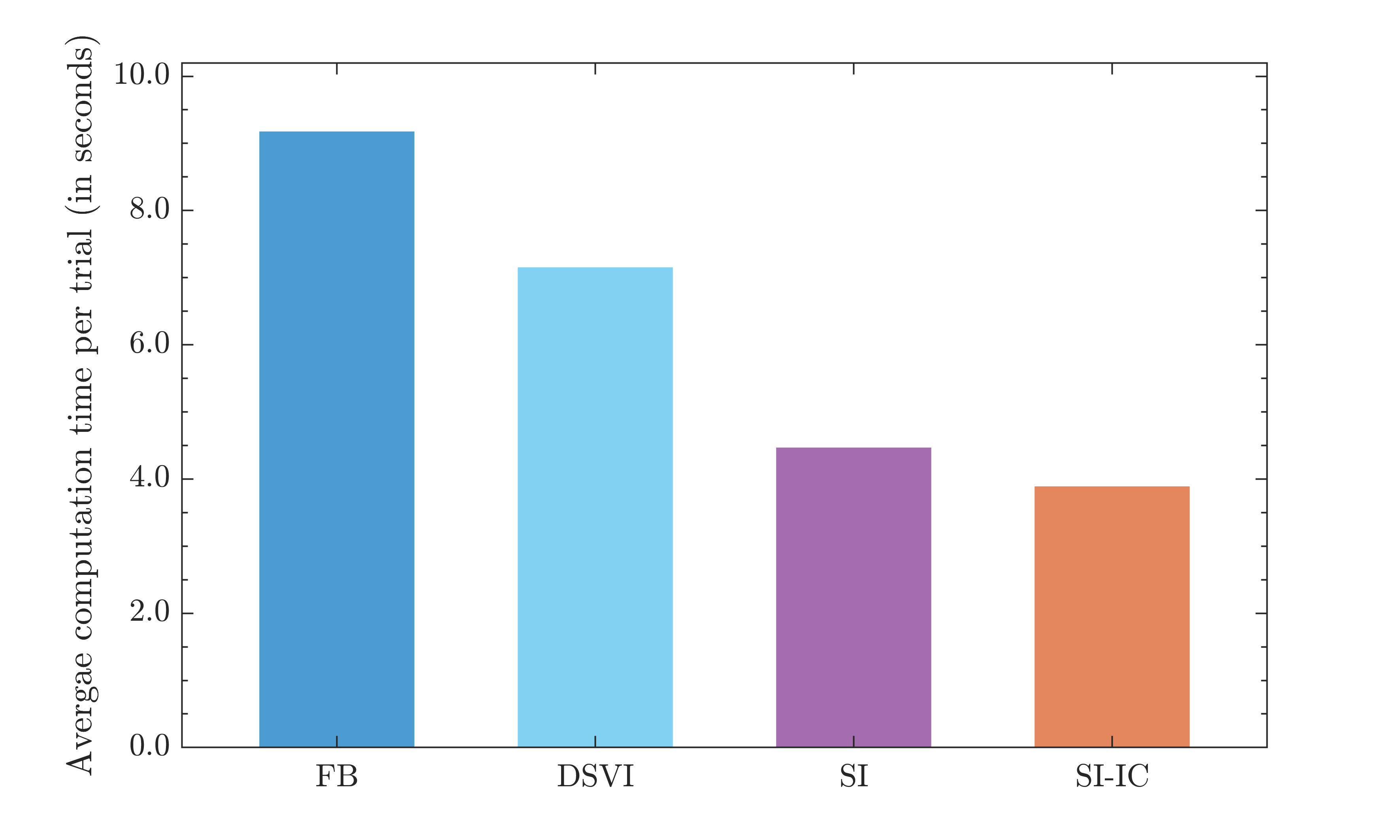}} 
\caption{Comparison of FB, DSVI, SI, and SI-IC across $100$ repeatedly trained DGP emulators and the corresponding implementation packages' computation time. \emph{(a)}: Violin plots of Normalized Root Mean Squared Error of Predictions (NRMSEPs). The dash-dot line represents the trained conventional GP. \emph{(b)}: Average computation time (including training and prediction) per trial.}
\label{fig:nrmse_time}
\end{figure}

Figure~\subref*{fig:nrmse} summarizes the NRMSEPs of the $100$ DGP emulators produced by different approaches. We observe that DGPs trained by FB, followed by DSVI, give the best overall performance in terms of mean prediction accuracy. Although DGPs produced by SI present the least accurate mean predictions on average, their accuracy is clearly improved with SI-IC, approaching average NRMSEPs of FB and DSVI with moderate sacrifices of uncertainties (as shown in Figure~\ref{fig:sd}). For practicality, we compare in Figure~\subref*{fig:time} the single-core computation time (including training and prediction) taken by the packages (i.e., \texttt{deepgp}, \texttt{GPflux}, and \texttt{dgpsi}) that implement the four inference methods on a MacBook Pro with Apple M1 Max processor and $32$GB RAM. We note that SI-IC is generally faster than SI because ESS updates in SI have faster acceptances when the input connection is considered.

\section{Option Greeks from the Heston Model}
\label{sec:heston}
Option Greeks are important quantities used in financial engineering to measure the sensitivity of an option's price to features of the underlying asset such as the spot price or volatility. The Greeks are commonly used by financial engineers for risk hedging strategies and are essential elements of modern quantitative risk management. Popular Greeks include Vega that quantifies the sensitivity of an option's price to the volatility of the underlying asset, Delta that controls the sensitivity of an option's price to the underlying spot price, and Gamma that measures the sensitivity of an option's Delta to the underlying spot price. However, analytical calculations of Greeks are rarely available and one often needs to solve Partial Differential Equations (PDE) with numerical approaches, such as finite-difference methods or Monte-Carlo techniques, that could be computationally expensive~\citep{capriotti2017aad}, especially when fast evaluations of Greeks are desired under a large number of different option scenarios. Thus, building cheap-to-evaluate surrogates of Greeks is needed. 

\begin{figure}[!ht]
\centering 
\subfloat[Vega]{\label{fig:vega}\includegraphics[width=0.3\linewidth]{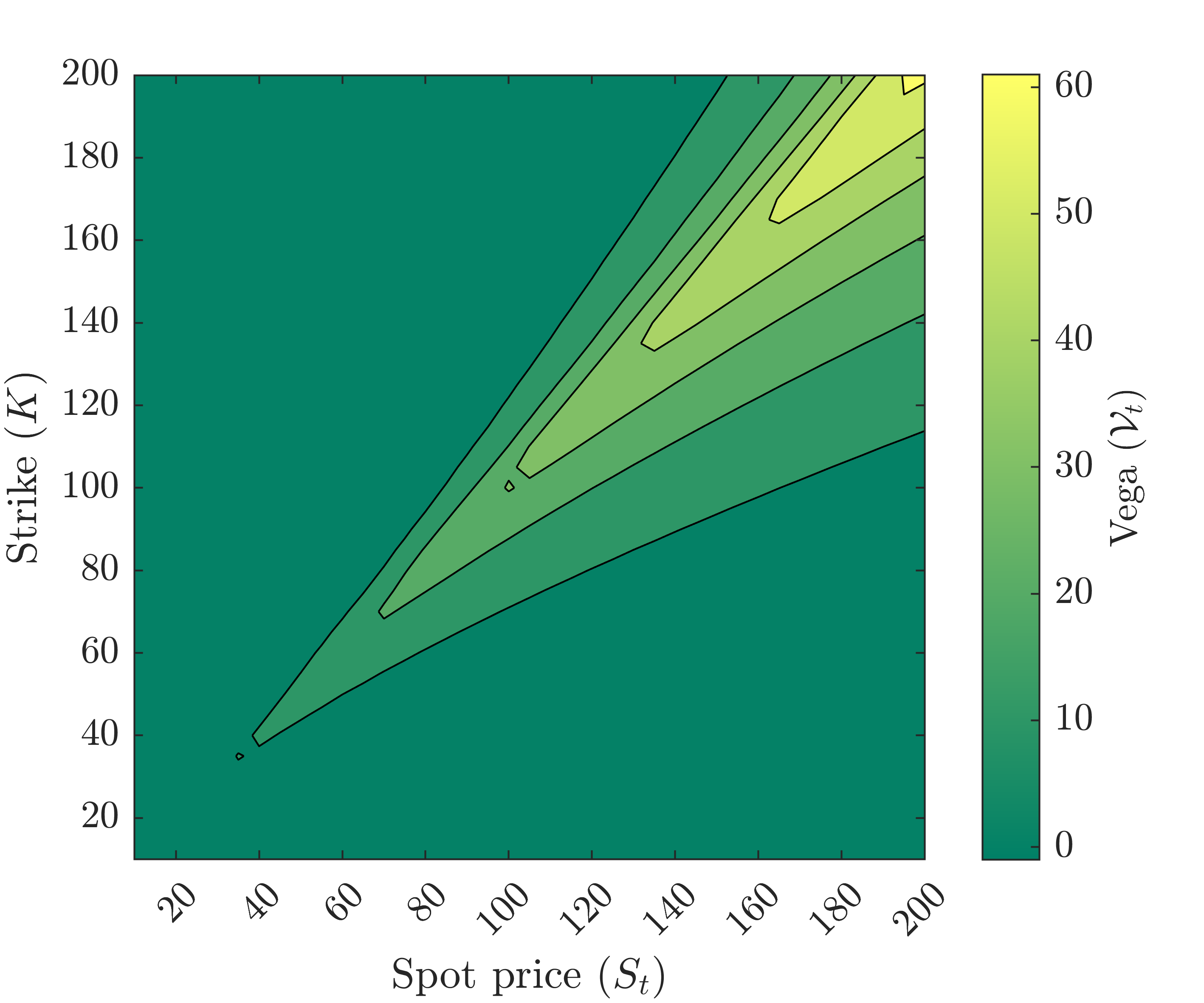}}\hspace{1em}
\subfloat[Delta]{\label{fig:delta}\includegraphics[width=0.3\linewidth]{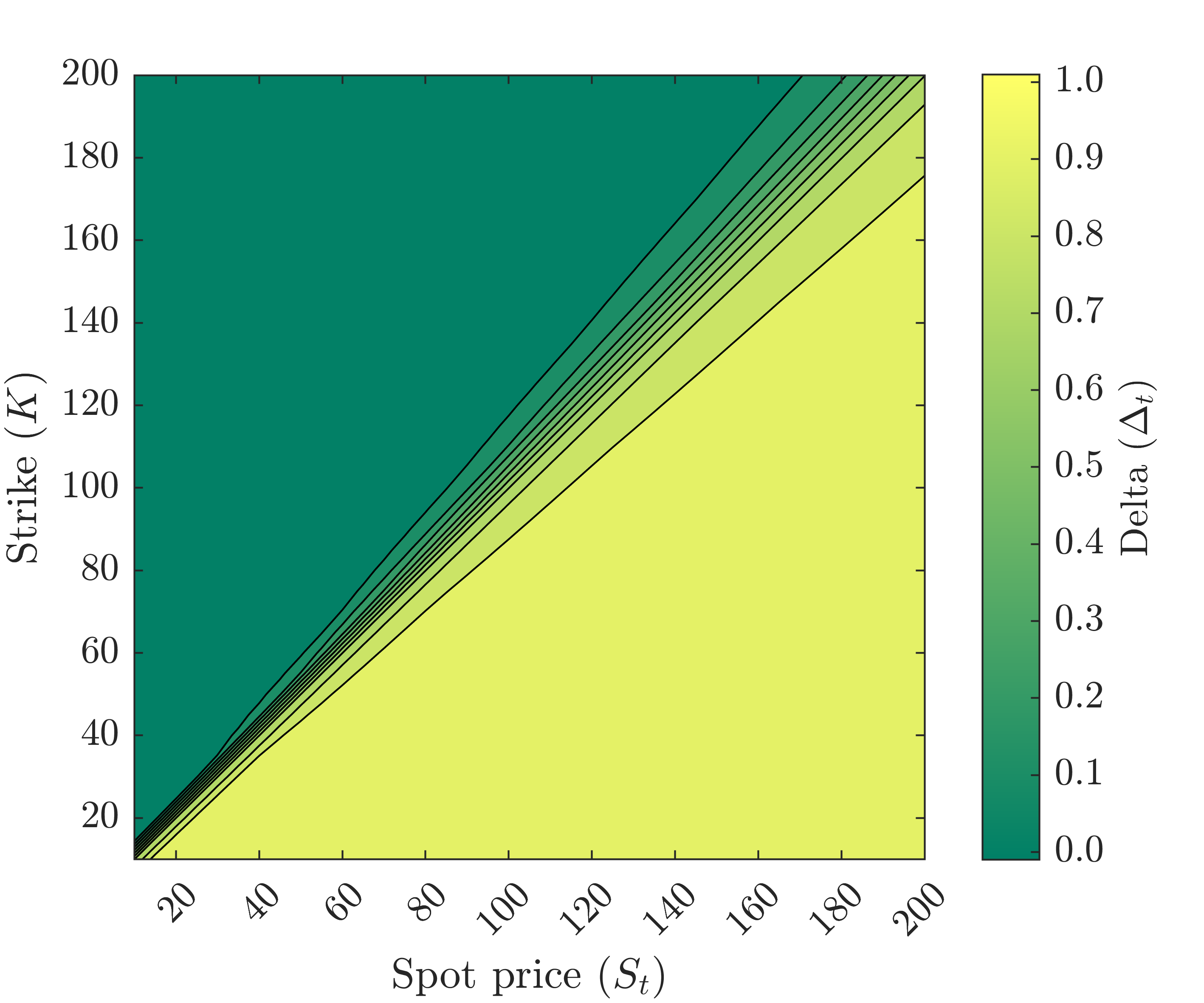}}\hspace{1em}
\subfloat[Gamma]{\label{fig:gamma}\includegraphics[width=0.3\linewidth]{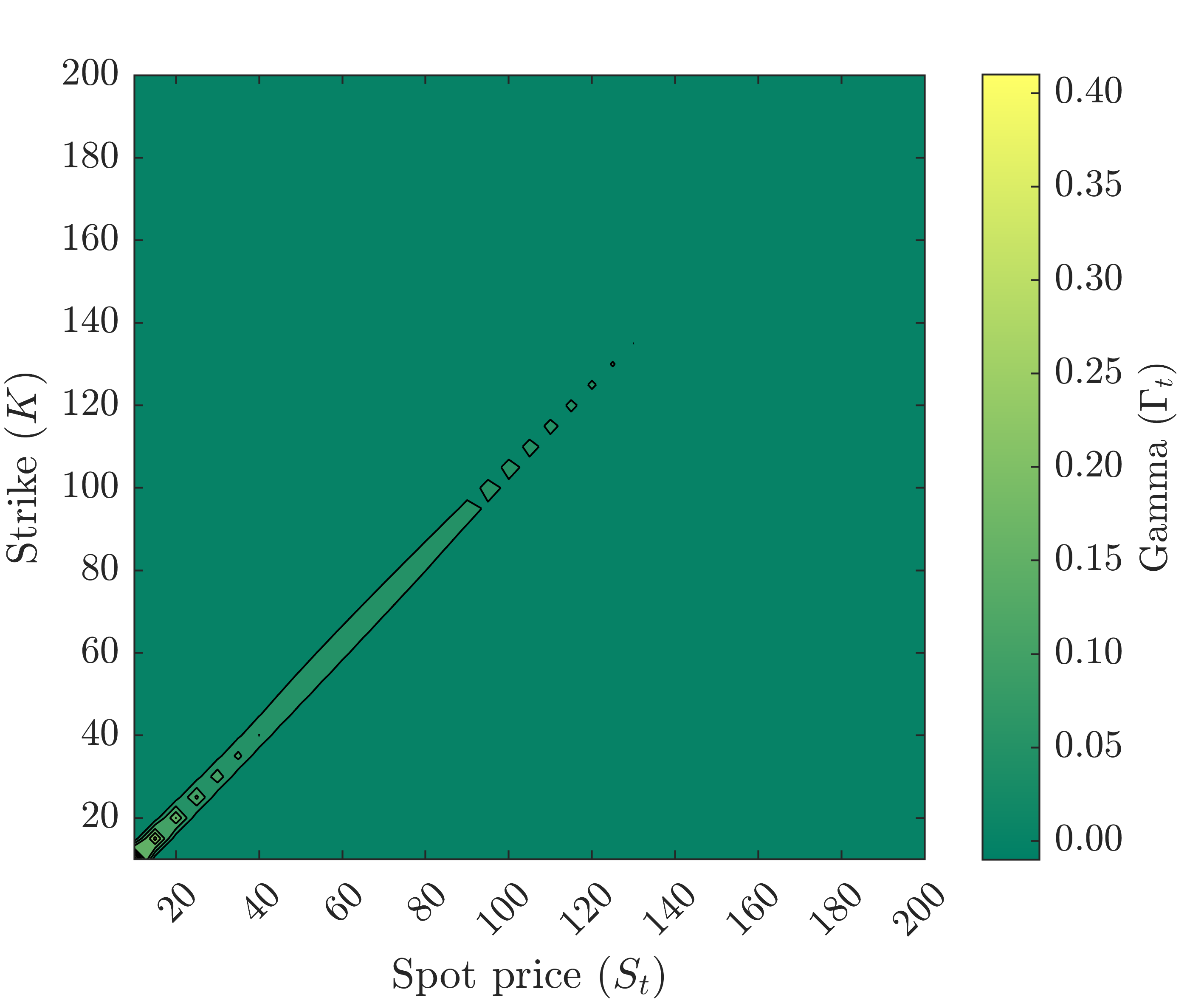}}
\caption{Contour plots of a slice of Vega, Delta and Gamma produced by~\eqref{eq:heston_pde} over $(S_t,K)\in[10,200]^2$ when $\tau=1$.}
\label{fig:heston_slice}
\end{figure}

Consider a European call option with strike price $K$ (in \$) and time-to-maturity $\tau$ (in years) whose price $C_t(S_t,K,\tau)$ at time $t$ depends on underlying asset price $S_t$ (in \$), which follows the Heston model~\citep{heston1993closed}:
\begin{align*}
dS_t&=(r-q)S_tdt+\sqrt{V_t}S_tdW^S_t\\
dV_t&=\kappa(\theta-V_t)dt+\sigma_V\sqrt{V_t}dW^V_t,
\end{align*}
where $r$ is the risk-free rate; $q$ is the dividend yield;  $V_t$ is the asset price variance with initial level $V_0=v_0$; $\kappa>0$ is the mean reversion rate of $V_t$; $\theta>0$ is the long-term variance; $\sigma_V>0$ is the volatility of $V_t$; $W^S_t$ and $W^V_t$ are Wiener processes with correlation $\rho$. Then, the computation of Greeks at time $t$ requires solving the Heston PDE given by~\citep{rouah2013heston}:
\begin{equation}
\label{eq:heston_pde}
    \frac{\partial C_t}{\partial t}+\frac{1}{2}S^2_tV_t\frac{\partial^2 C_t}{\partial S^2_t}+\rho S_t\sigma_VV_t\frac{\partial^2 C_t}{\partial S_t\partial V_t}+\frac{1}{2}\sigma^2_VV_t\frac{\partial^2 C_t}{\partial V^2_t}+(r-q)S_t\frac{\partial C_t}{\partial S_t}+\kappa(\theta-V_t)\frac{\partial C_t}{\partial V_t}=rC_t
\end{equation}
with the terminal condition $C_T=\max(0,S_T-K)$ at maturity $T$. Figure~\ref{fig:heston_slice} visualizes respectively a slice of Vega ($\mathcal{V}_t={\partial C_t}/{\partial v_0}$), Delta ($\Delta_t={\partial C_t}/{\partial S_t}$), and Gamma ($\Gamma_t={\partial^2C_t}/{\partial S^2_t}$) produced by~\eqref{eq:heston_pde} over $(S_t,K)\in[10,200]^2$ when $\tau$ is fixed to $1$. It can be seen that Vega, Delta and Gamma exhibit non-stationarity because at-the-money (ATM) options (i.e., options with strike prices close to the underlying asset prices) are most sensitive to asset price changes and oscillations, and thus cause a mountain to Vega, a cliff to Delta, and a spike to Gamma over the input domain. We compare SI and SI-IC to the other two inference approaches (i.e., DSVI and FB) for DGP emulation of the relationship between Greeks and $(S_t,K,\tau)$. In the remainder of this section, we focus on $\mathcal{V}_t$. Results for $\Delta_t$ and $\Gamma_t$ are given in Section~\ref{sec:delta} and~\ref{sec:gamma} of the supplement.

\begin{figure}[!ht]
\centering 
\begin{minipage}[b]{.3\linewidth}
\centering
\subfloat[2-layer]{
\scalebox{0.7}{
\begin{tikzpicture}[shorten >=1pt,->,draw=black!50, node distance=4cm]
    \tikzstyle{every pin edge}=[<-,shorten <=1pt]
    \tikzstyle{neuron}=[circle,fill=black!25,minimum size=27.5pt,inner sep=0pt]
    \tikzstyle{layer1}=[neuron, fill=green!50];
    \tikzstyle{layer2}=[neuron, fill=red!50];
    \tikzstyle{layer3}=[neuron, fill=blue!50];
    \node[layer1, pin=left:] (l1-0) at (0,0) {$\mathcal{GP}^{(1)}_{1}$};
    \node[layer1, pin=left:] (l1-1) at (0,-1.25) {$\mathcal{GP}^{(2)}_{1}$};
    \node[layer1, pin=left:] (l1-2) at (0,-2.5) {$\mathcal{GP}^{(3)}_{1}$};
    \node[layer2,pin={[pin edge={->}]right:}] (l2-0) at (1.5,-1.25) {$\mathcal{GP}^{(1)}_{2}$};
\draw[->] (l1-0) -- (l2-0);
\draw[->] (l1-1) -- (l2-0);
\draw[->] (l1-2) -- (l2-0);
\end{tikzpicture}}}
\end{minipage}
\begin{minipage}[b]{.3\linewidth}
\centering
\subfloat[3-layer]{
\scalebox{0.7}{
\begin{tikzpicture}[shorten >=1pt,->,draw=black!50, node distance=4cm]
    \tikzstyle{every pin edge}=[<-,shorten <=1pt]
    \tikzstyle{neuron}=[circle,fill=black!25,minimum size=27.5pt,inner sep=0pt]
    \tikzstyle{layer1}=[neuron, fill=green!50];
    \tikzstyle{layer2}=[neuron, fill=red!50];
    \tikzstyle{layer3}=[neuron, fill=blue!50];
    \node[layer1, pin=left:] (l1-0) at (0,0) {$\mathcal{GP}^{(1)}_{1}$};
    \node[layer1, pin=left:] (l1-1) at (0,-1.25) {$\mathcal{GP}^{(2)}_{1}$};
    \node[layer1, pin=left:] (l1-2) at (0,-2.5) {$\mathcal{GP}^{(3)}_{1}$};
    \node[layer2] (l2-0) at (1.5,0) {$\mathcal{GP}^{(1)}_{2}$};
    \node[layer2] (l2-1) at (1.5,-1.25) {$\mathcal{GP}^{(2)}_{2}$};
    \node[layer2] (l2-2) at (1.5,-2.5) {$\mathcal{GP}^{(3)}_{2}$};
    \node[layer3,pin={[pin edge={->}]right:}] (l3-0) at (3,-1.25) {$\mathcal{GP}^{(1)}_{3}$};
\draw[->] (l1-0) -- (l2-0);
\draw[->] (l1-0) -- (l2-1);
\draw[->] (l1-0) -- (l2-2);
\draw[->] (l1-1) -- (l2-0);
\draw[->] (l1-1) -- (l2-1);
\draw[->] (l1-1) -- (l2-2);
\draw[->] (l1-2) -- (l2-0);
\draw[->] (l1-2) -- (l2-1);
\draw[->] (l1-2) -- (l2-2);
\draw[->] (l2-0) -- (l3-0);
\draw[->] (l2-1) -- (l3-0);
\draw[->] (l2-2) -- (l3-0);
\end{tikzpicture}}}
\end{minipage}
\begin{minipage}[b]{.35\linewidth}
\centering
\subfloat[4-layer]{
\scalebox{0.7}{
\begin{tikzpicture}[shorten >=1pt,->,draw=black!50, node distance=4cm]
    \tikzstyle{every pin edge}=[<-,shorten <=1pt]
    \tikzstyle{neuron}=[circle,fill=black!25,minimum size=27.5pt,inner sep=0pt]
    \tikzstyle{layer1}=[neuron, fill=green!50];
    \tikzstyle{layer2}=[neuron, fill=red!50];
    \tikzstyle{layer3}=[neuron, fill=blue!50];
    \tikzstyle{layer4}=[neuron, fill=purple!50];
    \node[layer1, pin=left:] (l1-0) at (0,0) {$\mathcal{GP}^{(1)}_{1}$};
    \node[layer1, pin=left:] (l1-1) at (0,-1.25) {$\mathcal{GP}^{(2)}_{1}$};
    \node[layer1, pin=left:] (l1-2) at (0,-2.5) {$\mathcal{GP}^{(3)}_{1}$};
    \node[layer2] (l2-0) at (1.5,0) {$\mathcal{GP}^{(1)}_{2}$};
    \node[layer2] (l2-1) at (1.5,-1.25) {$\mathcal{GP}^{(2)}_{2}$};
    \node[layer2] (l2-2) at (1.5,-2.5) {$\mathcal{GP}^{(3)}_{2}$};
    \node[layer3] (l3-0) at (3,0) {$\mathcal{GP}^{(1)}_{3}$};
    \node[layer3] (l3-1) at (3,-1.25) {$\mathcal{GP}^{(2)}_{3}$};
    \node[layer3] (l3-2) at (3,-2.5) {$\mathcal{GP}^{(3)}_{3}$};
    \node[layer4,pin={[pin edge={->}]right:}] (l4-0) at (4.5,-1.25) {$\mathcal{GP}^{(1)}_{4}$};
\draw[->] (l1-0) -- (l2-0);
\draw[->] (l1-0) -- (l2-1);
\draw[->] (l1-0) -- (l2-2);
\draw[->] (l1-1) -- (l2-0);
\draw[->] (l1-1) -- (l2-1);
\draw[->] (l1-1) -- (l2-2);
\draw[->] (l1-2) -- (l2-0);
\draw[->] (l1-2) -- (l2-1);
\draw[->] (l1-2) -- (l2-2);
\draw[->] (l2-0) -- (l3-0);
\draw[->] (l2-0) -- (l3-1);
\draw[->] (l2-0) -- (l3-2);
\draw[->] (l2-1) -- (l3-0);
\draw[->] (l2-1) -- (l3-1);
\draw[->] (l2-1) -- (l3-2);
\draw[->] (l2-2) -- (l3-0);
\draw[->] (l2-2) -- (l3-1);
\draw[->] (l2-2) -- (l3-2);
\draw[->] (l3-0) -- (l4-0);
\draw[->] (l3-1) -- (l4-0);
\draw[->] (l3-2) -- (l4-0);
\end{tikzpicture}}}
\end{minipage}
\caption{Three different DGP formations considered to build the emulator of Vega.}
\label{fig:archs}
\end{figure}
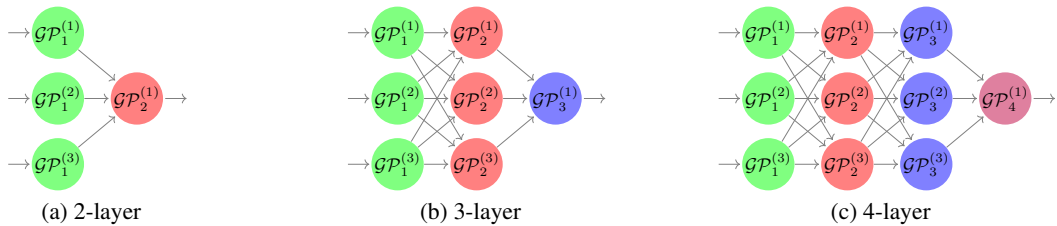

To train DGP emulators of Vega, we generate $100$ training data points by first drawing $100$ input positions over $(S_t,K,\tau)\in[10,200]\times[10,200]\times[1/12,3]$ with Latin-hypercube-sampler (LHS), and then compute numerically the corresponding $\mathcal{V}_t$ from the Heston model using the \texttt{Financial Instruments Toolbox} of \texttt{MATLAB}. $500$ testing data points are obtained in the same fashion. The model parameters $(r,q,v_0,\kappa,\theta,\sigma_V,\rho)$ in~\eqref{eq:heston_pde} are set to $(0.03, 0.02, 0.04, 0.04, 0.3, 0.9, -0.5)$, following~\citet{teng2018numerical}. We adopt three formations (in which each individual GP has its one-dimensional kernel functions across different input dimensions sharing a common range parameter) shown in Figure~\ref{fig:archs} for DGP emulators. For each combination of formation and inference approach we conduct $40$ inference trials. However, only DSVI, SI, and SI-IC are implemented for the four-layer formation because \texttt{deepgp} only allows DGP hierarchies up to three layers. 

\subsection{Results}
\begin{figure}[!ht]
\centering 
\subfloat[NRMSEP]{\label{fig:vega_rmse}\includegraphics[width=0.45\linewidth]{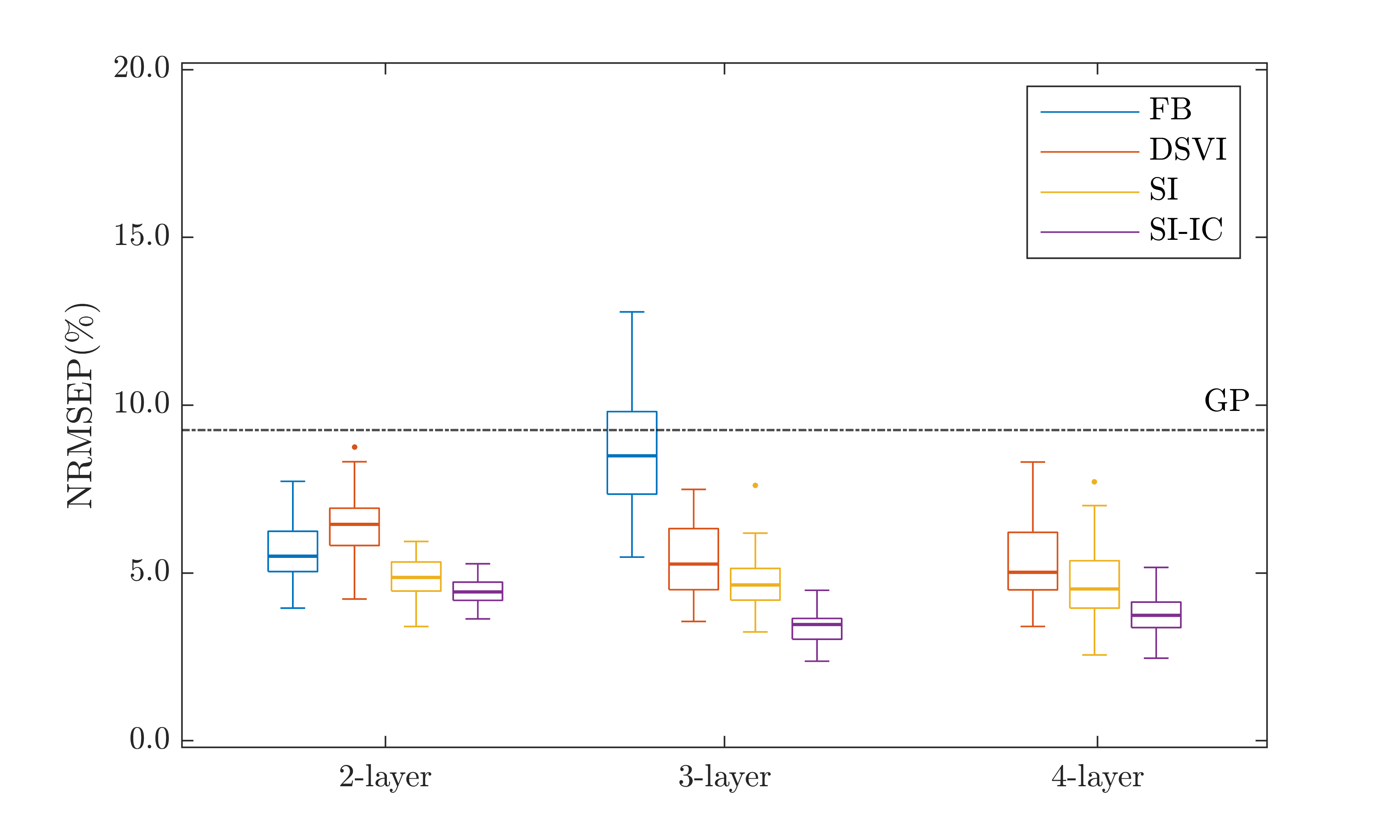}}
\subfloat[Computation time]{\label{fig:vega_time}\includegraphics[width=0.45\linewidth]{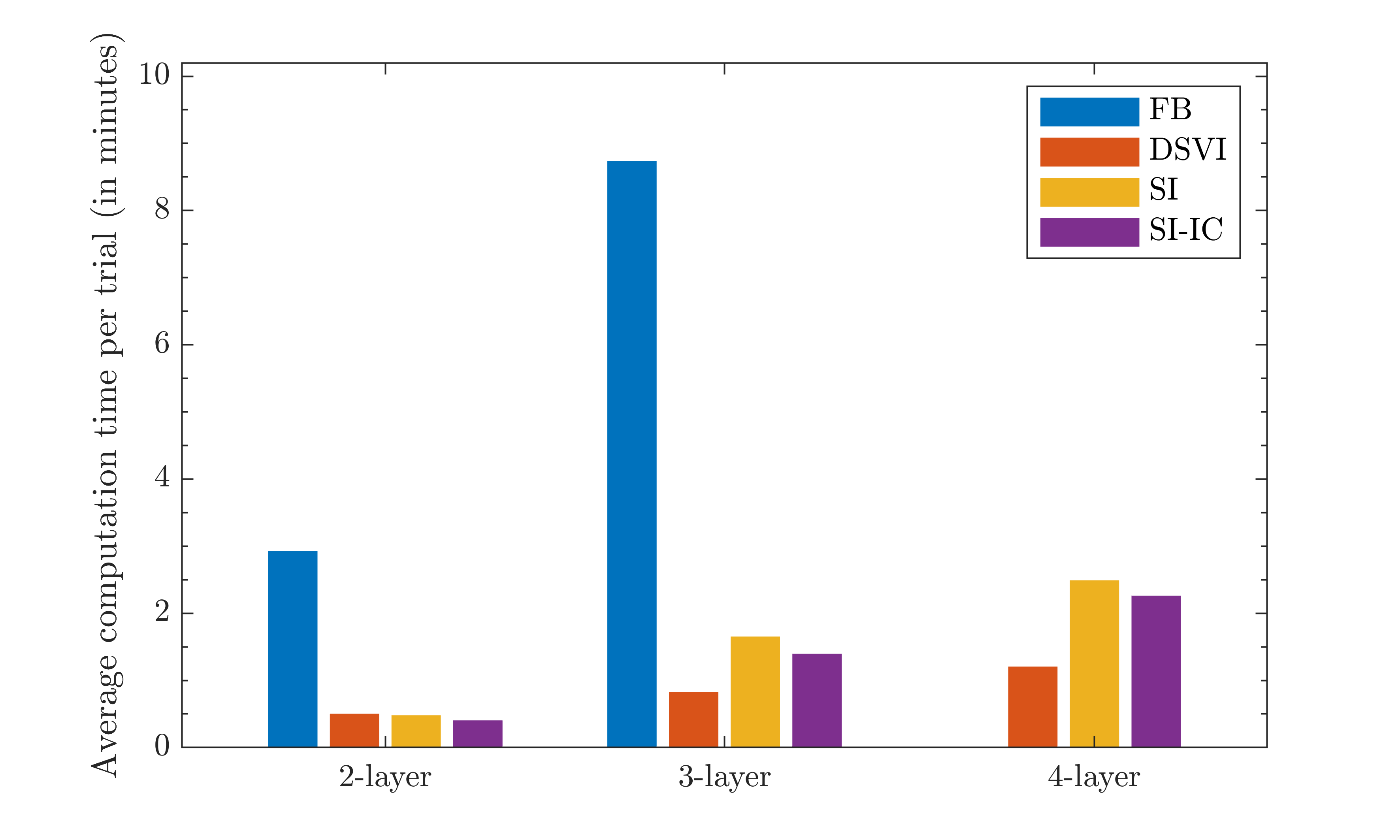}} 
\caption{Comparison of FB, DSVI, SI, and SI-IC for $40$ repeatedly trained DGP emulators (i.e., $40$ inference trials) of Vega ($\mathcal{V}_t$) from the Heston model. FB is not implemented for the 4-layer formation because \texttt{deepgp} only allows DGPs up to three layers. The dash-dot line represents the NRMSEP of a trained conventional GP emulator.}
\label{fig:vega_rmse_time}
\end{figure}

It can be seen from Figure~\subref*{fig:vega_rmse} that emulators produced by SI outperform those trained by FB and DSVI under all experimental settings. Figure~\subref*{fig:vega_rmse} also shows that with the input connection, SI could produce DGP emulators with even lower NRMSEPs. In addition, we observe that under DSVI and SI-IC, three-layered DGP emulators have systemically lower NRMSEPs than two-layered emulators. However, for both DSVI and SI-IC increasing DGP depth to four layers shows no improvement on NRMSEP. 

\begin{figure}[!ht]
\centering 
\subfloat[2-layer (FB)]{\label{fig:vega_fb2}\includegraphics[width=0.25\linewidth]{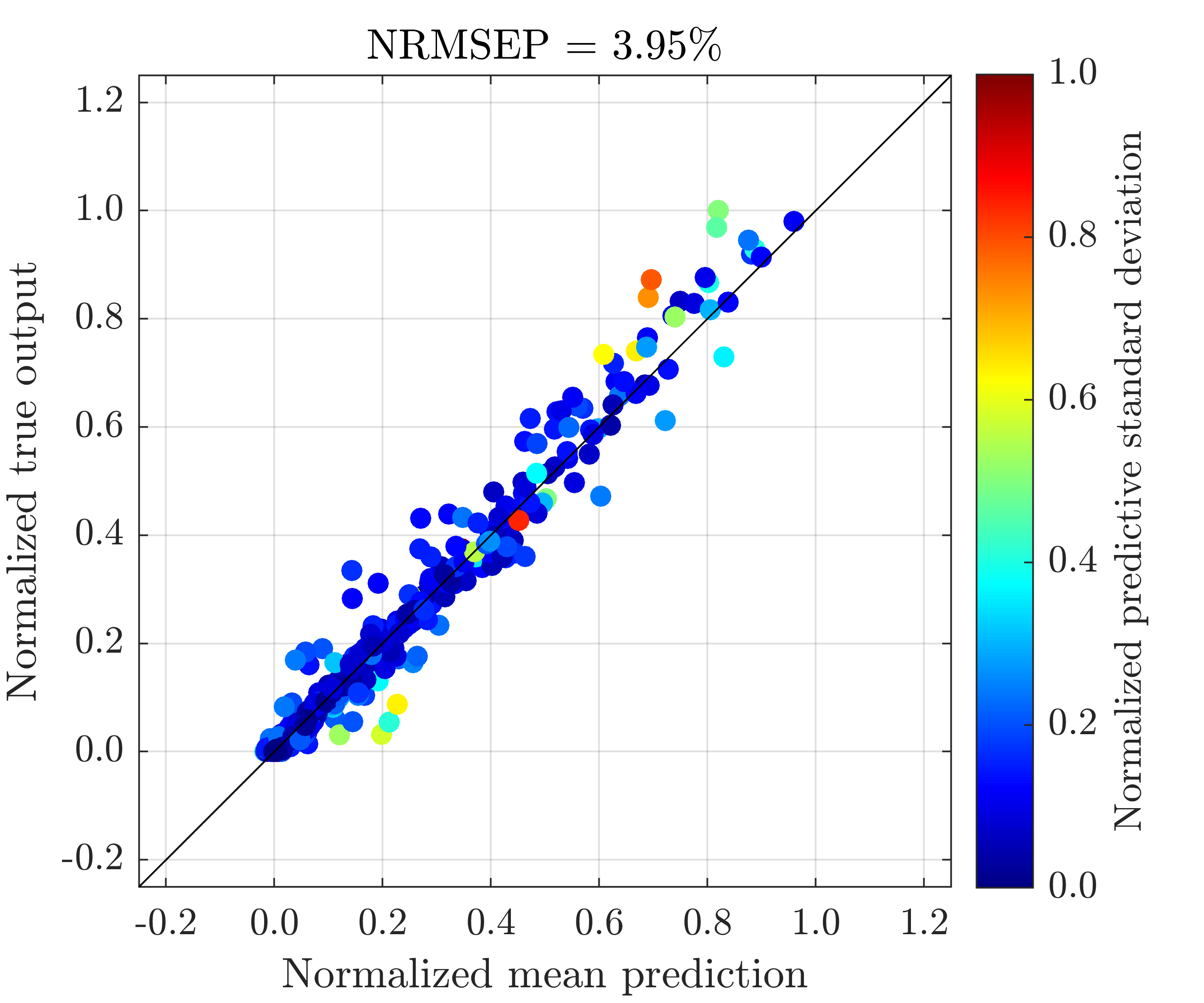}}
\subfloat[2-layer (DSVI)]{\label{fig:vega_vi2}\includegraphics[width=0.25\linewidth]{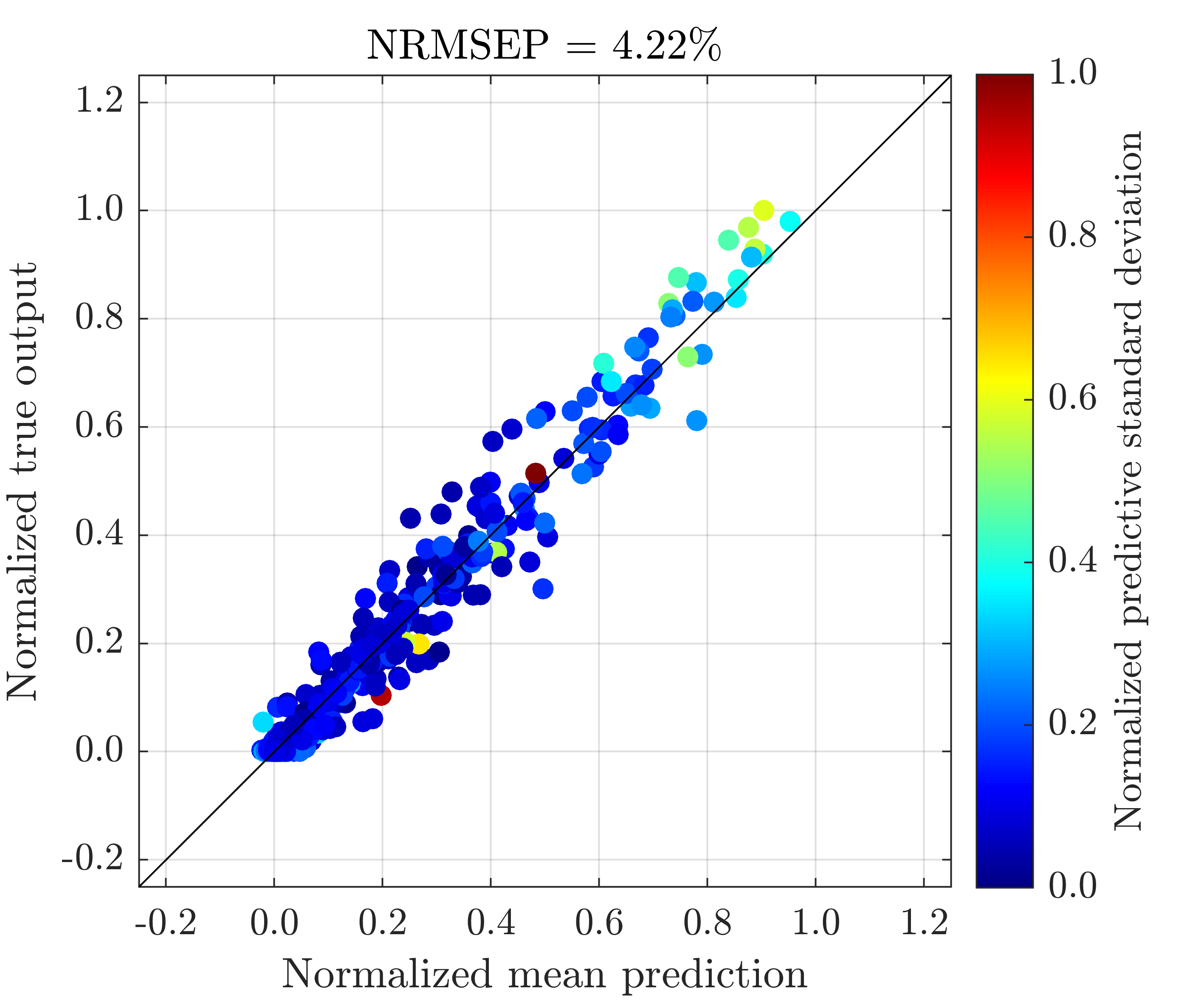}}
\subfloat[2-layer (SI)]{\label{fig:vega_si2}\includegraphics[width=0.25\linewidth]{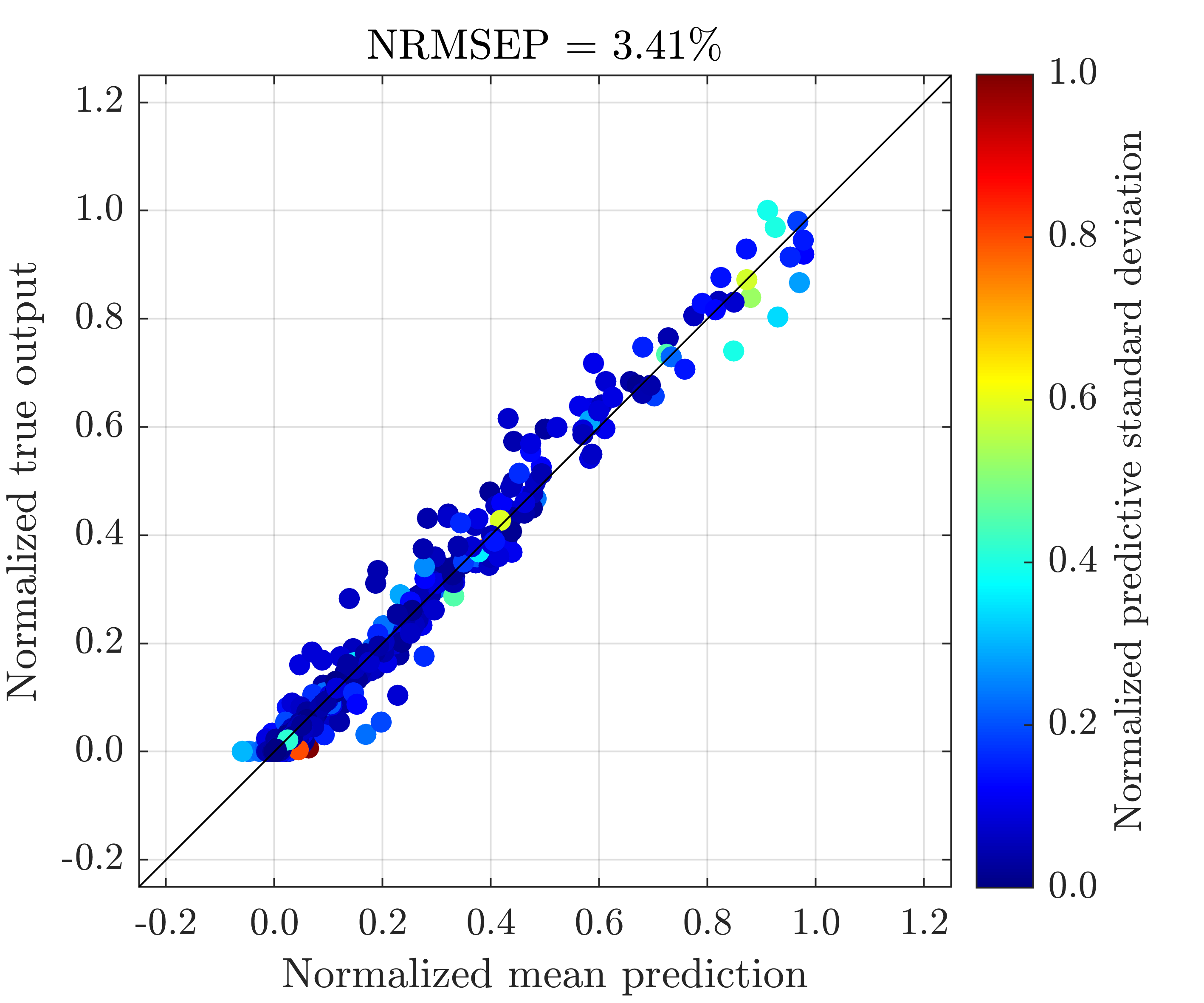}}
\subfloat[2-layer (SI-IC)]{\label{fig:vega_si2_con}\includegraphics[width=0.25\linewidth]{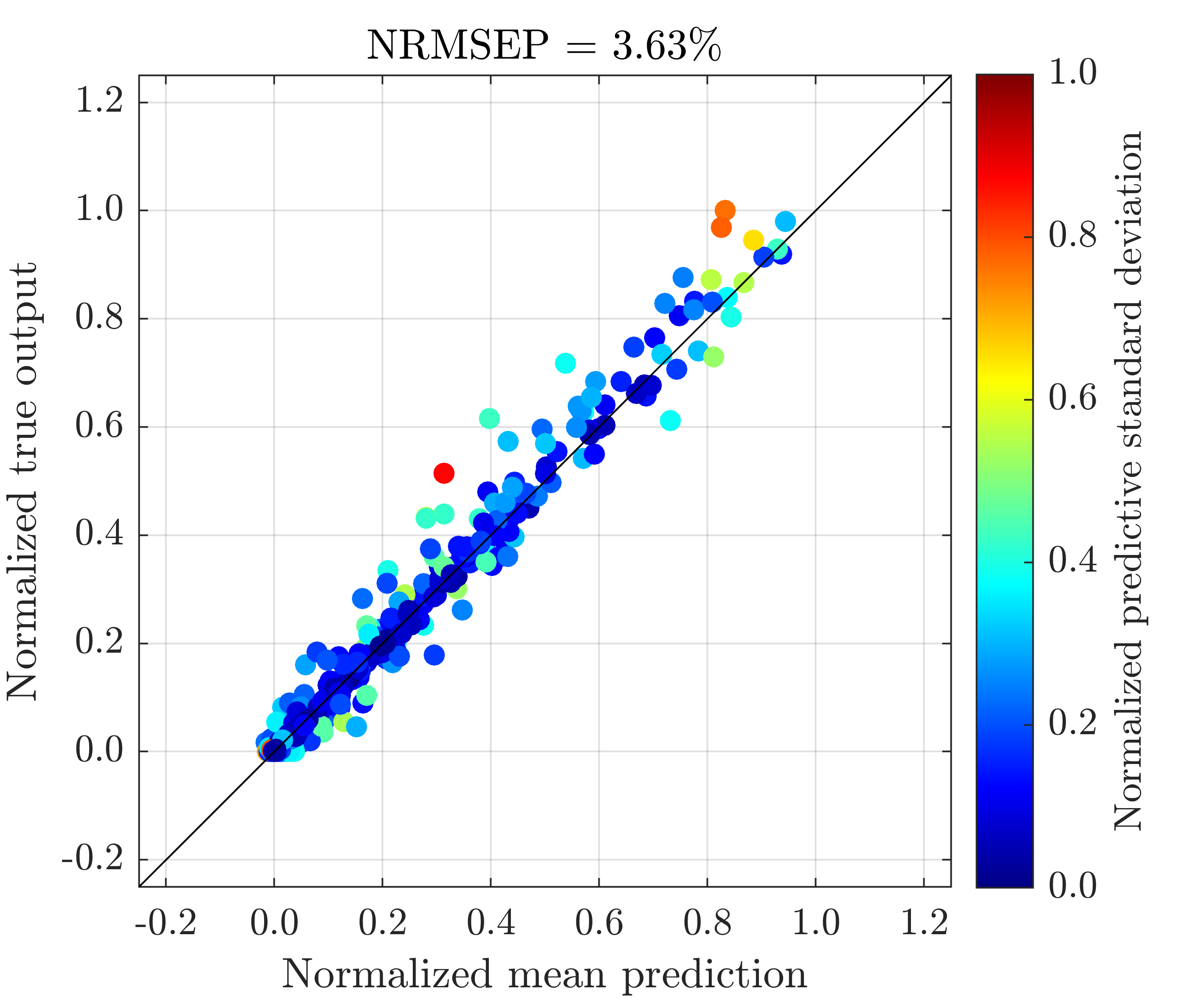}}\\
\subfloat[3-layer (FB)]{\label{fig:vega_fb3}\includegraphics[width=0.25\linewidth]{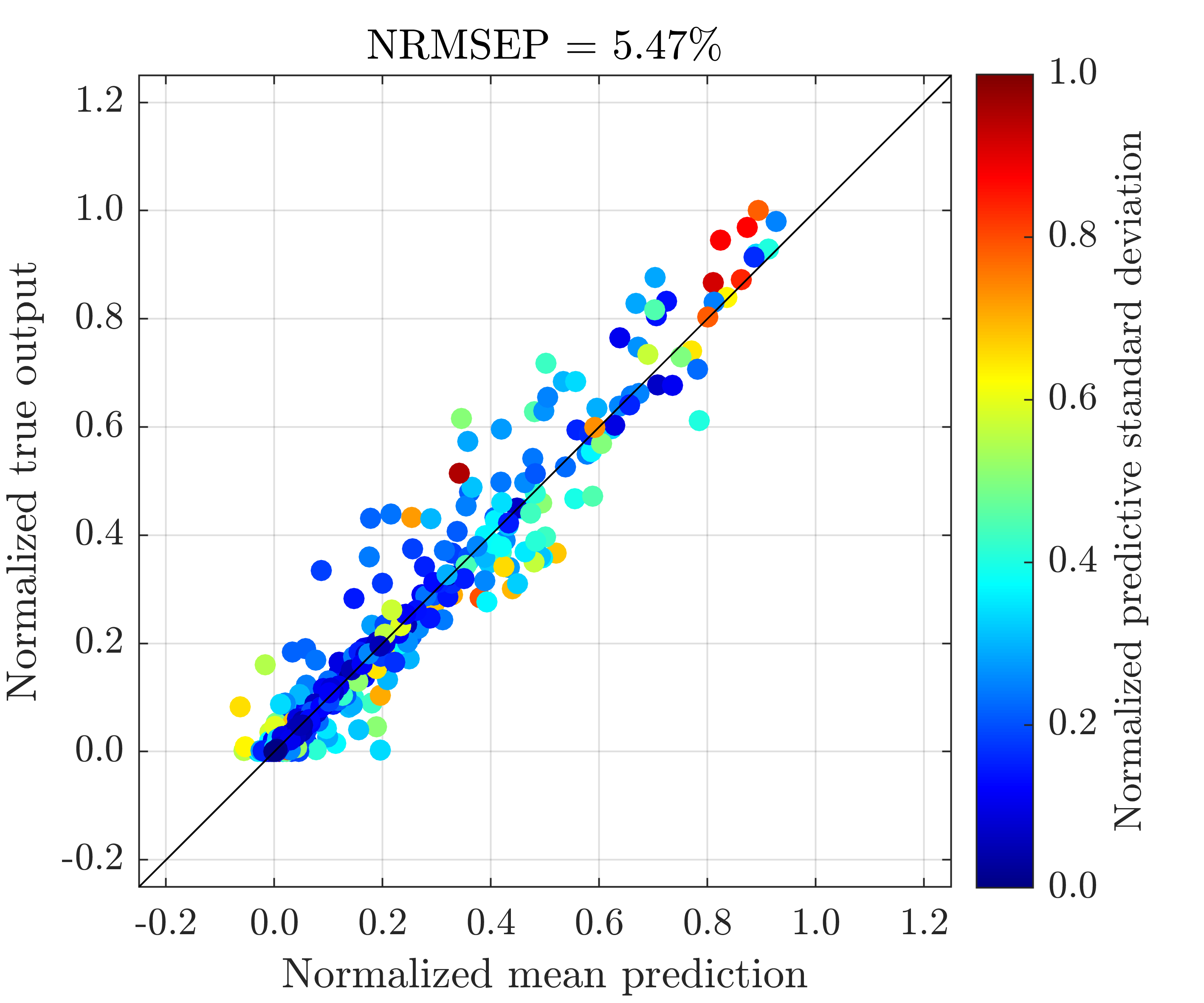}}
\subfloat[3-layer (DSVI)]{\label{fig:vega_vi3}\includegraphics[width=0.25\linewidth]{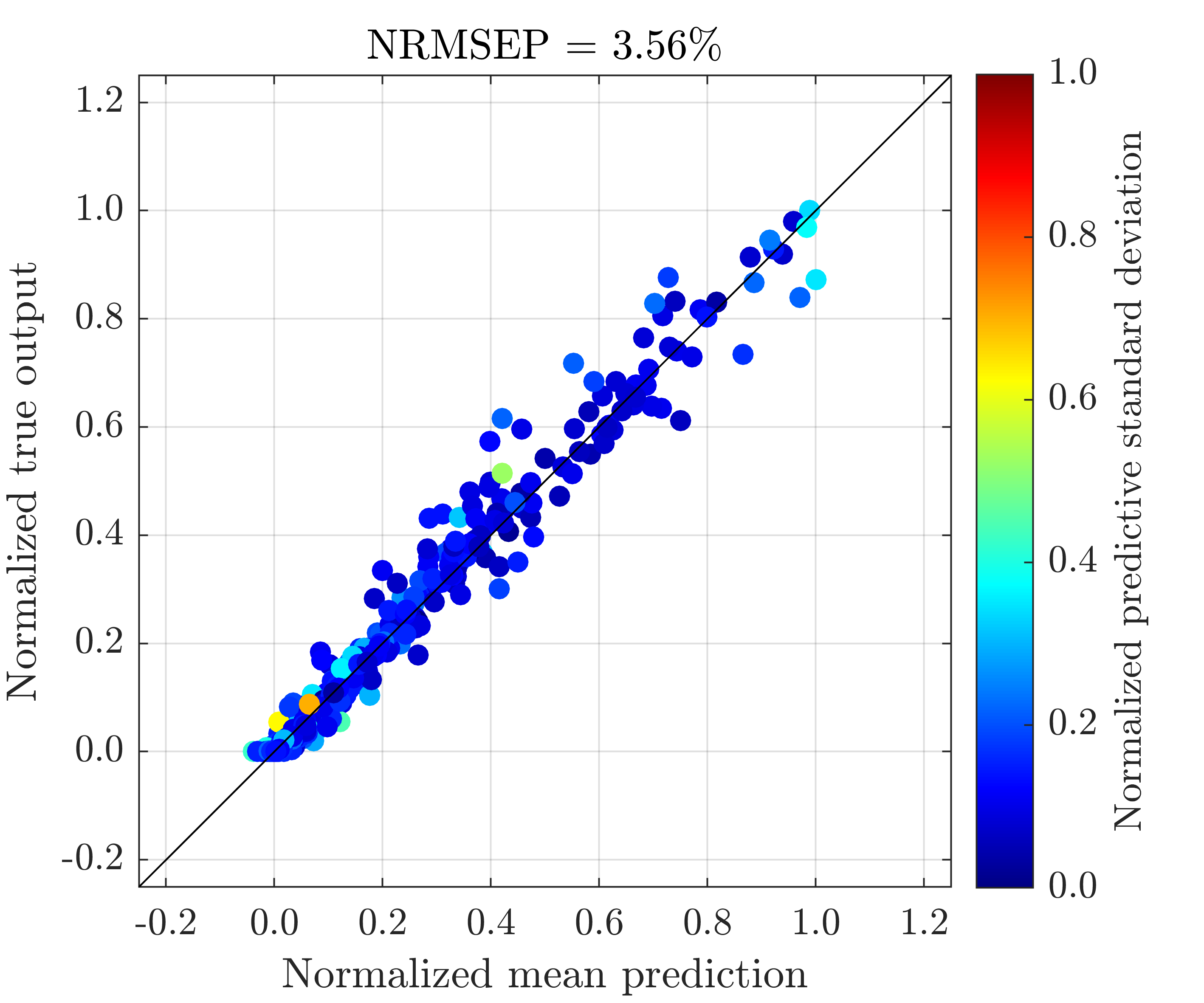}}
\subfloat[3-layer (SI)]{\label{fig:vega_si3}\includegraphics[width=0.25\linewidth]{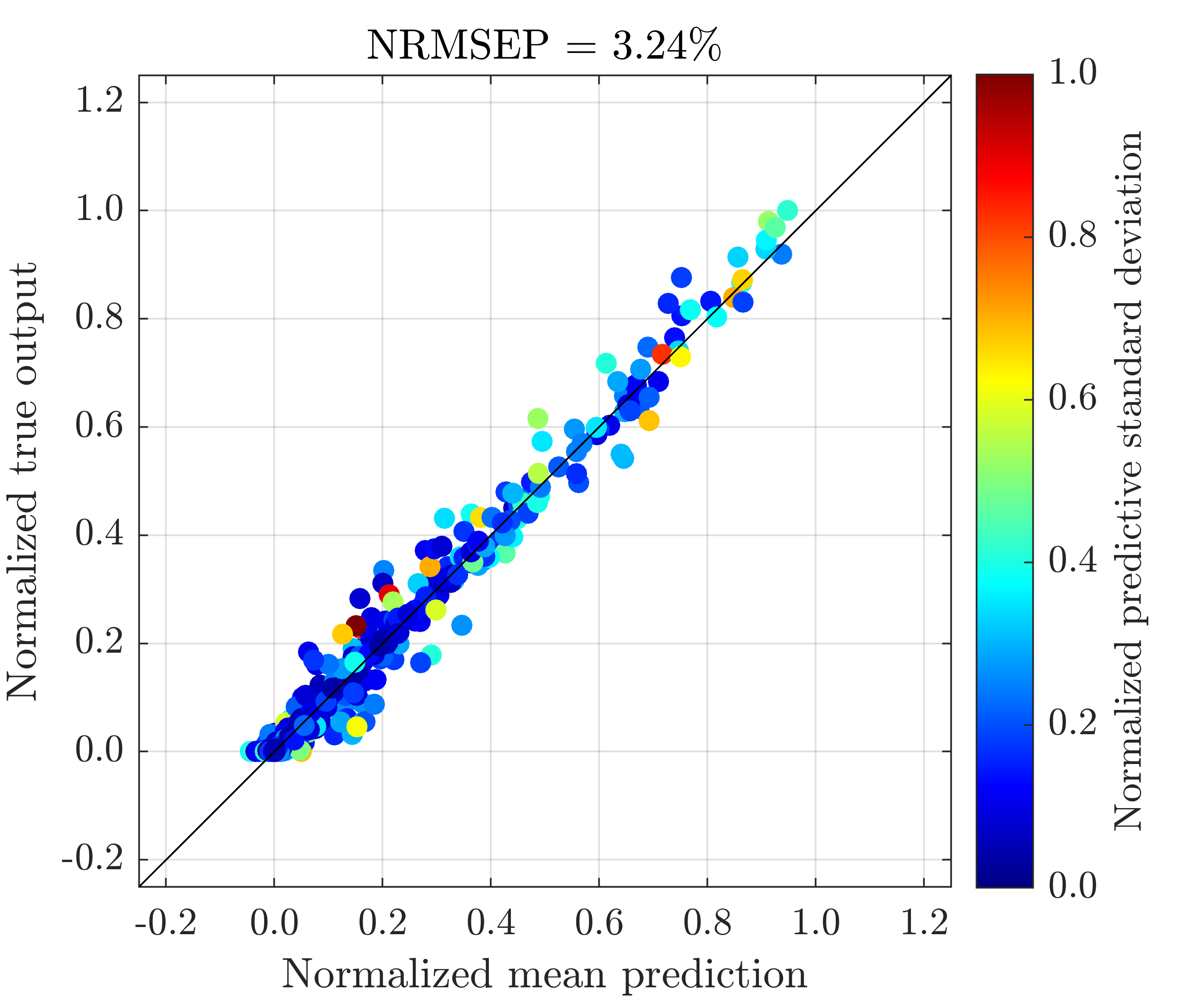}}
\subfloat[3-layer (SI-IC)]{\label{fig:vega_si3_con}\includegraphics[width=0.25\linewidth]{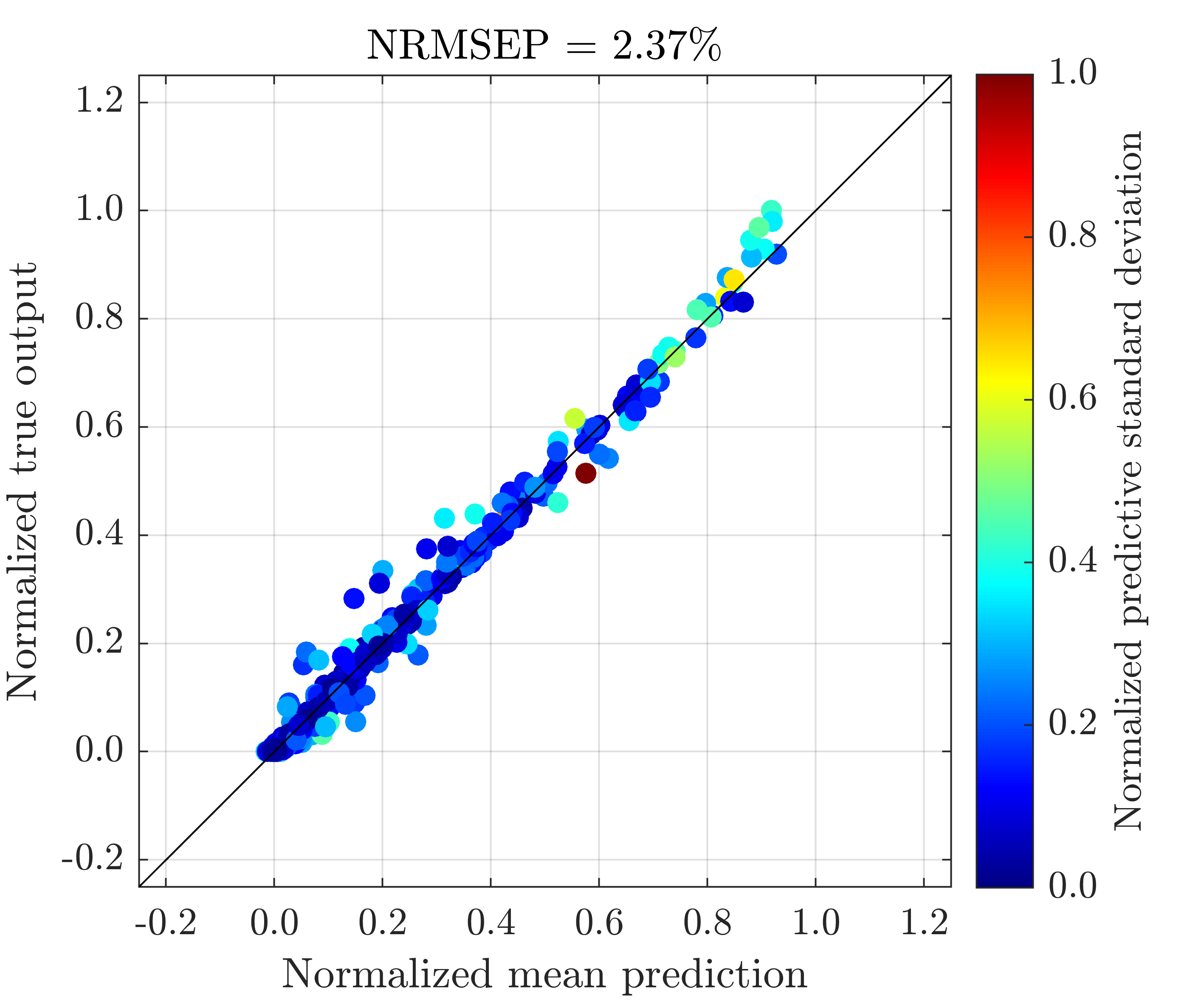}}\\
\subfloat[GP]{\label{fig:vega_gp}\includegraphics[width=0.25\linewidth]{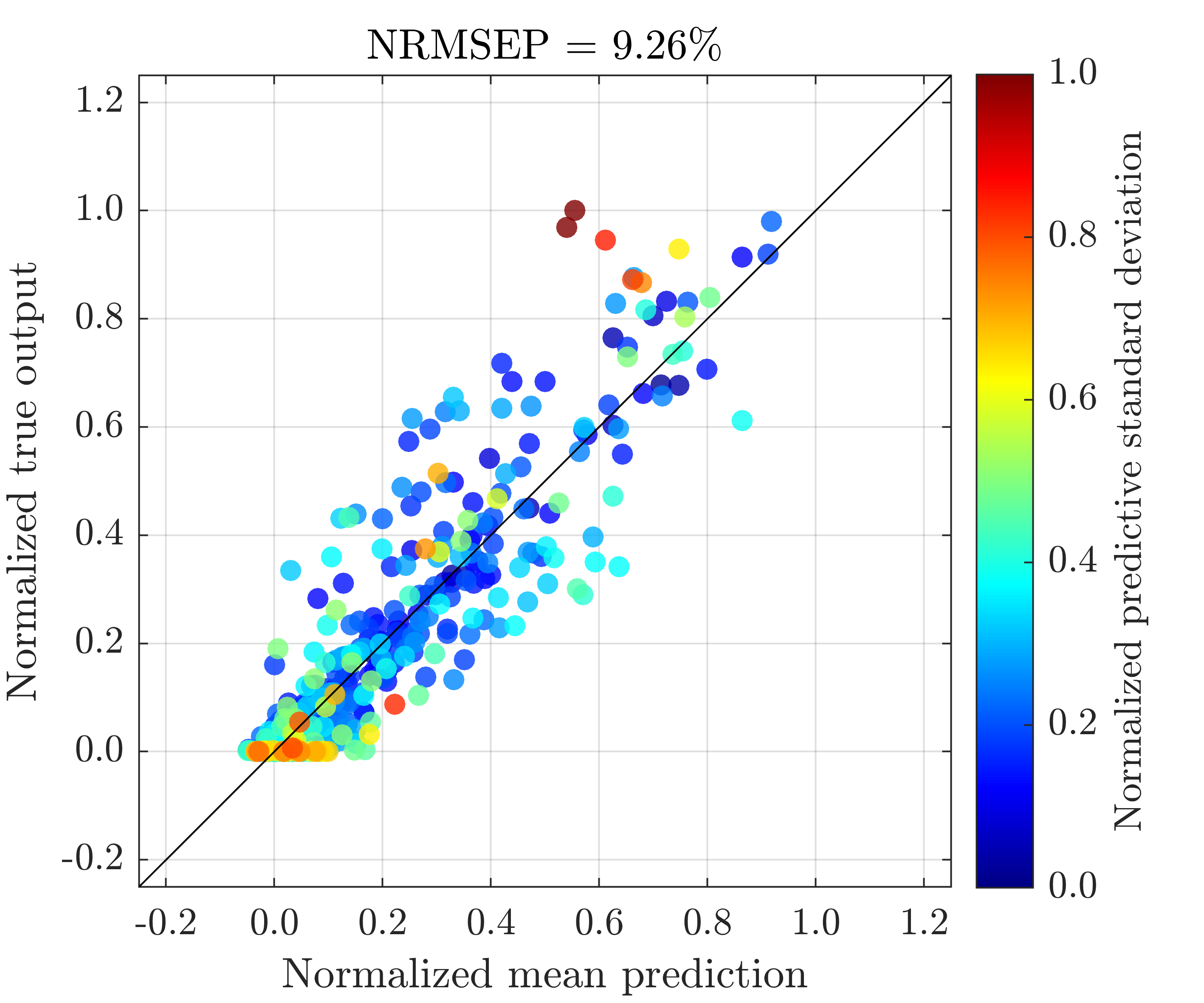}} 
\subfloat[4-layer (DSVI)]{\label{fig:vega_vi4}\includegraphics[width=0.25\linewidth]{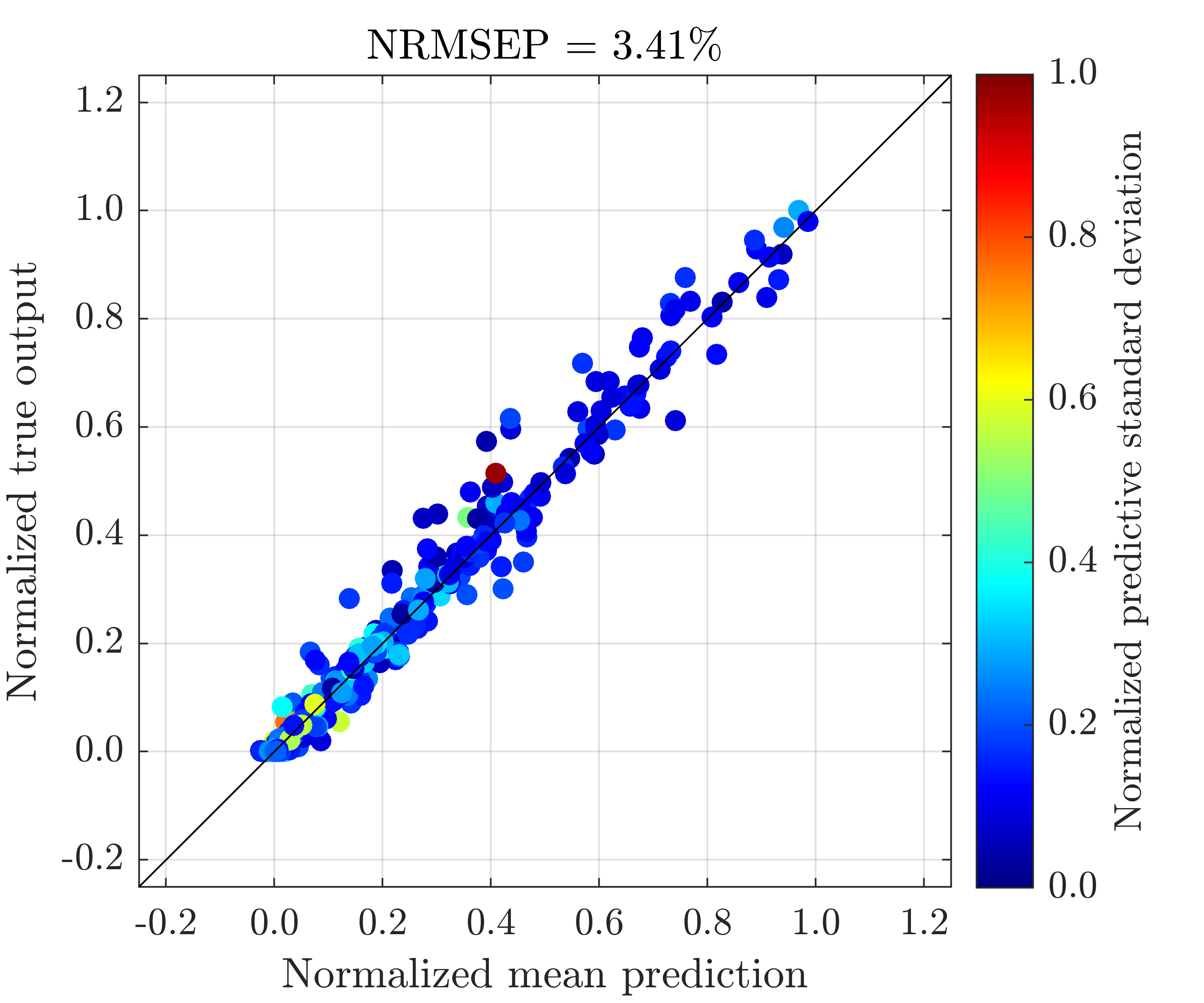}}
\subfloat[4-layer (SI)]{\label{fig:vega_si4}\includegraphics[width=0.25\linewidth]{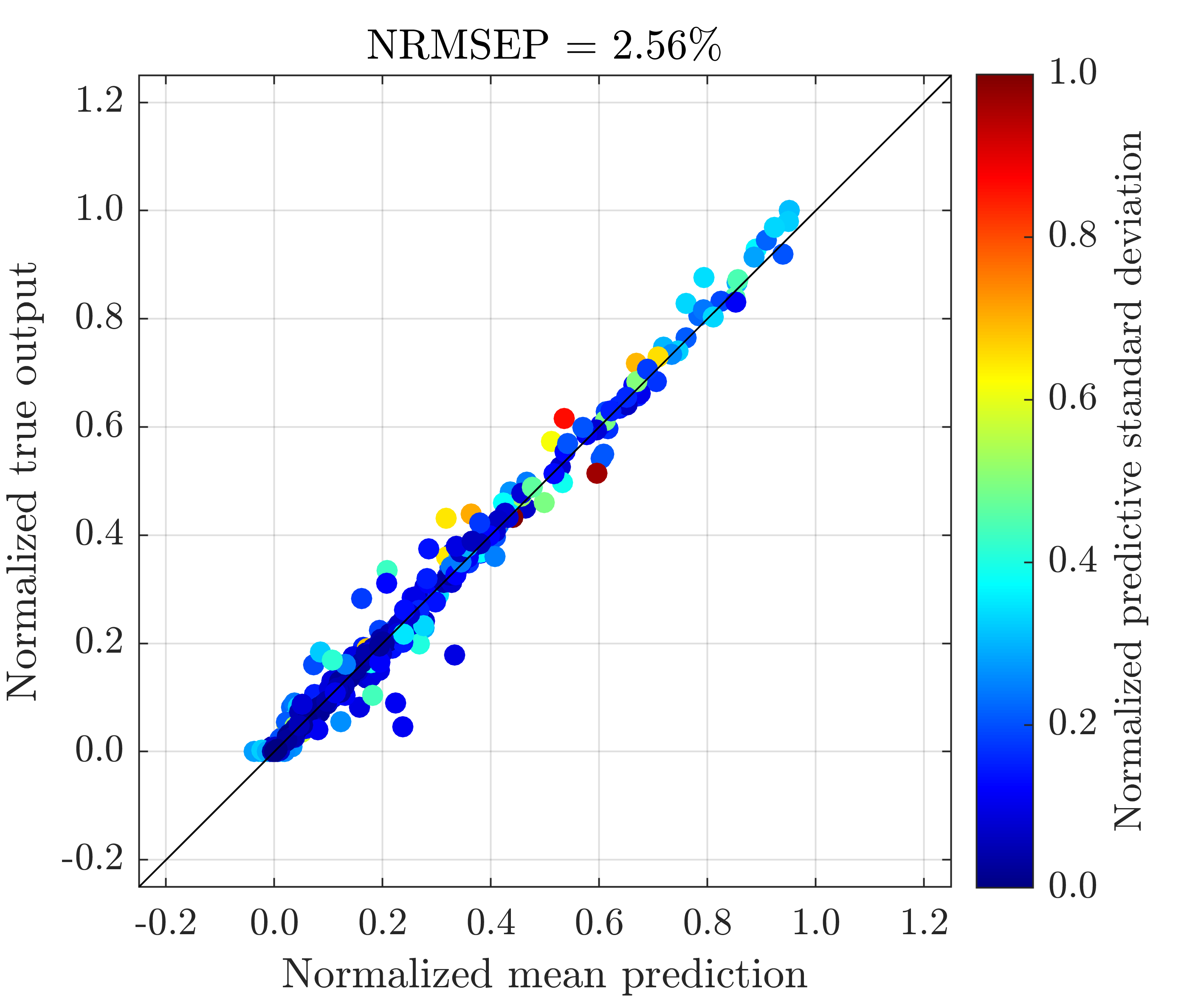}}
\subfloat[4-layer (SI-IC)]{\label{fig:vega_si4_con}\includegraphics[width=0.25\linewidth]{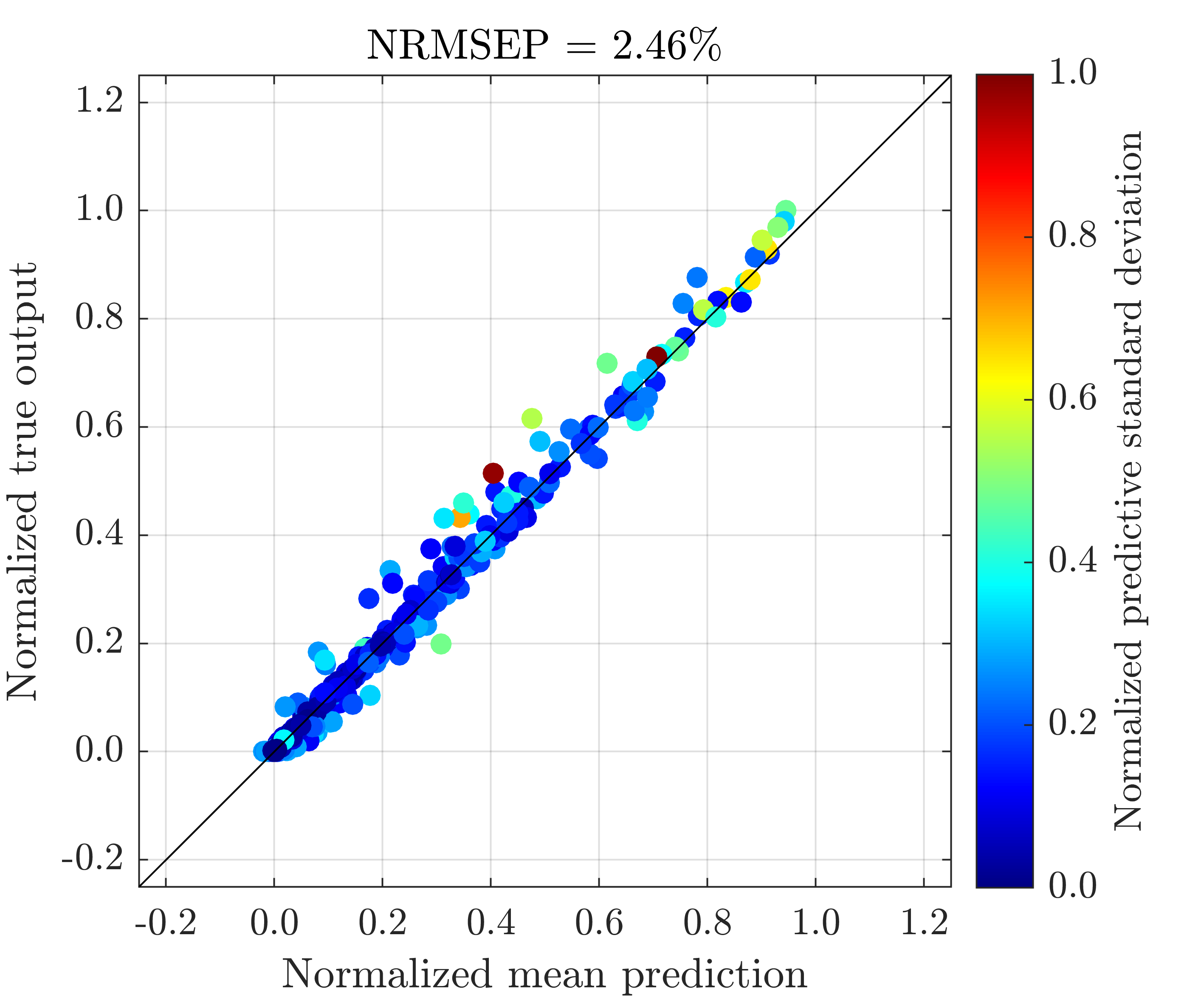}}
\caption{Plots of numerical solutions of Vega ($\mathcal{V}_t$) (normalized by their max and min values) from the Heston model at $500$ testing positions \emph{vs} the mean predictions (normalized by the max and min values of numerical solutions of Vega), along with predictive standard deviations (normalized by their max and min values), made by the best emulator (with the lowest NRMSEP out of $40$ inference trials) produced by FB, DSVI, SI and SI-IC. \emph{GP} represents a conventional GP emulator.}
\label{fig:vega_diag}
\end{figure}

Figure~\ref{fig:vega_diag} presents the profiles of uncertainty quantified by best DGP emulators, which are trained by different methods, from $40$ inference trials. The profiles show similar uncertainty behaviors of DGPs to those in Section~\ref{sec:step_fct}. In comparison to FB and SI (with or without the input connection), DSVI produces DGPs with lower uncertainties at locations where mean predictions are poor (e.g., dots in Figure~\subref*{fig:vega_vi2},~\subref*{fig:vega_vi3}, and~\subref*{fig:vega_vi4} that deviate from the diagonal lines have low predictive standard deviations) and in regions where Vega value becomes larger and exhibits more variations, across different formations. This can be problematic for tasks such as active learning in which DGP emulators trained by DSVI could unnecessarily evaluate the Heston PDE over input space where the DGP predictions are well-behaved. Although DGPs (e.g., the three-layered one in Figure~\subref*{fig:vega_fb3}) from FB provide more distinct predictive standard deviations that better distinguish the qualities of mean predictions, the three-layered DGP (in Figure~\subref*{fig:vega_si3_con}) trained by SI-IC seems to have the overall best performance by balancing NRMSEP, uncertainty quantification, and computation (see Figure~\subref*{fig:vega_time}).

\section{Conclusion}
\label{sec:conclusion}
In this study, a novel inference method, called stochastic imputation, for DGP emulation is introduced. By converting DGP emulations to linked GP emulations through stochastic imputations of latent layers using ESS, we simplify the training of a DGP emulator with constructions of conventional GP emulators. As a result, predictions from a DGP emulator can be made analytically tractable by computing the closed form predictive mean and variance of the corresponding linked GP emulators. We show in both synthetic and empirical examples that our method is a competitive candidate (in terms of predictive accuracy, uncertainty, and computational cost) for DGP surrogate modeling, in comparison to other state-of-the-art inferences such as DSVI and FB. In particular, we find some evidence that it can be beneficiary to implement SI with the input connection for better emulation performance. Empirical results suggest that SI may not give significant predictive improvement on DGP emulators as the number of layers in DGP increases (up to 4), and two- or three-layered DGP emulators trained by SI with the input-connected structure can often be satisfactory in terms of predictive accuracy and computational expense.

SI is algorithmically simple and it is natural to treat inference for DGP emulators as a missing data problem in which we have missingness on internal I/O of a network of conventional GP surrogates. This simplicity and interpretability makes SI generally applicable to any DGP hierarchies formed by feed-forward connected GPs, and thus allows various potential emulation scenarios, such as multi-fidelity emulation, multi-output emulation, linked emulation and their hybrids, to be implemented and explored under the same inference framework. The \texttt{Python} package \texttt{dgpsi} we developed as a by-product of this work is generally applicable to these advanced emulation problems and publicly available on GitHub (at \url{https://github.com/mingdeyu/DGP}).

Although we only discuss the emulation of deterministic models in this work, extension to stochastic models is straightforward using SI. One could add an extra Gaussian likelihood layer to the tail of DGP hierarchy to account for either homoscedastic or heteroscedastic~\citep{goldberg1997regression} noise exhibited in the stochastic computer simulators. Non-Gaussian likelihoods are a natural extension and are available in \texttt{dgpsi}. Future work worthy of investigation include DGP emulator-based sensitivity analysis, Bayesian optimization, and calibration, taking advantage of the DGP emulators' analytically tractable mean and variance implemented in SI. Coupling SI with sequential design~\citep{beck2016sequential, salmanidou2021probabilistic} to further reinforce the predictive performance of DGP emulators with reduced computational costs is another promising research direction. Applications of sequential designs to FB-based DGP emulation are explored by~\citet{sauer2020active}.  

Although SI utilizes all data points in the dataset, this does not pose a serious computational problem to typical computer model experiments because the involved datasets are often of small-to-moderate sizes given limited computational budgets. However, when one has a big dataset, the method can become practically infeasible due to the high computational complexity associated to the storage, processing and analysis of the huge amount of data points. Therefore, it would be an interesting future work to scale the stochastic imputation method to big data, e.g., via sparse approximation~\citep{snelson2005sparse} or GPU acceleration.

\bibliographystyle{agsm}  
\bibliography{references}  


\vfill

\pagebreak
\begin{center}
\label{supp}
\textbf{\Large Supplementary Materials}
\end{center}
\setcounter{equation}{0}
\setcounter{section}{0}
\setcounter{figure}{0}
\makeatletter
\renewcommand{\thesection}{S.\arabic{section}}
\renewcommand{\theequation}{S\arabic{equation}} 
\renewcommand{\thefigure}{S.\arabic{figure}} 

\section{Five-Dimensional Example}
\label{sec:5d-case}
Consider a modified 5-D function from~\citet{montagna2016computer} that is given by
\begin{equation}
\label{eq:5d}
    f(x_1,x_2,x_3,x_4,x_5)=\begin{cases*}
    \exp\left\{\sum^5_{i=1}\left(\frac{1}{i}\right)^2 x_i\right\},\quad \mathrm{if}\;x_1,x_2,x_3,x_4,x_5>0.2\\
    0,\quad \mathrm{otherwise}
    \end{cases*}
\end{equation}
on the hypercube $\mathbf{x}=(x_1,x_2,x_3,x_4,x_5)\in[0,1]^5$. To form our training dataset, we draw $100$ Latin hypercube input positions, at which we evaluate~\eqref{eq:5d} to obtain the corresponding output. We then construct both two- and three-layered DGP emulators of~\eqref{eq:5d} using FB, DSVI, SI, and SI-IC respectively. We additionally draw $500$ Latin hypercube input positions over the domain $[0,1]^5$ to form our testing dataset. The trained DGP emulators are subsequently used to predict functional values (in terms of mean predictions and predictive standard deviations) at the drawn $500$ testing input positions. Due to the stochasticity of the methods, we conduct $20$ inference trials, resulting in $20$ DGP emulators, for each method.  

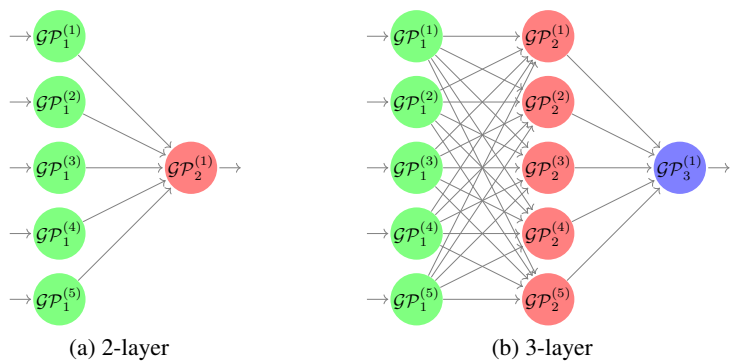
\begin{figure}[!ht]
\centering 
\subfloat[2-layer]{\label{fig:5d_2layer}
\scalebox{0.7}{
\begin{tikzpicture}[shorten >=1pt,->,draw=black!50, node distance=4cm]
    \tikzstyle{every pin edge}=[<-,shorten <=1pt]
    \tikzstyle{neuron}=[circle,fill=black!25,minimum size=27.5pt,inner sep=0pt]
    \tikzstyle{layer1}=[neuron, fill=green!50];
    \tikzstyle{layer2}=[neuron, fill=red!50];
    \tikzstyle{layer3}=[neuron, fill=blue!50];
    \tikzstyle{annot} = [text width=2em, text centered];
    \node[layer1,pin={[pin edge={<-}]left:}] (l1-0) at (0,0) {$\mathcal{GP}^{(1)}_{1}$};
    \node[layer1,pin={[pin edge={<-}]left:}] (l1-1) at (0,-1.25) {$\mathcal{GP}^{(2)}_{1}$};
    \node[layer1,pin={[pin edge={<-}]left:}] (l1-2) at (0,-2.5) {$\mathcal{GP}^{(3)}_{1}$};
    \node[layer1,pin={[pin edge={<-}]left:}] (l1-3) at (0,-3.75) {$\mathcal{GP}^{(4)}_{1}$};
    \node[layer1,pin={[pin edge={<-}]left:}] (l1-4) at (0,-5) {$\mathcal{GP}^{(5)}_{1}$};
    \node[layer2,pin={[pin edge={->}]right:}] (l2-0) at (2.5,-2.5) {$\mathcal{GP}^{(1)}_{2}$};
\draw[->] (l1-0) -- (l2-0);
\draw[->] (l1-1) -- (l2-0);
\draw[->] (l1-2) -- (l2-0);
\draw[->] (l1-3) -- (l2-0);
\draw[->] (l1-4) -- (l2-0);
\end{tikzpicture}}}\hspace{3em}
\subfloat[3-layer]{\label{fig:5d_3layer}
\scalebox{0.7}{
\begin{tikzpicture}[shorten >=1pt,->,draw=black!50, node distance=4cm]
    \tikzstyle{every pin edge}=[<-,shorten <=1pt]
    \tikzstyle{neuron}=[circle,fill=black!25,minimum size=27.5pt,inner sep=0pt]
    \tikzstyle{layer1}=[neuron, fill=green!50];
    \tikzstyle{layer2}=[neuron, fill=red!50];
    \tikzstyle{layer3}=[neuron, fill=blue!50];
    \tikzstyle{annot} = [text width=2em, text centered];
    \node[layer1,pin={[pin edge={<-}]left:}] (l1-0) at (0,0) {$\mathcal{GP}^{(1)}_{1}$};
    \node[layer1,pin={[pin edge={<-}]left:}] (l1-1) at (0,-1.25) {$\mathcal{GP}^{(2)}_{1}$};
    \node[layer1,pin={[pin edge={<-}]left:}] (l1-2) at (0,-2.5) {$\mathcal{GP}^{(3)}_{1}$};
    \node[layer1,pin={[pin edge={<-}]left:}] (l1-3) at (0,-3.75) {$\mathcal{GP}^{(4)}_{1}$};
    \node[layer1,pin={[pin edge={<-}]left:}] (l1-4) at (0,-5) {$\mathcal{GP}^{(5)}_{1}$};
    \node[layer2] (l2-0) at (2.5,0) {$\mathcal{GP}^{(1)}_{2}$};
    \node[layer2] (l2-1) at (2.5,-1.25) {$\mathcal{GP}^{(2)}_{2}$};
    \node[layer2] (l2-2) at (2.5,-2.5) {$\mathcal{GP}^{(3)}_{2}$};
    \node[layer2] (l2-3) at (2.5,-3.75) {$\mathcal{GP}^{(4)}_{2}$};
    \node[layer2] (l2-4) at (2.5,-5) {$\mathcal{GP}^{(5)}_{2}$};
    \node[layer3,pin={[pin edge={->}]right:}] (l3-0) at (5,-2.5) {$\mathcal{GP}^{(1)}_{3}$};
\draw[->] (l1-0) -- (l2-0);
\draw[->] (l1-1) -- (l2-0);
\draw[->] (l1-2) -- (l2-0);
\draw[->] (l1-3) -- (l2-0);
\draw[->] (l1-4) -- (l2-0);
\draw[->] (l1-0) -- (l2-1);
\draw[->] (l1-1) -- (l2-1);
\draw[->] (l1-2) -- (l2-1);
\draw[->] (l1-3) -- (l2-1);
\draw[->] (l1-4) -- (l2-1);
\draw[->] (l1-0) -- (l2-2);
\draw[->] (l1-1) -- (l2-2);
\draw[->] (l1-2) -- (l2-2);
\draw[->] (l1-3) -- (l2-2);
\draw[->] (l1-4) -- (l2-2);
\draw[->] (l1-0) -- (l2-3);
\draw[->] (l1-1) -- (l2-3);
\draw[->] (l1-2) -- (l2-3);
\draw[->] (l1-3) -- (l2-3);
\draw[->] (l1-4) -- (l2-3);
\draw[->] (l1-0) -- (l2-4);
\draw[->] (l1-1) -- (l2-4);
\draw[->] (l1-2) -- (l2-4);
\draw[->] (l1-3) -- (l2-4);
\draw[->] (l1-4) -- (l2-4);
\draw[->] (l2-0) -- (l3-0);
\draw[->] (l2-1) -- (l3-0);
\draw[->] (l2-2) -- (l3-0);
\draw[->] (l2-3) -- (l3-0);
\draw[->] (l2-4) -- (l3-0);
\end{tikzpicture}}}
\caption{Two DGP structures used to construct DGP emulators of the 5-D function~\eqref{eq:5d}.}
\label{fig:5d_structure}
\end{figure}

\subsection{Results}
\begin{figure}[!ht]
\centering 
\subfloat[NRMSEP]{\label{fig:5d_rmse}\includegraphics[width=0.45\linewidth]{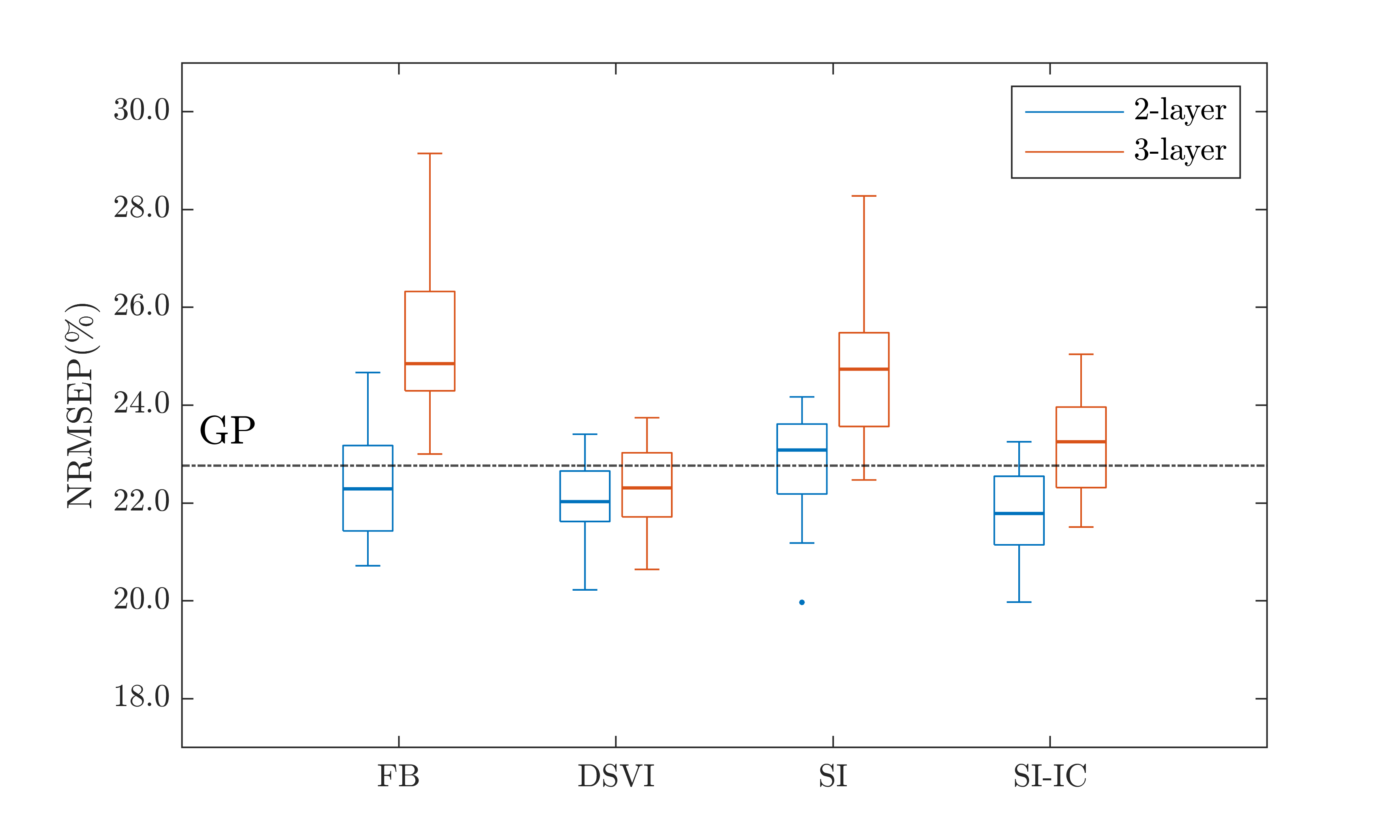}}
\subfloat[Computation time]{\label{fig:5d_time}\includegraphics[width=0.45\linewidth]{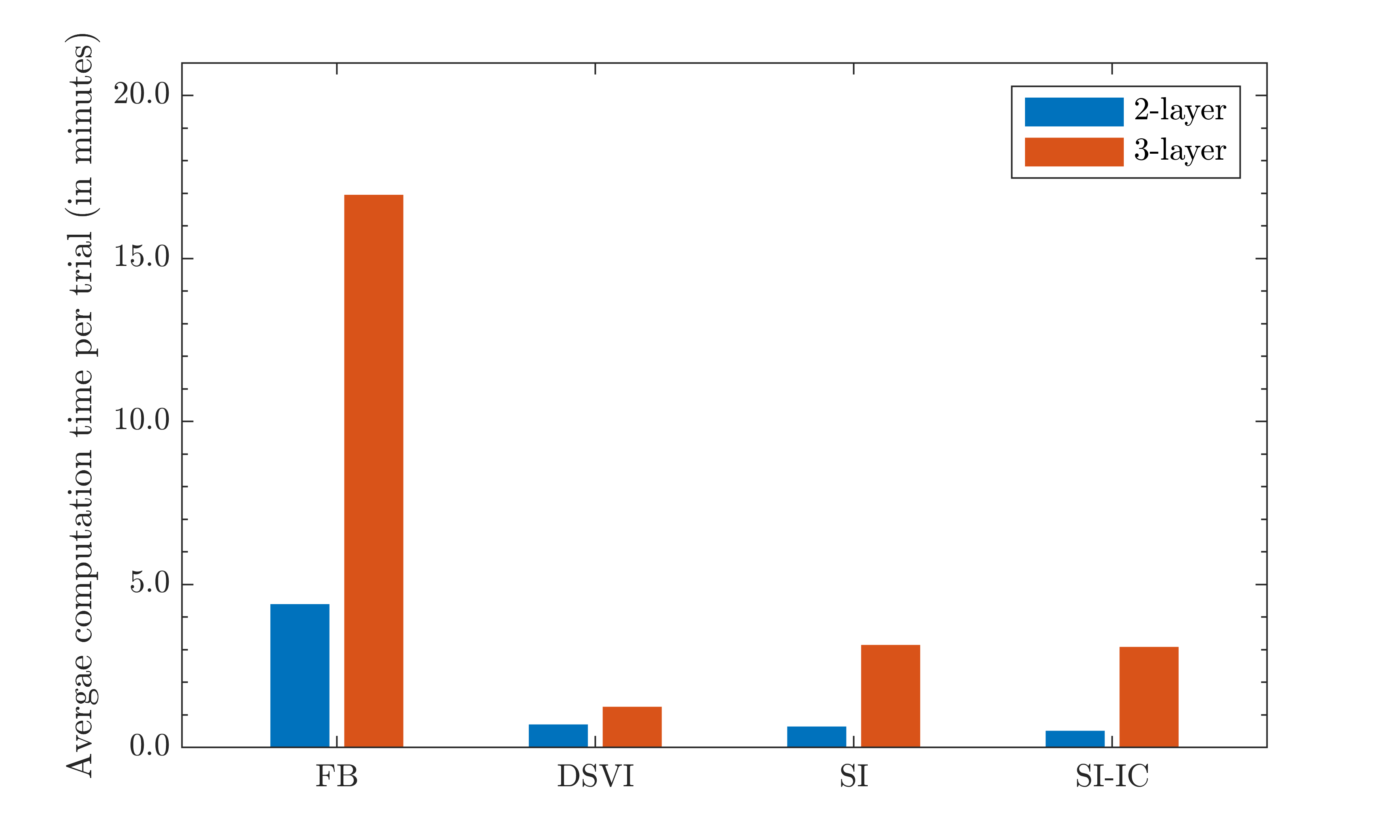}} 
\caption{Comparison of FB, DSVI, SI, and SI-IC in terms of NRMSEP for $20$ repeatedly trained DGP emulators and the computation time of corresponding implementation packages. The dash-dot line represents the NRMSEP of a trained conventional GP emulator.}
\label{fig:5d_compare}
\end{figure}

 There are several observations that can be drawn from Figure~\subref*{fig:5d_rmse} and~\subref*{fig:5d_time}. Firstly, regardless of the inference methods, the three-layered DGP emulators exhibit higher NRMSEP than the two-layered ones across different trials. Secondly, in comparison to FB, SI is generally faster to implement and can achieve comparable overall accuracy on mean predictions for both two- and three-layered DGP emulators. Finally, two-layered DGPs produced by SI-IC have the lowest overall NRMSEP whilst three-layered DGPs trained by DSVI give the best overall mean prediction accuracy. 

\begin{figure}[!ht]
\centering 
\subfloat[2-layer (FB)]{\label{fig:5d_fb2}\includegraphics[width=0.3\linewidth]{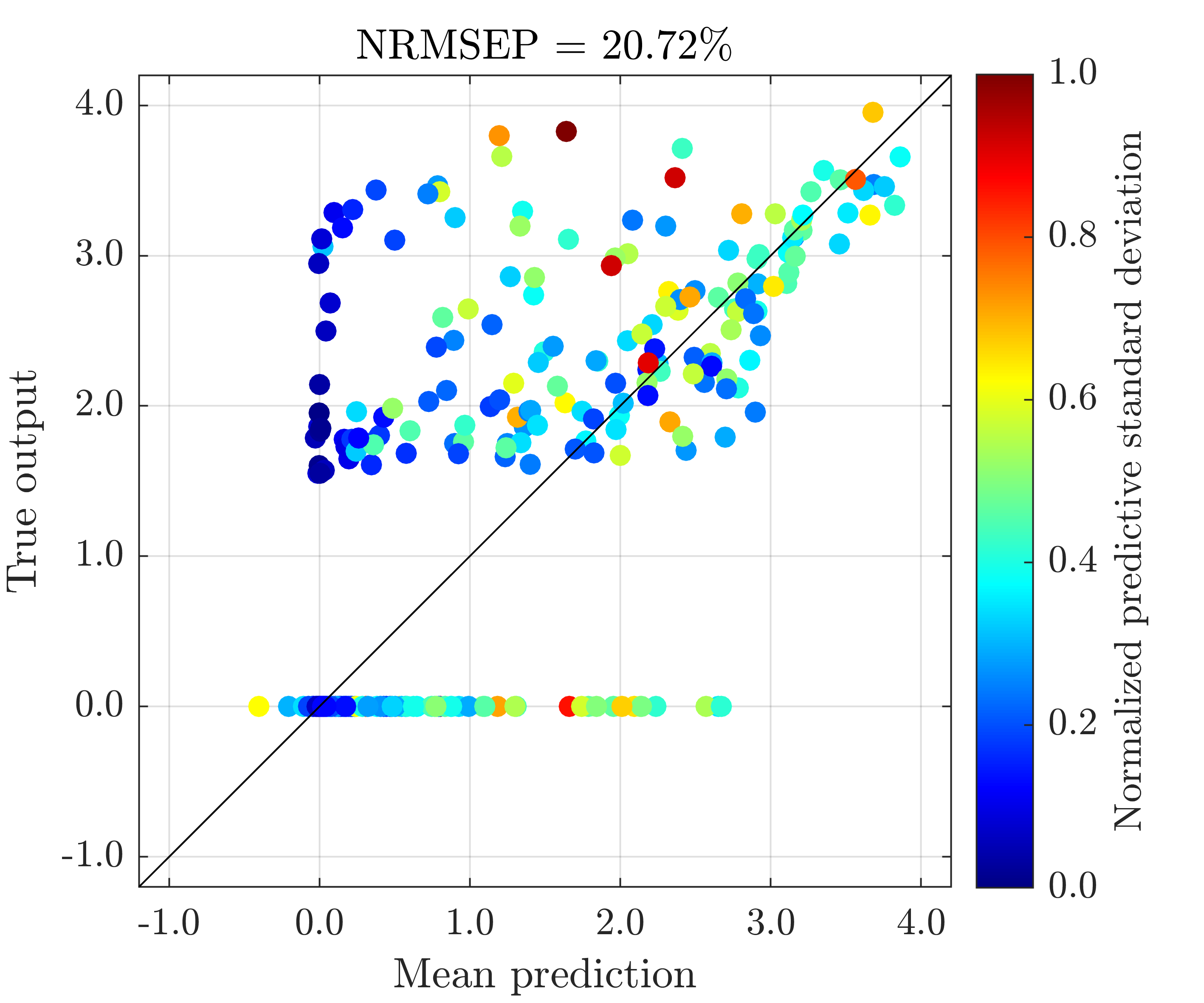}}\hspace{1.5em}
\subfloat[2-layer (DSVI)]{\label{fig:5d_vi2}\includegraphics[width=0.3\linewidth]{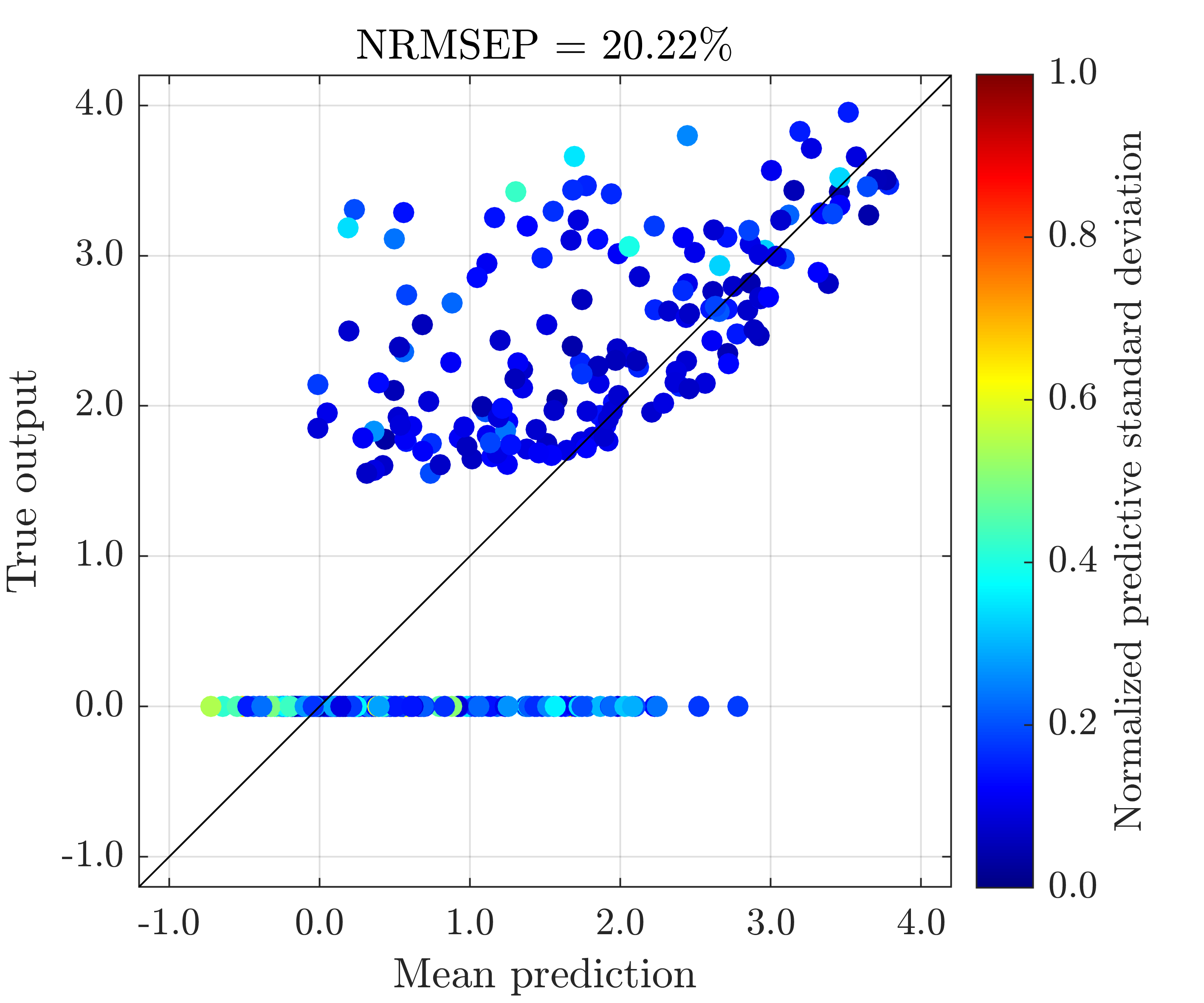}}\hspace{1.5em}
\subfloat[2-layer (SI)]{\label{fig:5d_si2}\includegraphics[width=0.3\linewidth]{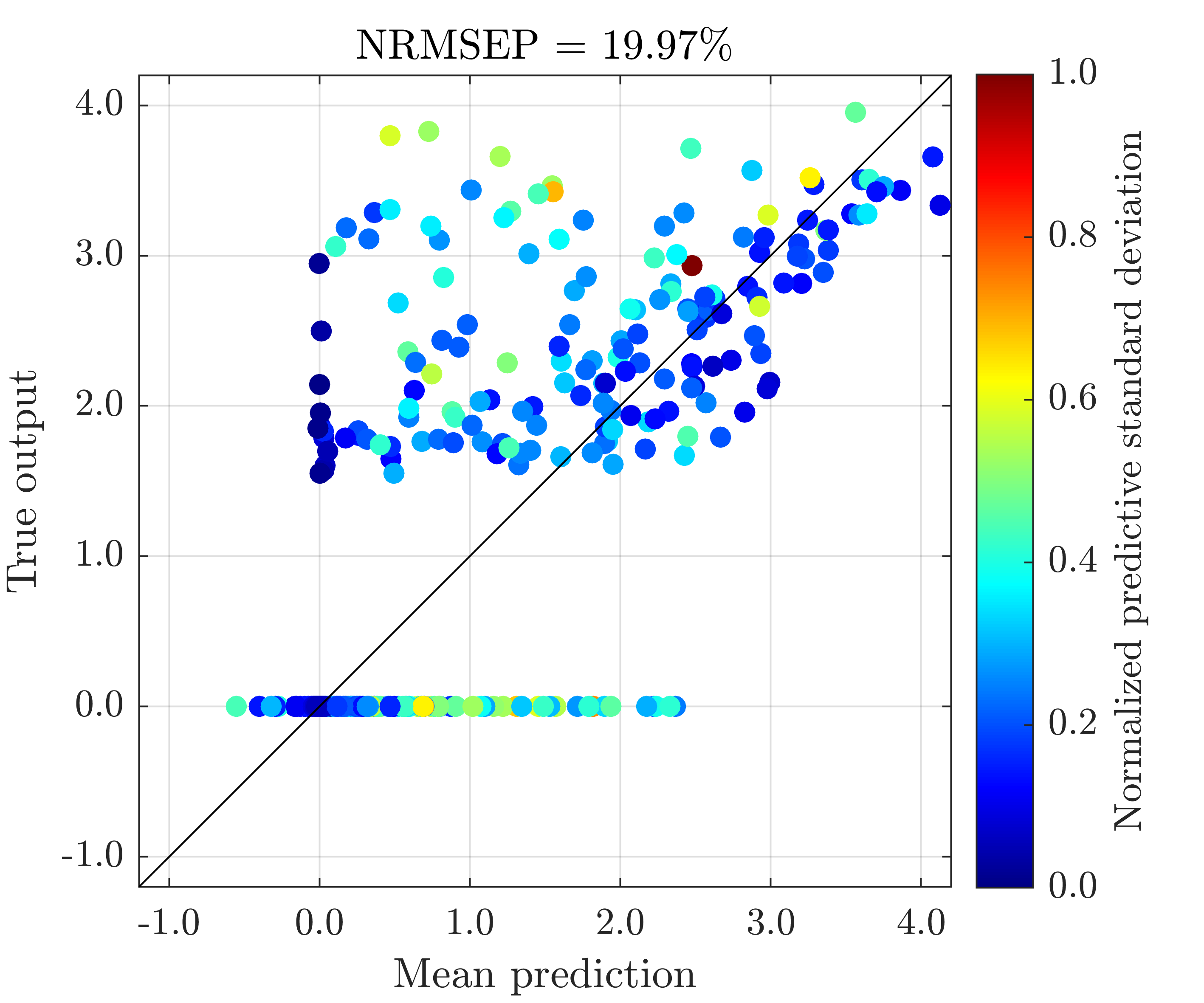}}\\
\subfloat[3-layer (FB)]{\label{fig:5d_fb3}\includegraphics[width=0.3\linewidth]{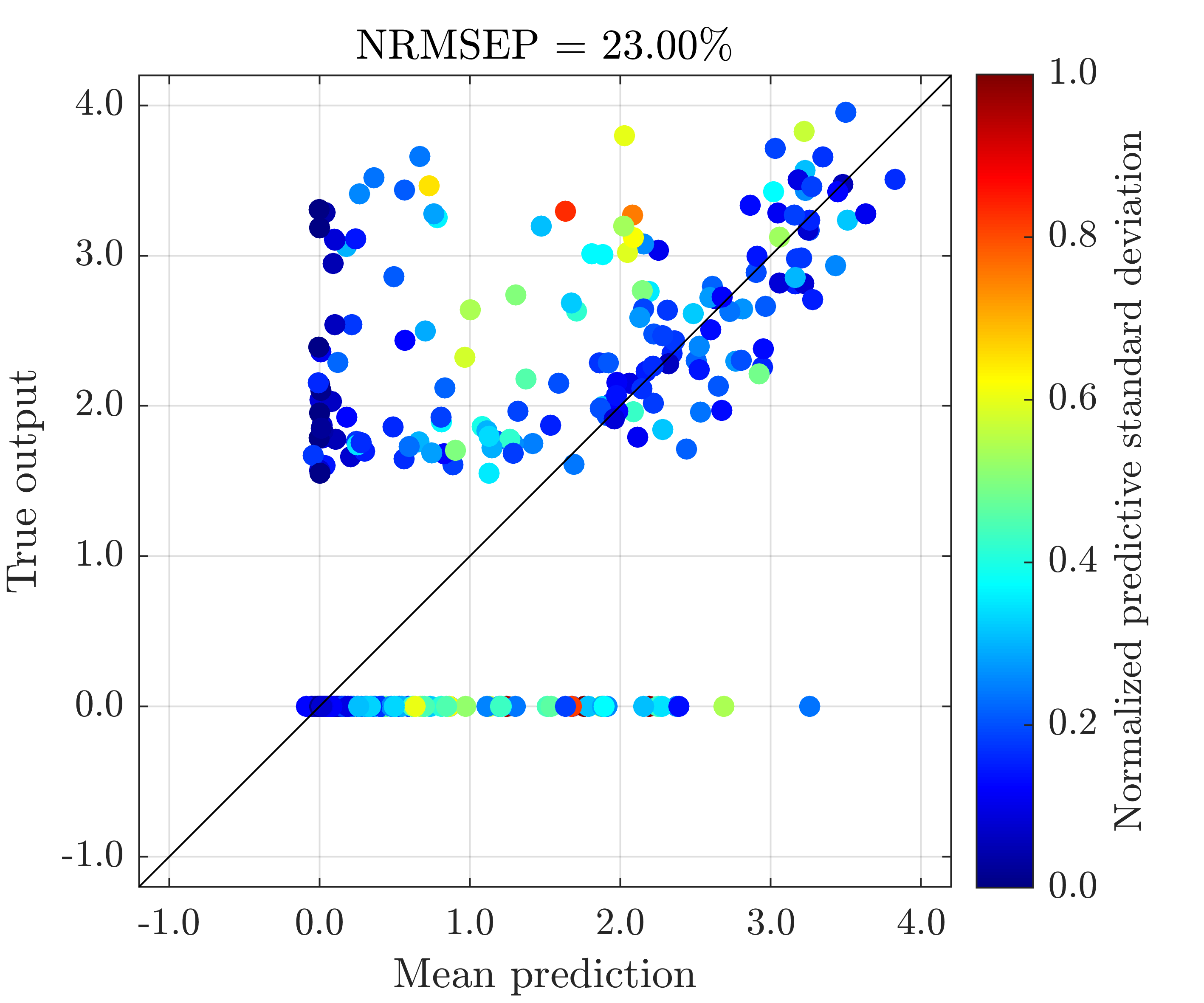}}\hspace{1.5em}
\subfloat[3-layer (DSVI)]{\label{fig:5d_vi3}\includegraphics[width=0.3\linewidth]{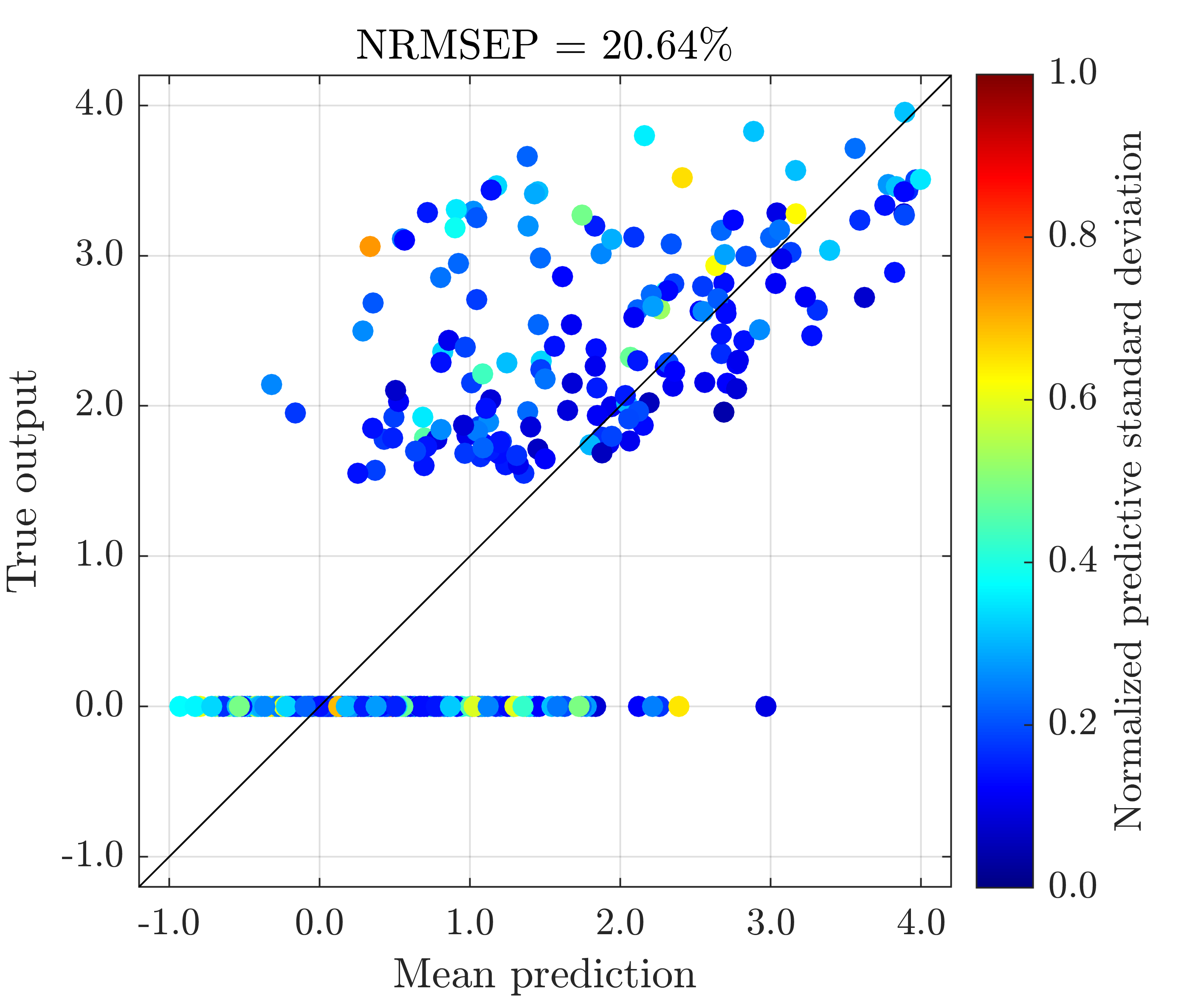}}\hspace{1.5em}
\subfloat[3-layer (SI)]{\label{fig:5d_si3}\includegraphics[width=0.3\linewidth]{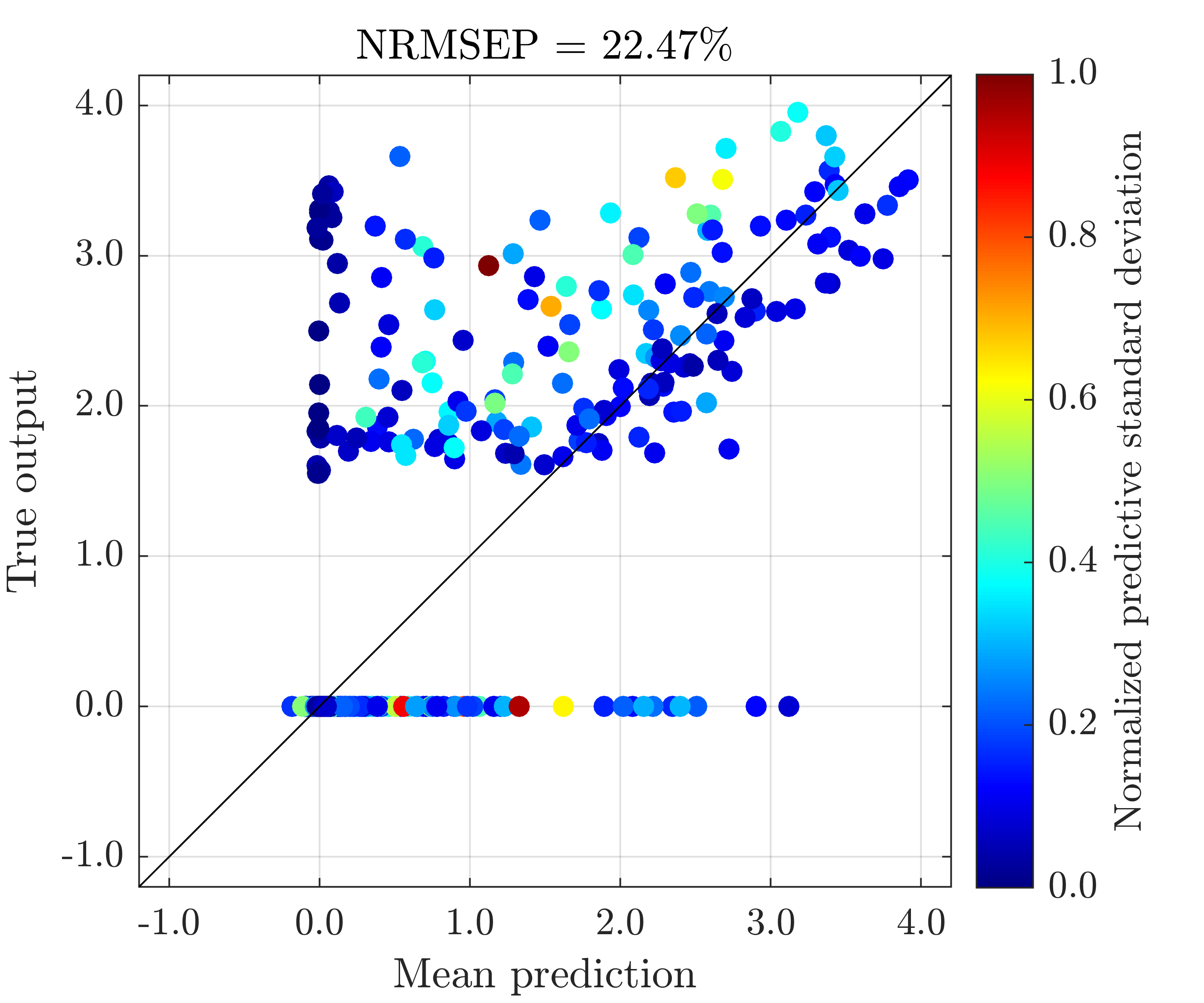}} \\
\subfloat[2-layer (SI-IC)]{\label{fig:5d_si2_con}\includegraphics[width=0.3\linewidth]{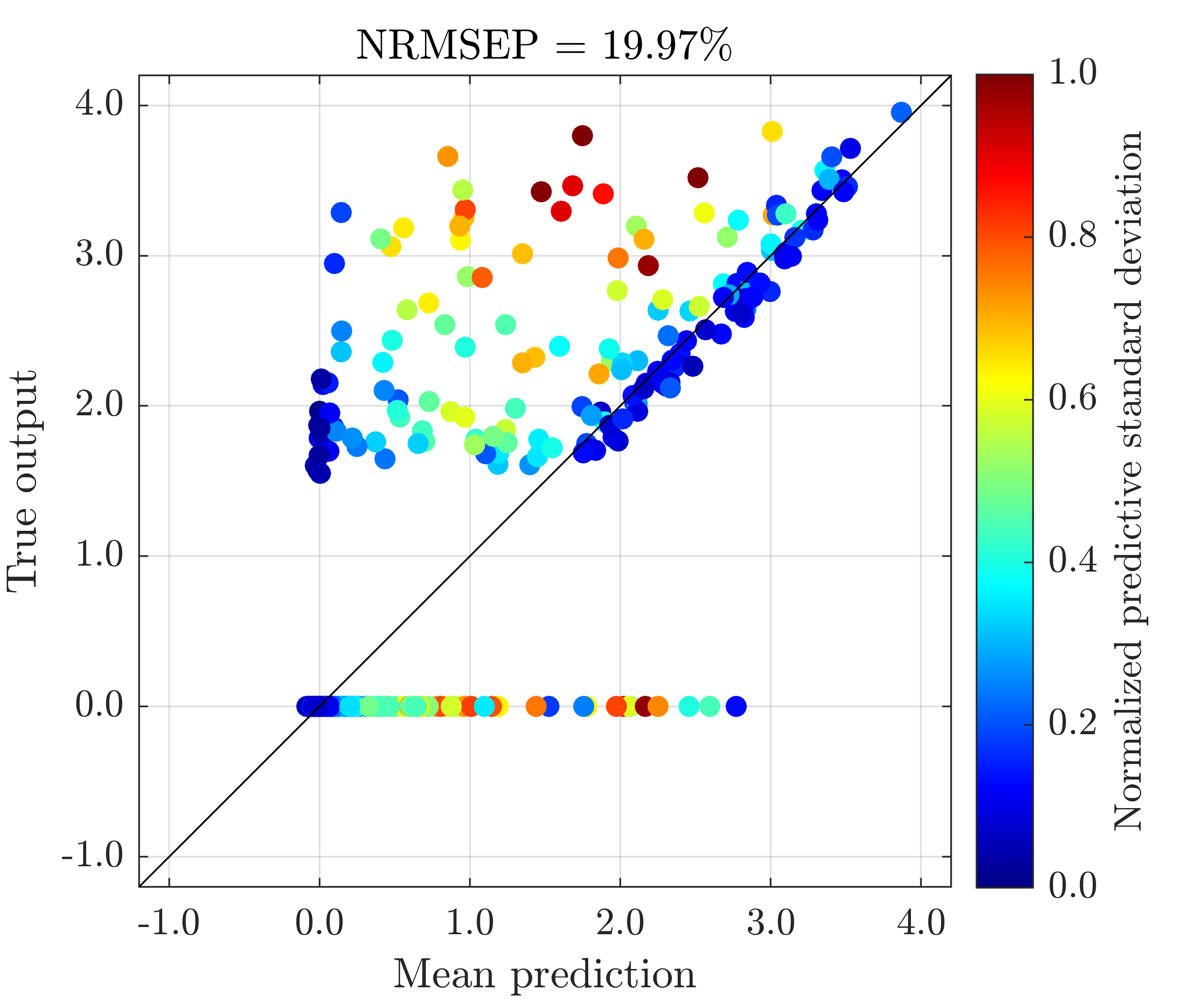}}\hspace{1.5em}
\subfloat[3-layer (SI-IC)]{\label{fig:5d_si3_con}\includegraphics[width=0.3\linewidth]{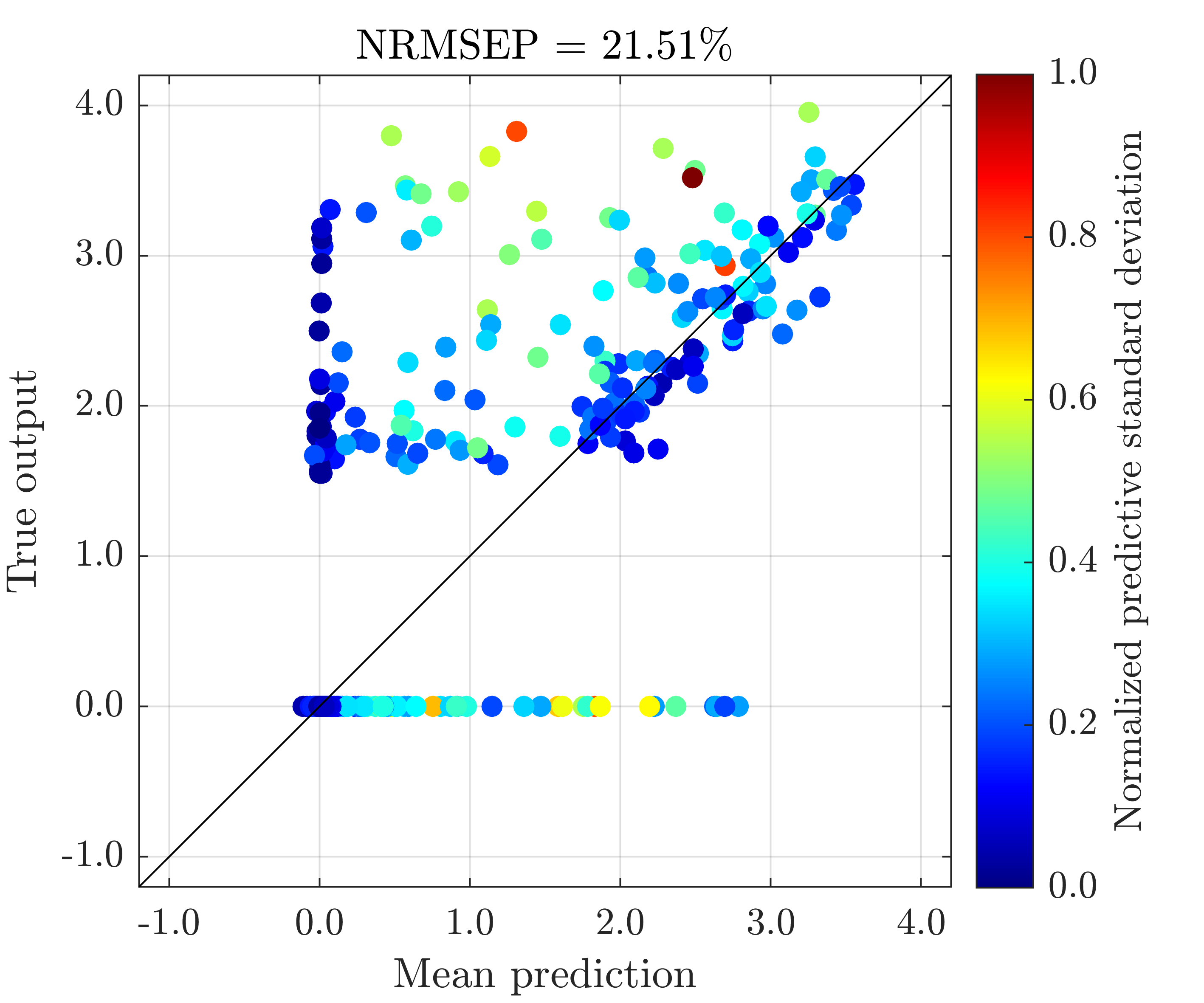}}\hspace{1.5em}
\subfloat[GP]{\label{fig:5d_gp}\includegraphics[width=0.3\linewidth]{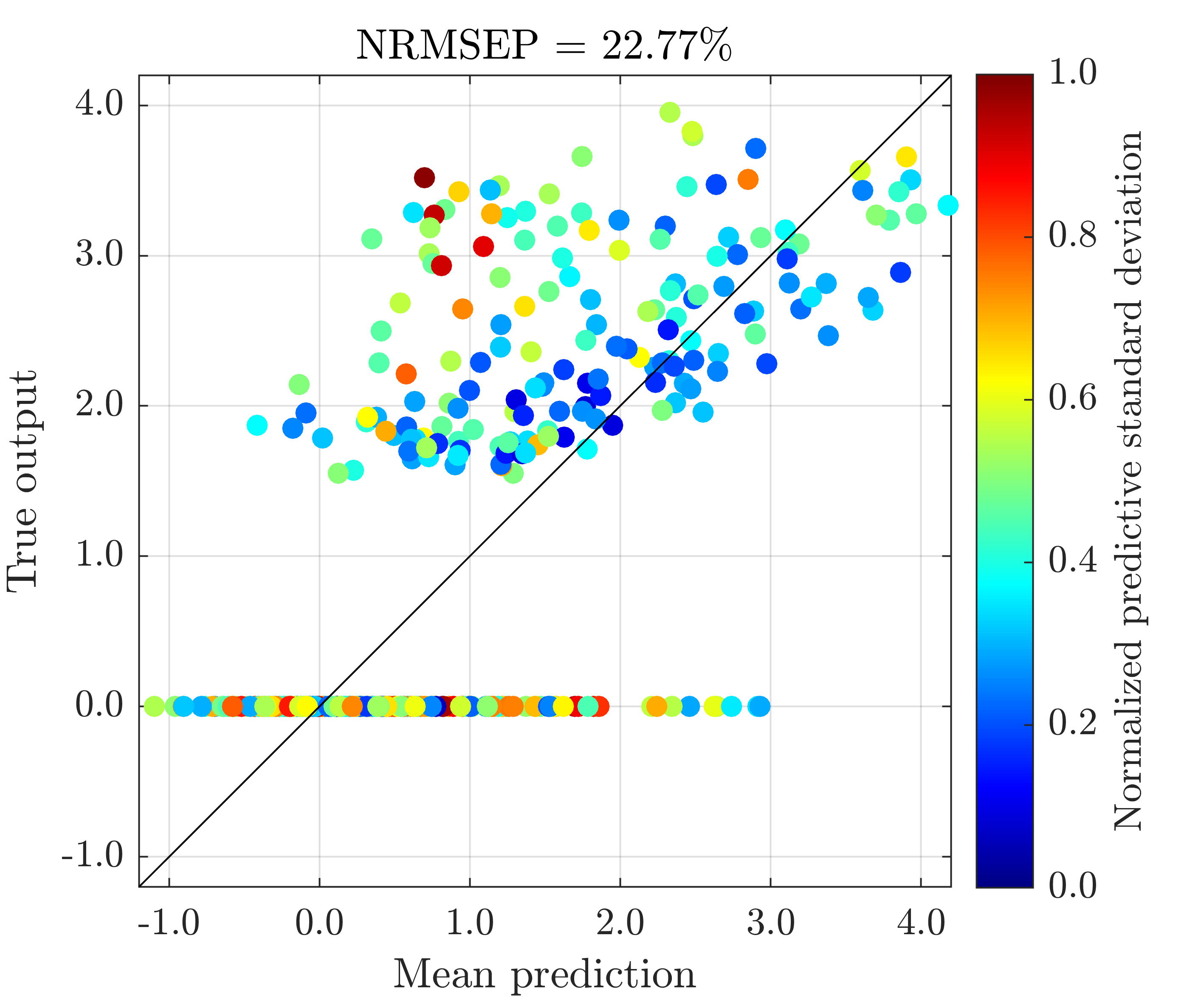}} 
\caption{Plots of true outputs from function~\eqref{eq:5d} at $500$ testing positions \emph{vs} mean predictions, along with predictive standard deviations (normalized by their max and min values), made by the best emulator (i.e., the one with the lowest NRMSEP, shown on top of each subfigure, out of $20$ inference trials) produced by FB, DSVI, SI and SI-IC. \emph{GP} represents a trained conventional GP emulator.}
\label{fig:5d_exp_diag}
\end{figure}

To examine the uncertainty quantified by the DGP emulators, we pick the best trial (i.e., the DGP emulator with the lowest NRMSEP) produced by each of the four approaches, and plot the mean predictions along with the predictive standard deviations against the true functional outputs in Figure~\ref{fig:5d_exp_diag}. Figure~\subref*{fig:5d_vi2} and~\subref*{fig:5d_vi3} show that in general DGP emulators trained by DSVI underestimate the uncertainties, giving low predictive standard deviations across different testing positions. DGP emulators trained by SI present slightly better uncertainty profiles, while those produced by FB exhibit fairly satisfactory profiles. DGP emulators from SI-IC appear to show the best performance on uncertainty quantification because their predictive standard deviations are significantly higher at testing positions where the corresponding true functional outputs are not well-predicted by the predictive means. Not surprisingly, the conventional GP emulator fails to distinguish good mean predictions from bad ones through its produced predictive standard deviations.

We note that DGP emulators trained by FB, SI, and SI-IC seem over-confident at some testing positions (e.g., the far left points with low predictive standard deviations in Figure~\subref*{fig:5d_fb3},~\subref*{fig:5d_si3} and~\subref*{fig:5d_si3_con}) where they actually make poor predictions. This is because these testing inputs (with non-zero responses) reside near the abrupt transitional regions of the underlying function and DGP emulators can incorrectly predict the corresponding output values with zeros on the flat boundary surface.

\section{Aircraft Engine Model}
\label{sec:aircraft}
Aircraft engine simulation is often involved in aircraft design for multi-disciplinary design optimization (MDO). However, the analysis of engine and its design optimization could require a large number of engine model evaluations, which can be computationally expensive~\citep{lyu2020flight}, across different flight conditions. Therefore, it is desirable to replace the engine model with a computationally cheaper surrogate model. In this section, we consider the Boeing 777 engine model~\citep{lyu2020flight}, which takes three inputs: aircraft altitude ($km$), Mach number and throttle, and produces two outputs: thrust ($N$) and Thrust-Specific Fuel Consumption (TSFC, $N/N/s$). The engine model exhibits non-stationarity with respect to output TSFC since TSFC quickly climbs up when Mach number is large whilst throttle position is small. We thus construct DGP emulators of the engine with respect to the output TSFC, with the same three formations shown in Figure~\ref{fig:archs} of the manuscript, using the dataset (available at \url{https://github.com/SMTorg/smt/tree/master/smt/examples/b777_engine}) published by~\citet{bouhlel2019python}. $100$ data points are drawn uniformly without replacement from the dataset to form the training data. After excluding the training data points from the dataset, we draw uniformly $500$ data points without replacement to form the testing data. For two-layer and three-layer formations, DGP emulators are constructed using FB, DSVI, SI, and SI-IC, while for four-layer formation only DSVI, SI, and SI-IC are implemented because \texttt{deepgp} only allows DGP hierarchies up to three layers. For each formation and method, we conduct $40$ inference trials, i.e., construct $40$ DGP emulators repeatedly.

\subsection{Results}
\begin{figure}[!ht]
\centering 
\subfloat[NRMSEP]{\label{fig:engine_rmse}\includegraphics[width=0.45\linewidth]{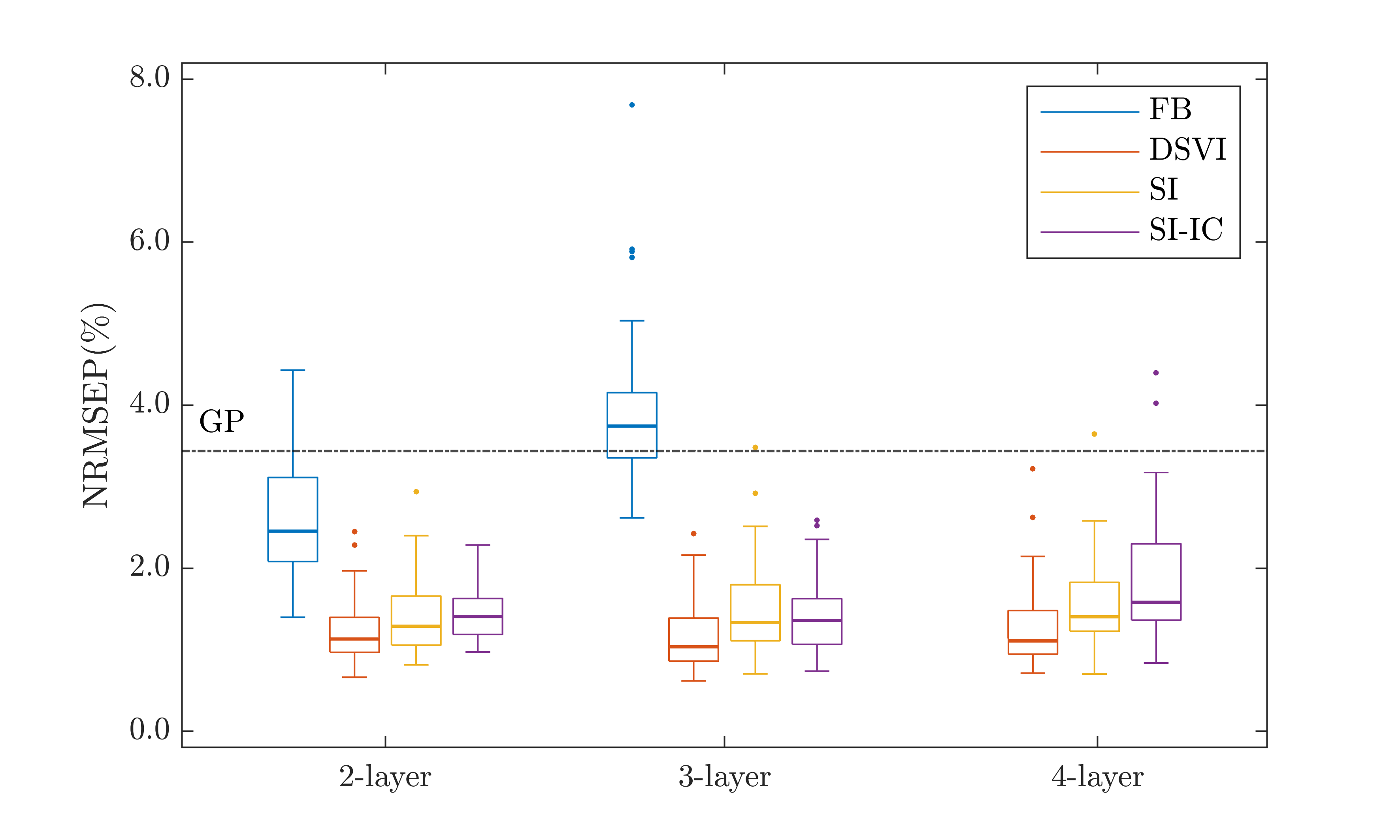}}
\subfloat[Computation time]{\label{fig:engine_time}\includegraphics[width=0.45\linewidth]{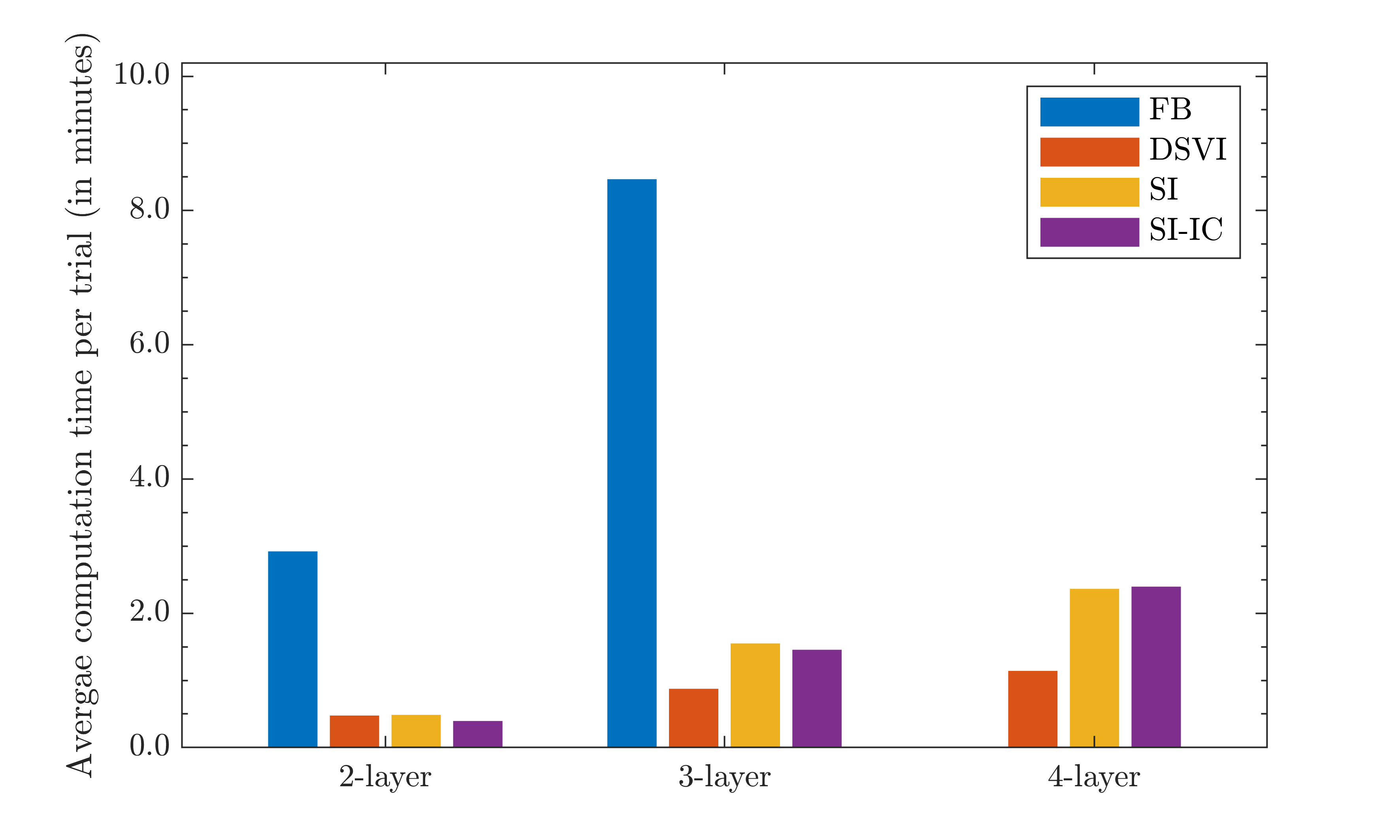}} 
\caption{Comparison of FB, DSVI, SI, and SI-IC for $40$ repeatedly trained DGP emulators (i.e., $40$ inference trials) of the engine model (with respect to TSFC). The dash-dot line represents the NRMSEP of a trained conventional GP emulator.}
\label{fig:engine_compare}
\end{figure}

Figure~\subref*{fig:engine_rmse} indicates that in general the mean predictions of DGP emulators outperform those of conventional GP emulator. For all formations, DSVI provides best overall accuracy on mean predictions. For two- and three-layer formations, SI produces DGP emulators with comparable accuracy to those trained by DSVI. It can also be seen that in this example input-connection does not give obvious improvements on mean predictions.

\begin{figure}[!ht]
\centering 
\subfloat[2-layer (FB)]{\label{fig:engine_fb2}\includegraphics[width=0.25\linewidth]{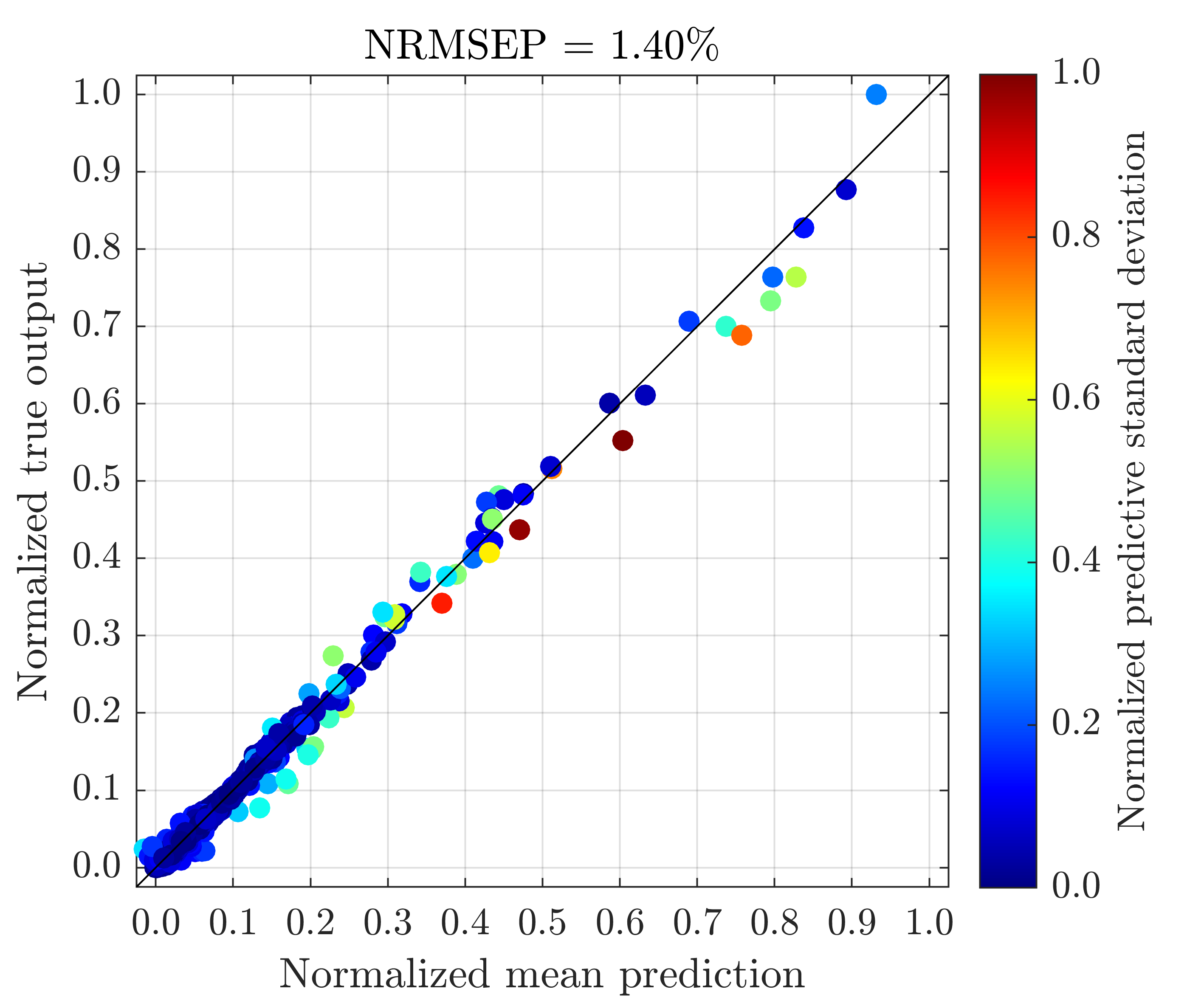}}
\subfloat[2-layer (DSVI)]{\label{fig:engine_vi2}\includegraphics[width=0.25\linewidth]{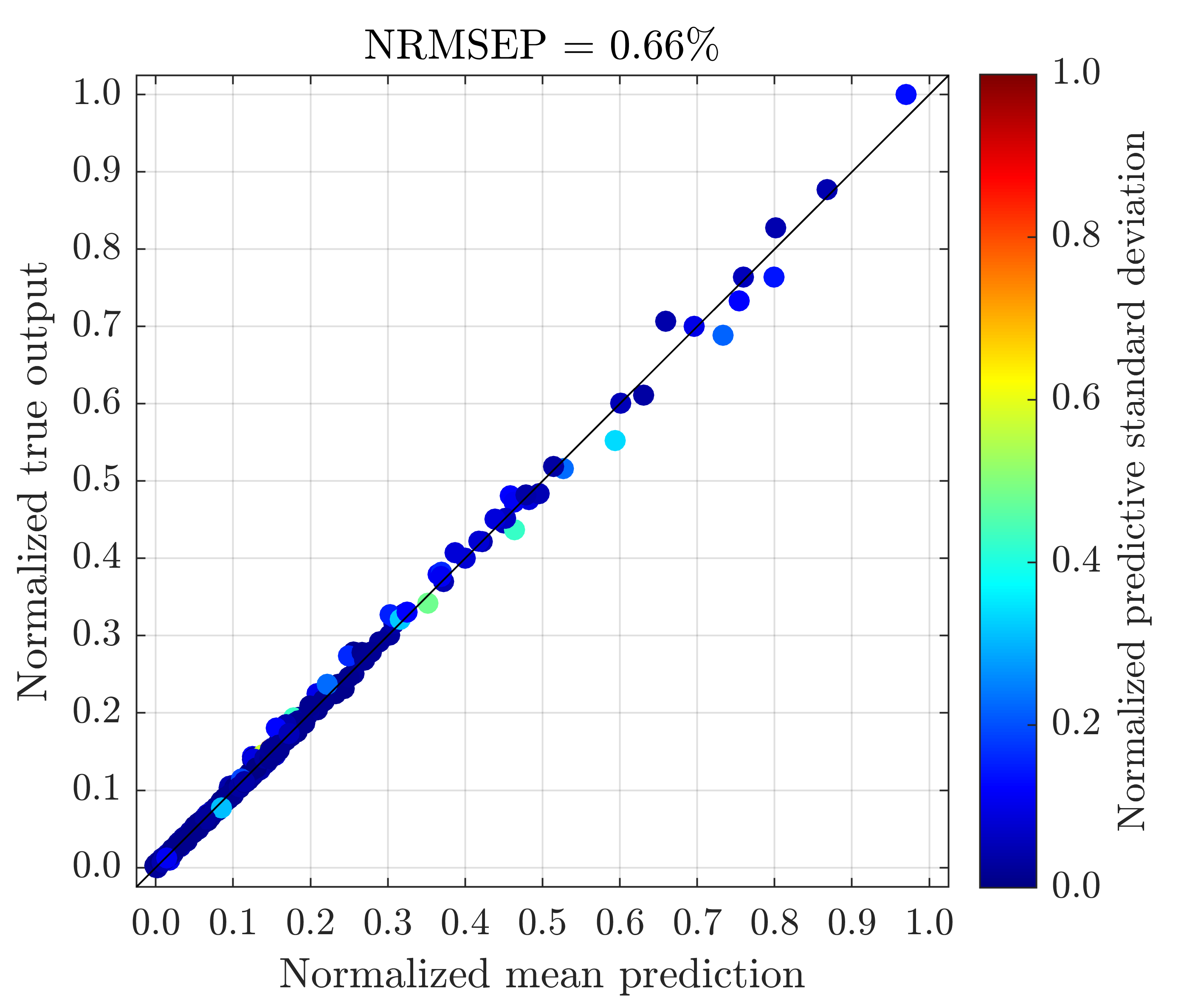}}
\subfloat[2-layer (SI)]{\label{fig:engine_si2}\includegraphics[width=0.25\linewidth]{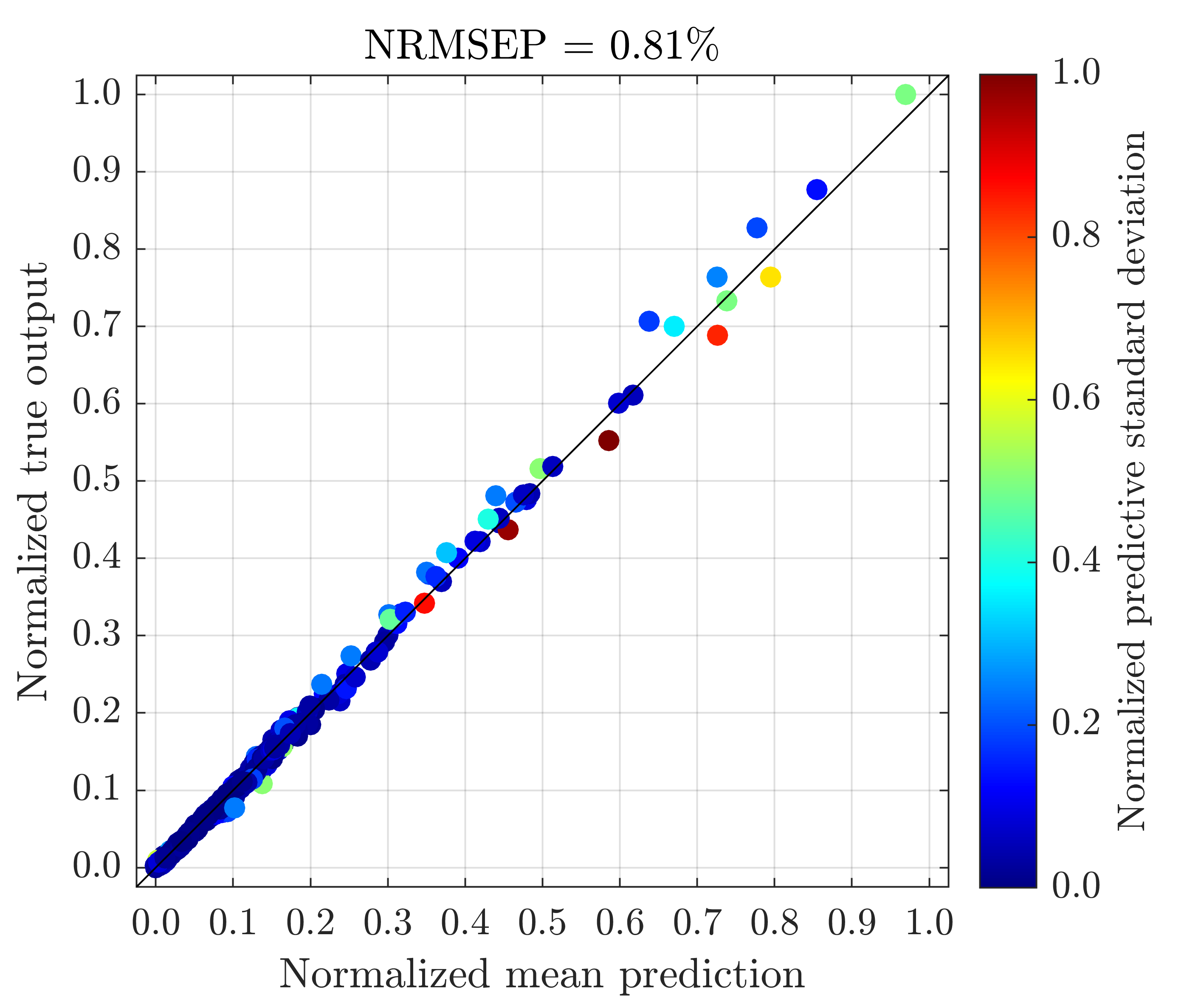}}
\subfloat[2-layer (SI-IC)]{\label{fig:engine_si2_con}\includegraphics[width=0.25\linewidth]{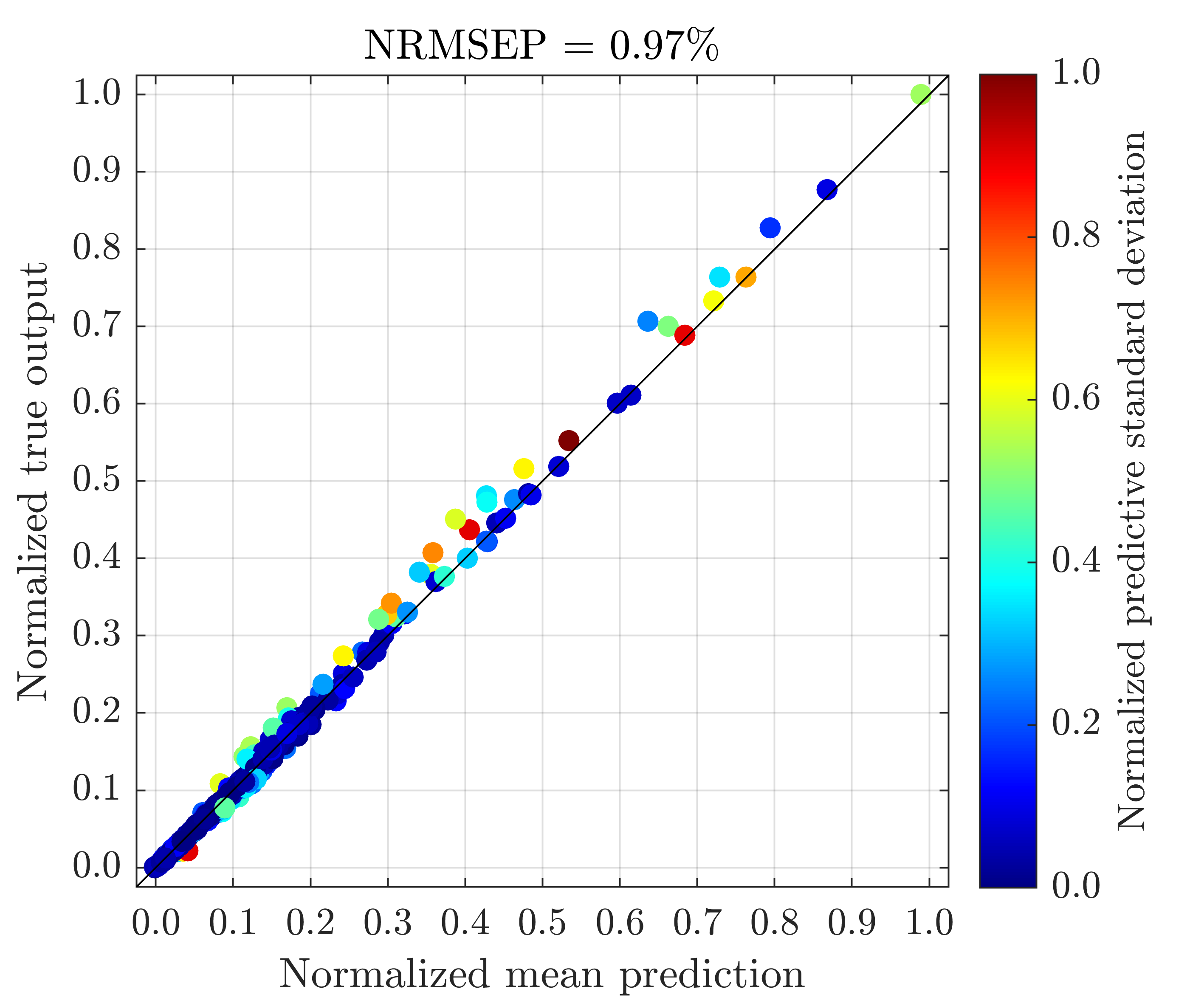}}\\
\subfloat[3-layer (FB)]{\label{fig:engine_fb3}\includegraphics[width=0.25\linewidth]{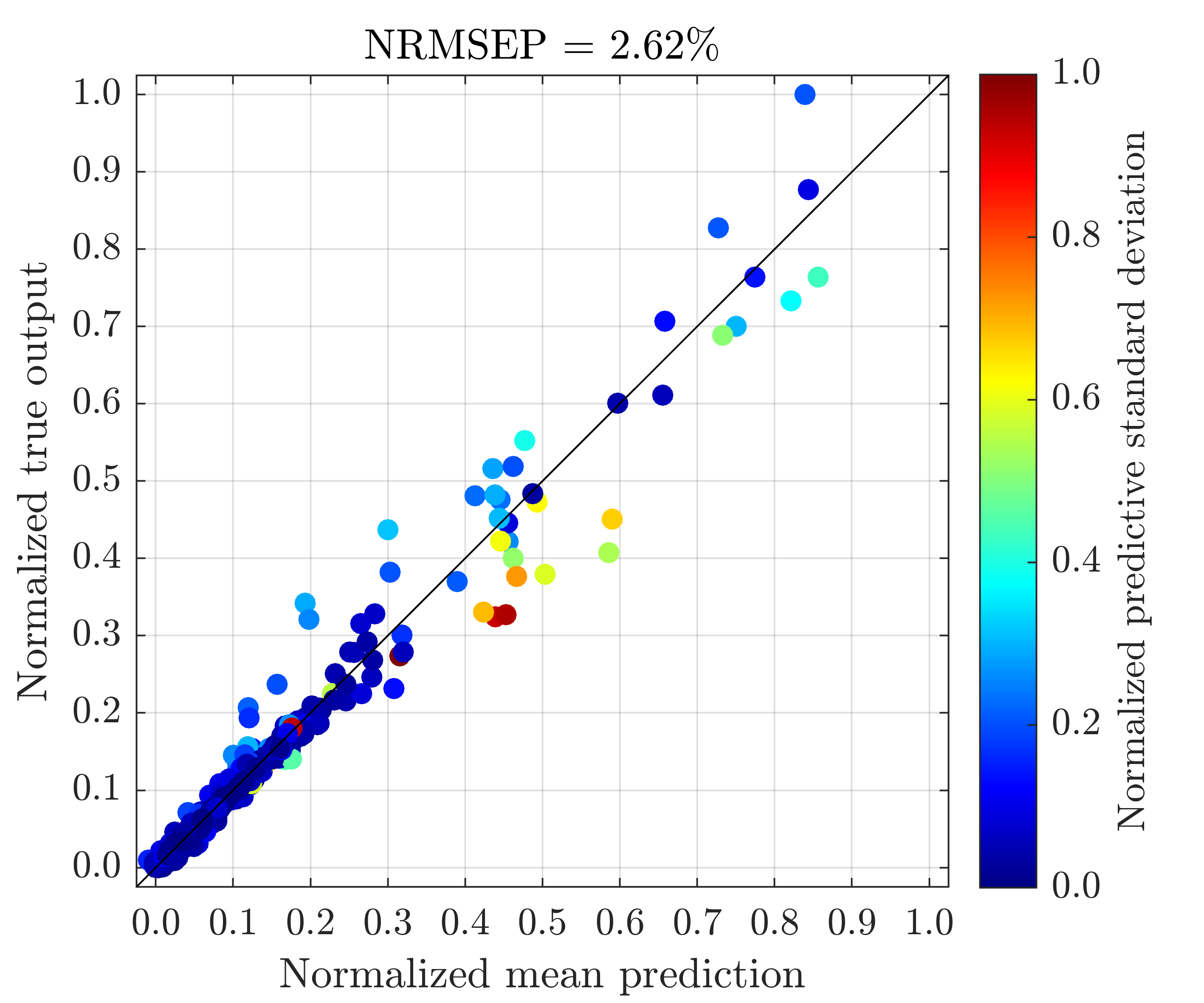}}
\subfloat[3-layer (DSVI)]{\label{fig:engine_vi3}\includegraphics[width=0.25\linewidth]{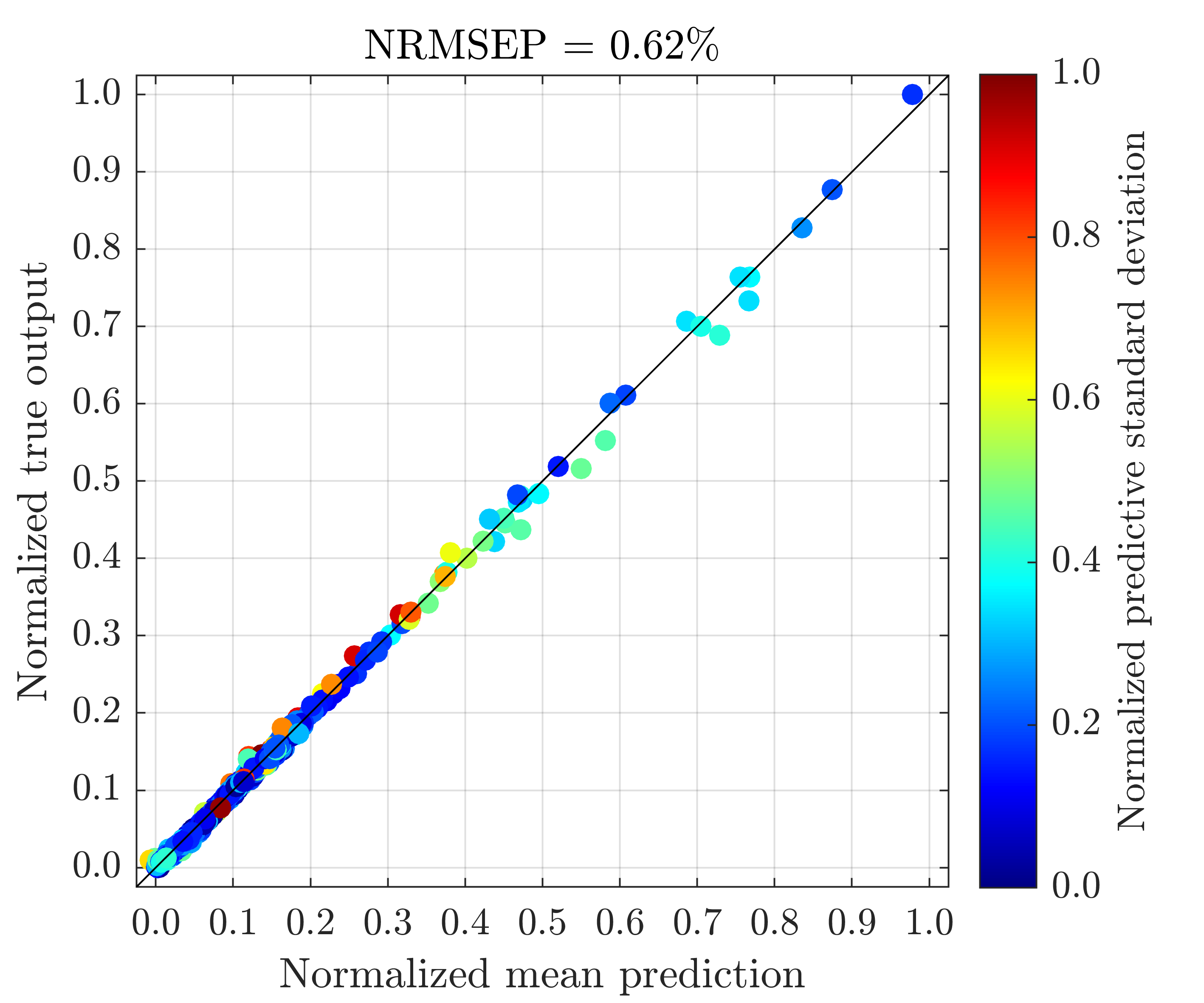}}
\subfloat[3-layer (SI)]{\label{fig:engine_si3}\includegraphics[width=0.25\linewidth]{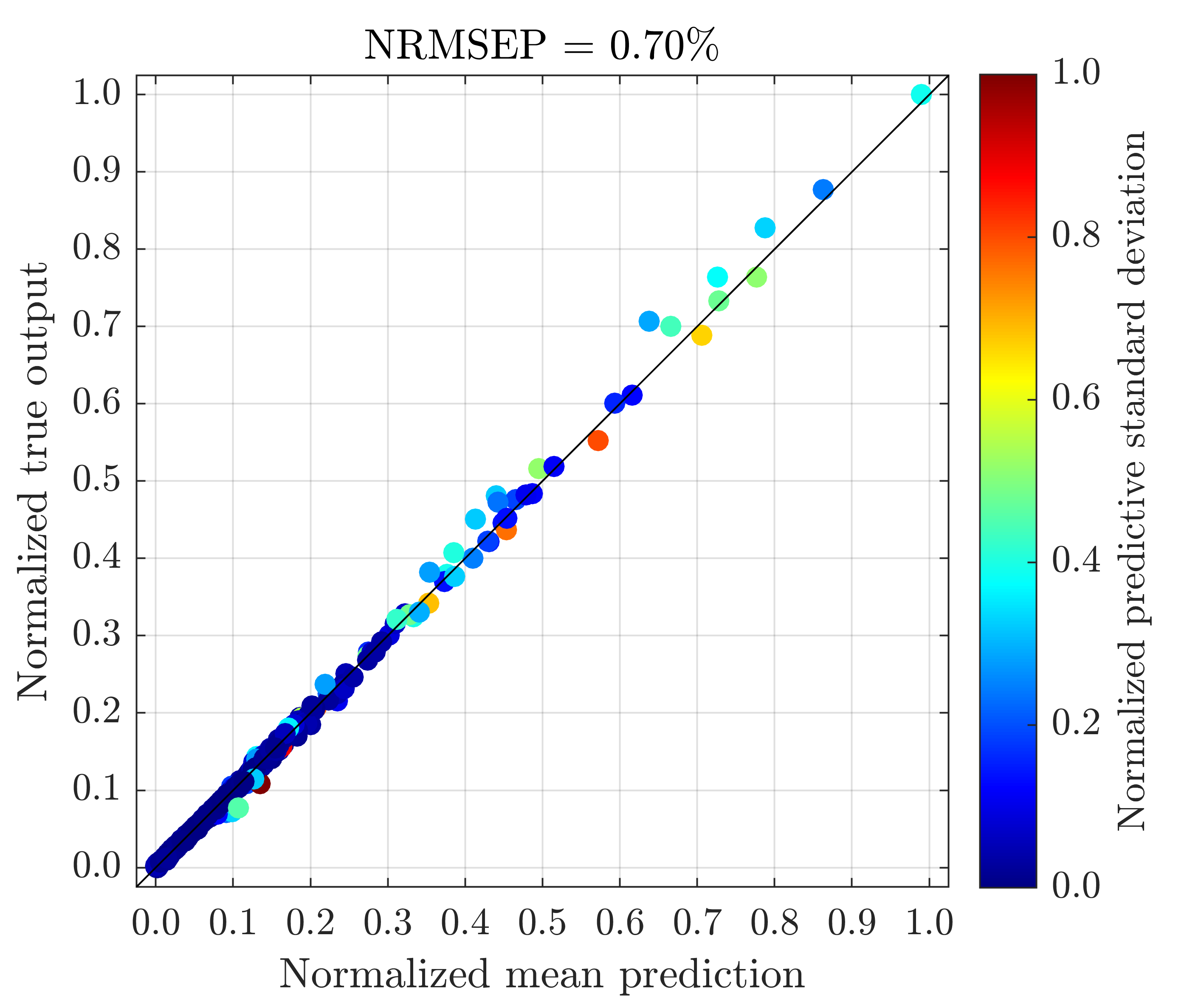}}
\subfloat[3-layer (SI-IC)]{\label{fig:engine_si3_con}\includegraphics[width=0.25\linewidth]{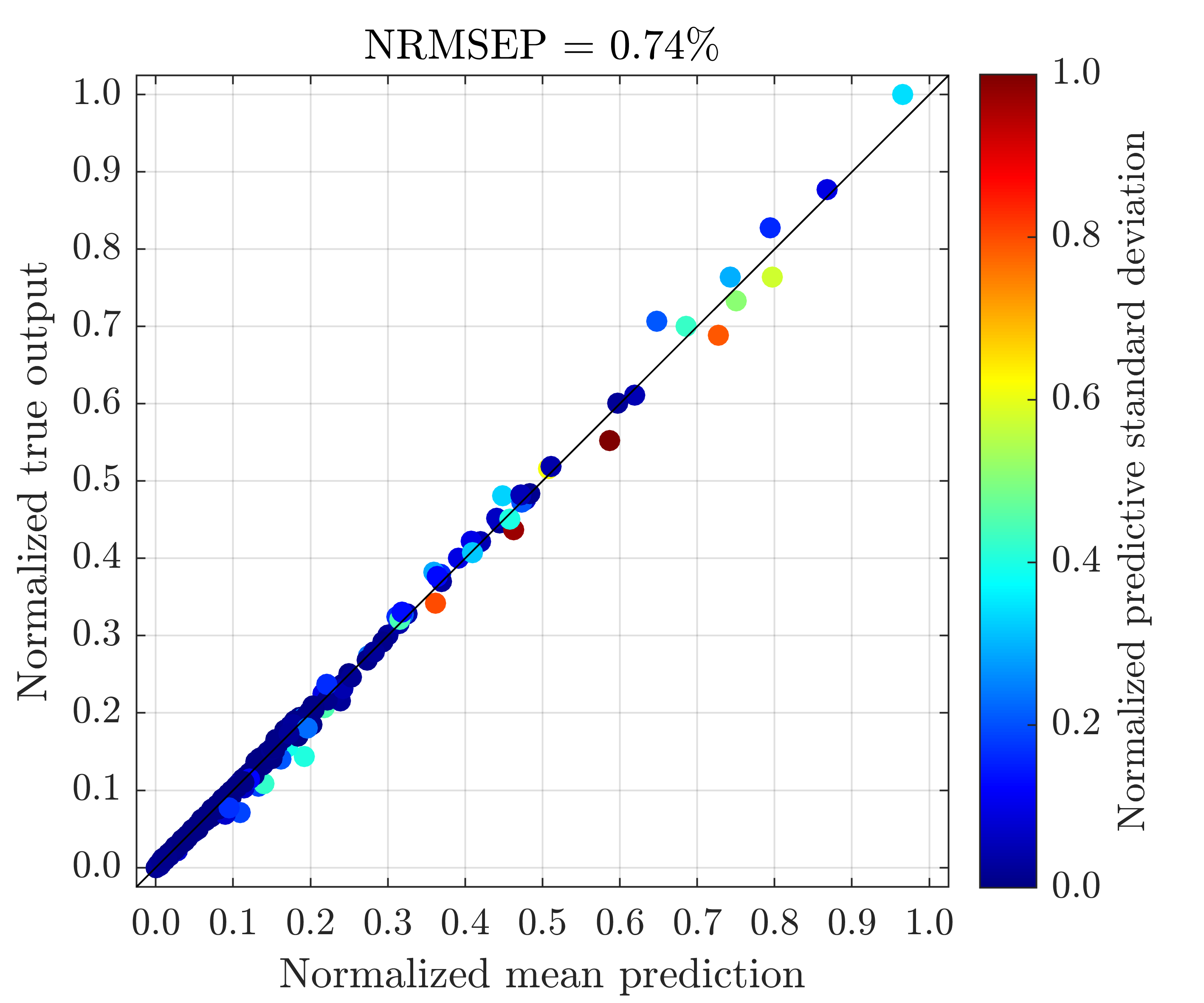}}\\
\subfloat[GP]{\label{fig:engine_gp}\includegraphics[width=0.25\linewidth]{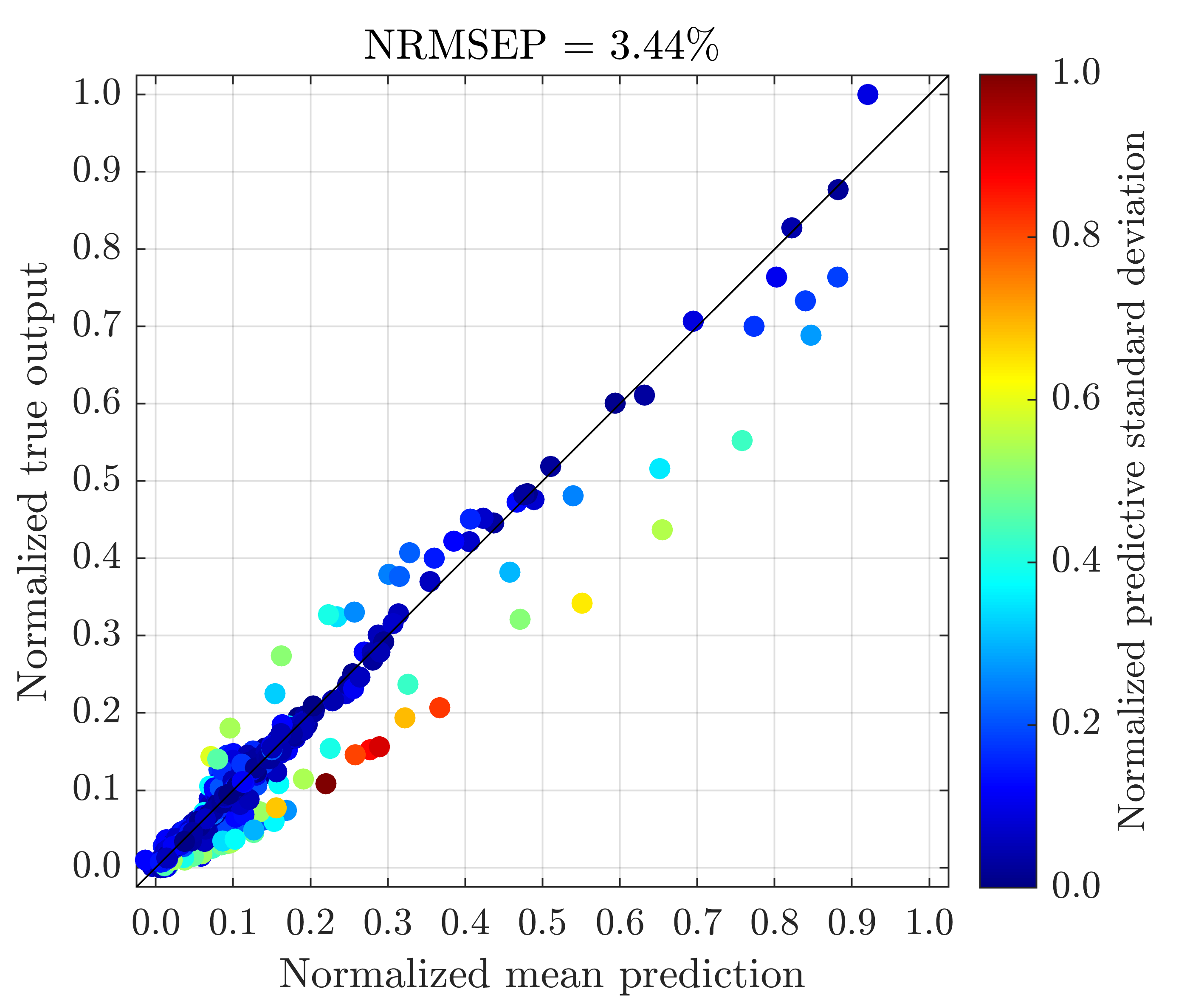}} 
\subfloat[4-layer (DSVI)]{\label{fig:engine_v4}\includegraphics[width=0.25\linewidth]{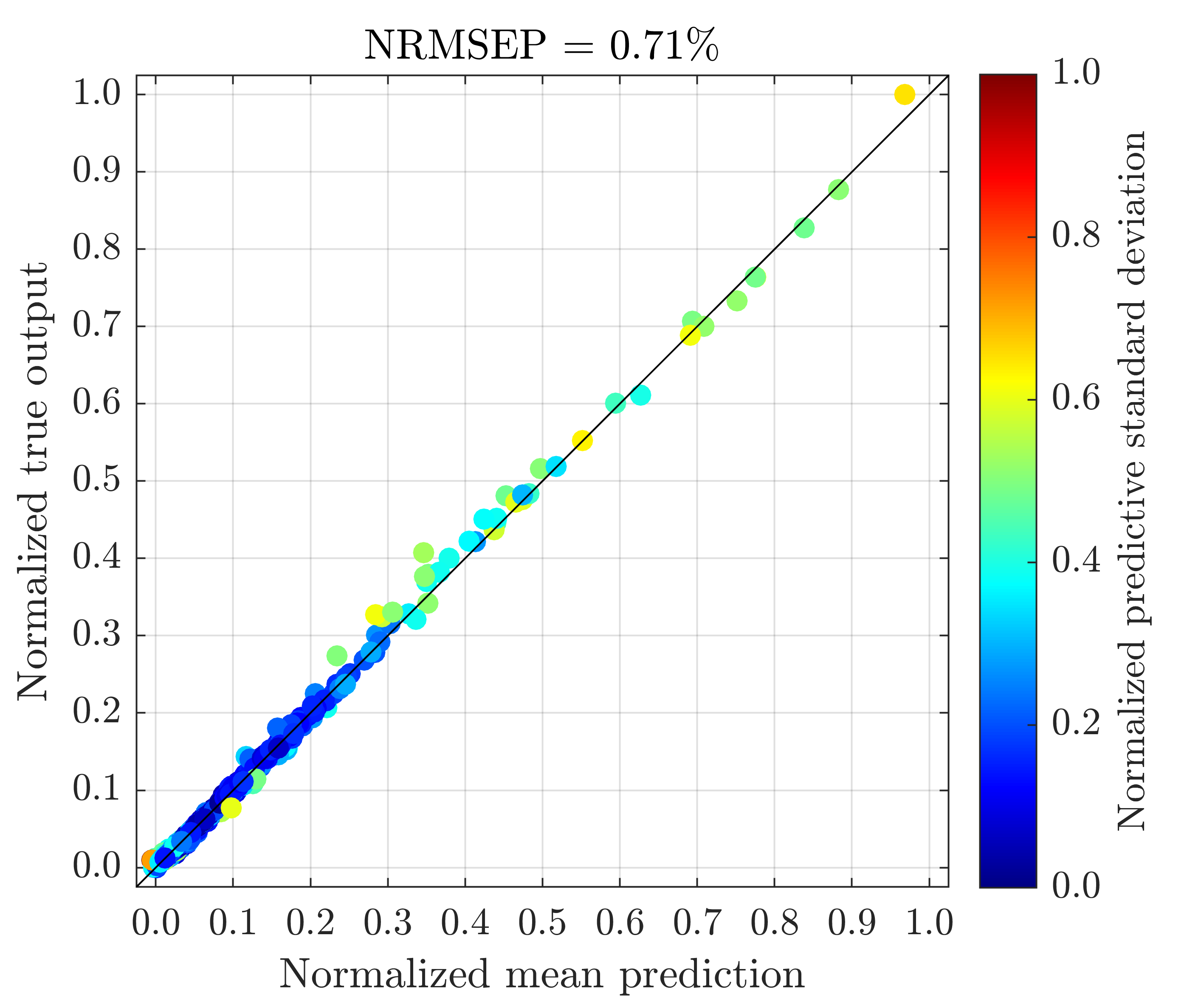}}
\subfloat[4-layer (SI)]{\label{fig:engine_si4}\includegraphics[width=0.25\linewidth]{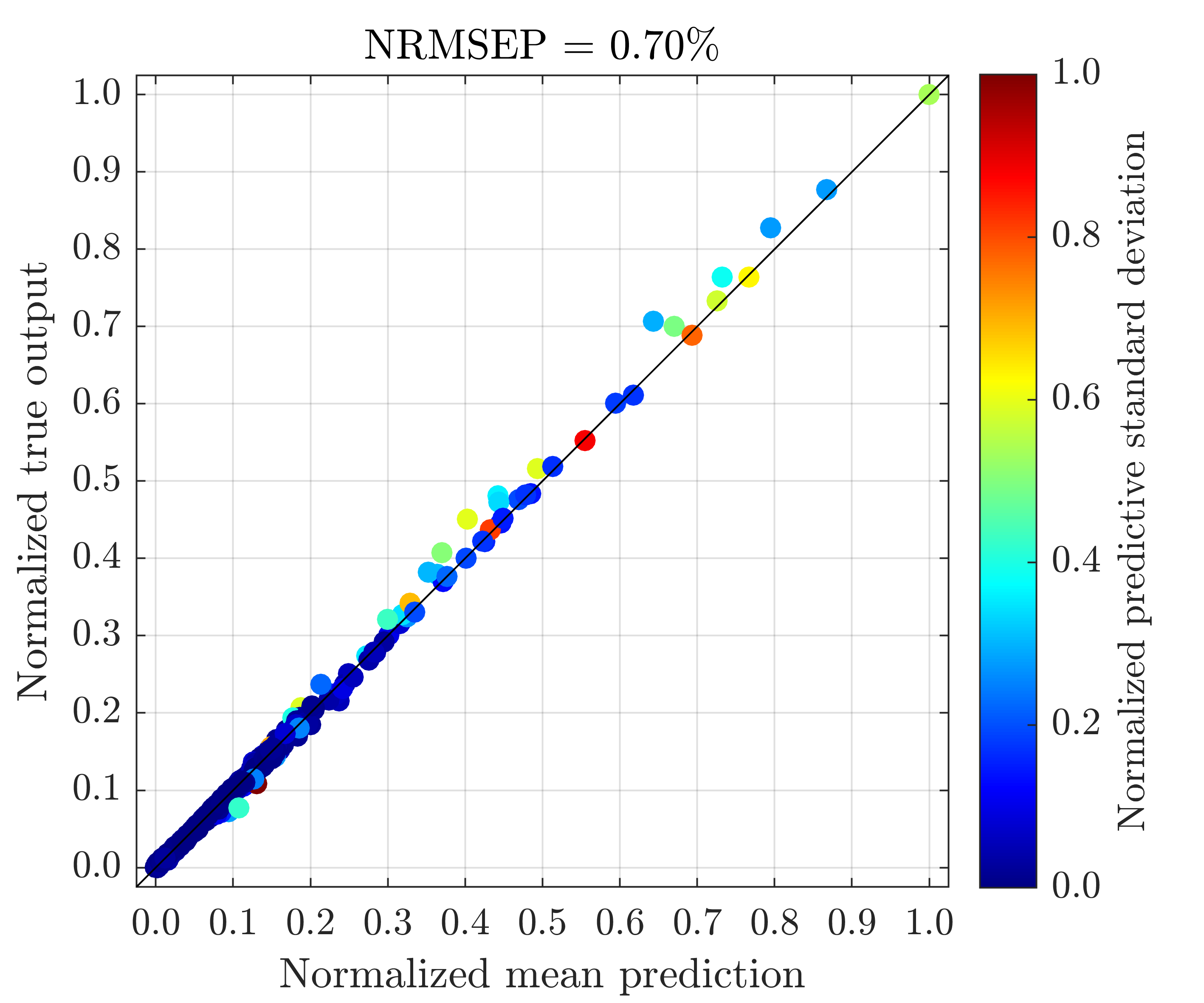}}
\subfloat[4-layer (SI-IC)]{\label{fig:engine_si4_con}\includegraphics[width=0.25\linewidth]{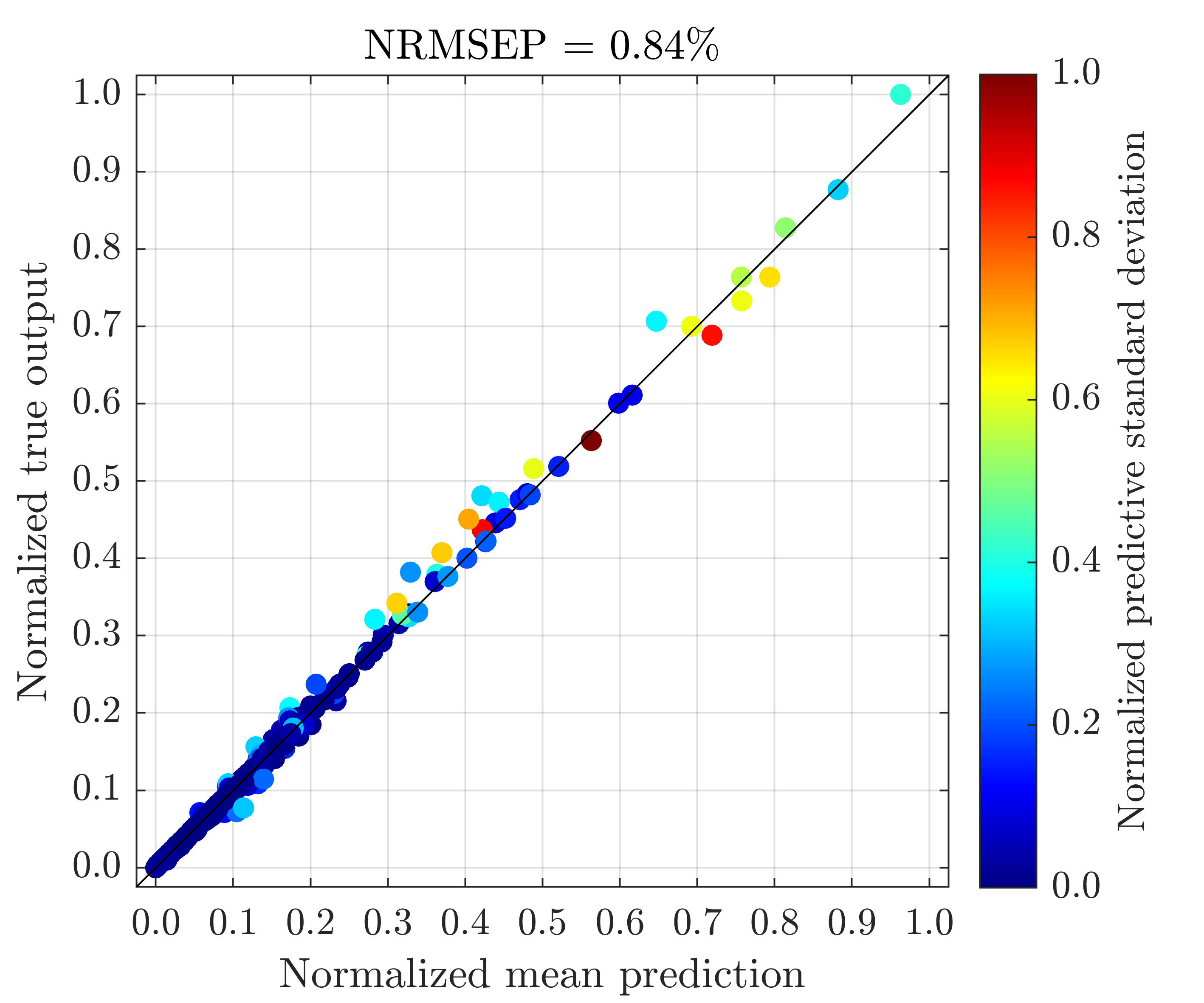}}
\caption{Plots of true TSFC outputs (normalized by their max and min values) of the engine model at $500$ testing positions \emph{vs} the mean predictions (normalized by the max and min values of true TSFC), along with predictive standard deviations (normalized by their max and min values), made by the best emulator (with the lowest NRMSEP out of $40$ inference trials) produced by FB, DSVI, SI and SI-IC. \emph{GP} represents a trained conventional GP emulator.}
\label{fig:engine_exp_diag}
\end{figure}

Figure~\ref{fig:engine_exp_diag} gives more detailed uncertainty profiles of best DGP emulators from different formations and inference approaches. It can be seen that the conventional GP emulator underestimates uncertainty at input positions where TSFC has large values and presents rapid increases. In contrast, most DGP emulators (particularly those from SI and SI-IC) highlight with much higher predictive standard deviations the input region where values of TSFC exhibits more variations, giving trustworthy information on input positions where additional data from the engine model can be obtained to improve the emulation performance. 

\section{Results for Delta}
\label{sec:delta}
\begin{figure}[!ht]
\centering 
\subfloat[NRMSEP]{\label{fig:delta_rmse}\includegraphics[width=0.45\linewidth]{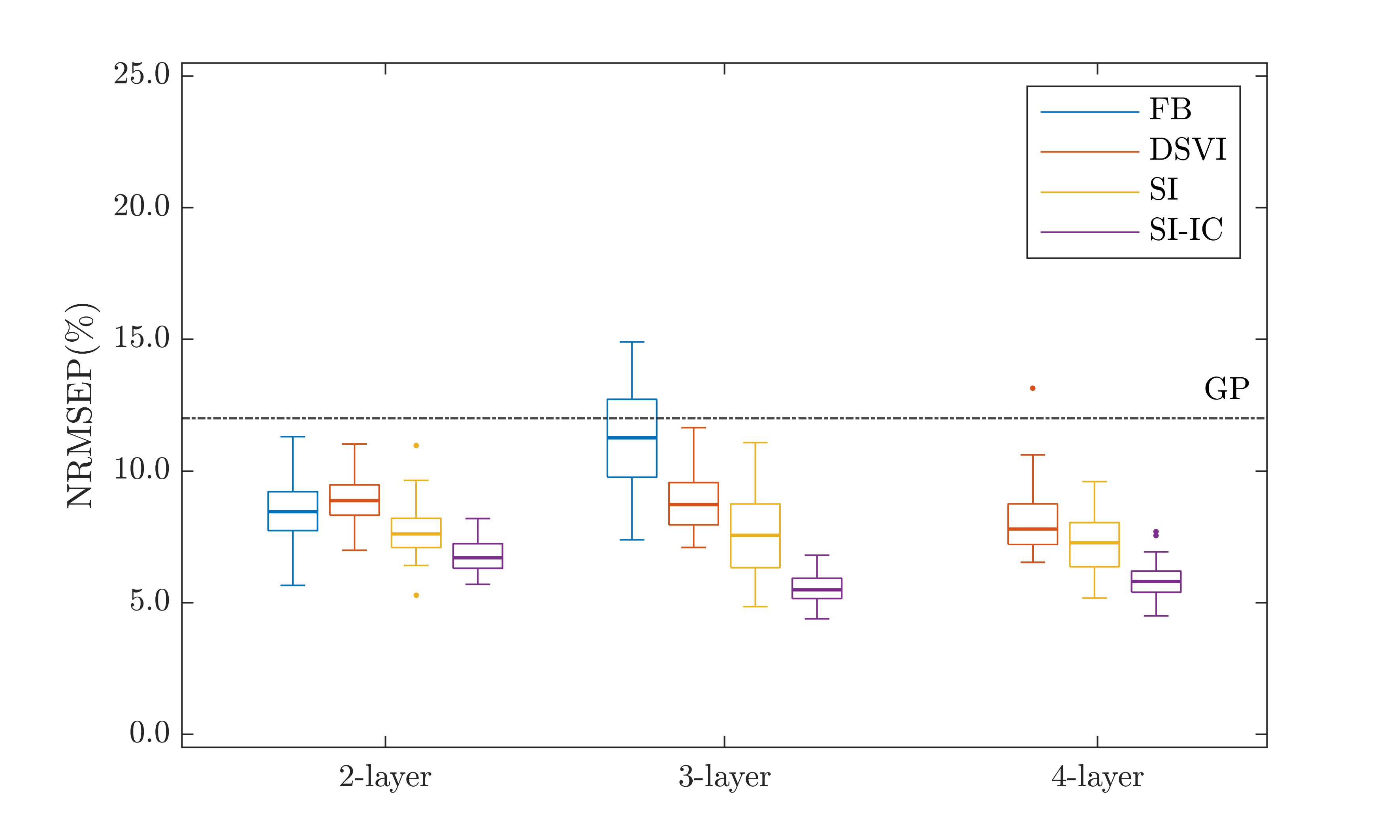}}
\subfloat[Computation time]{\label{fig:delta_time}\includegraphics[width=0.45\linewidth]{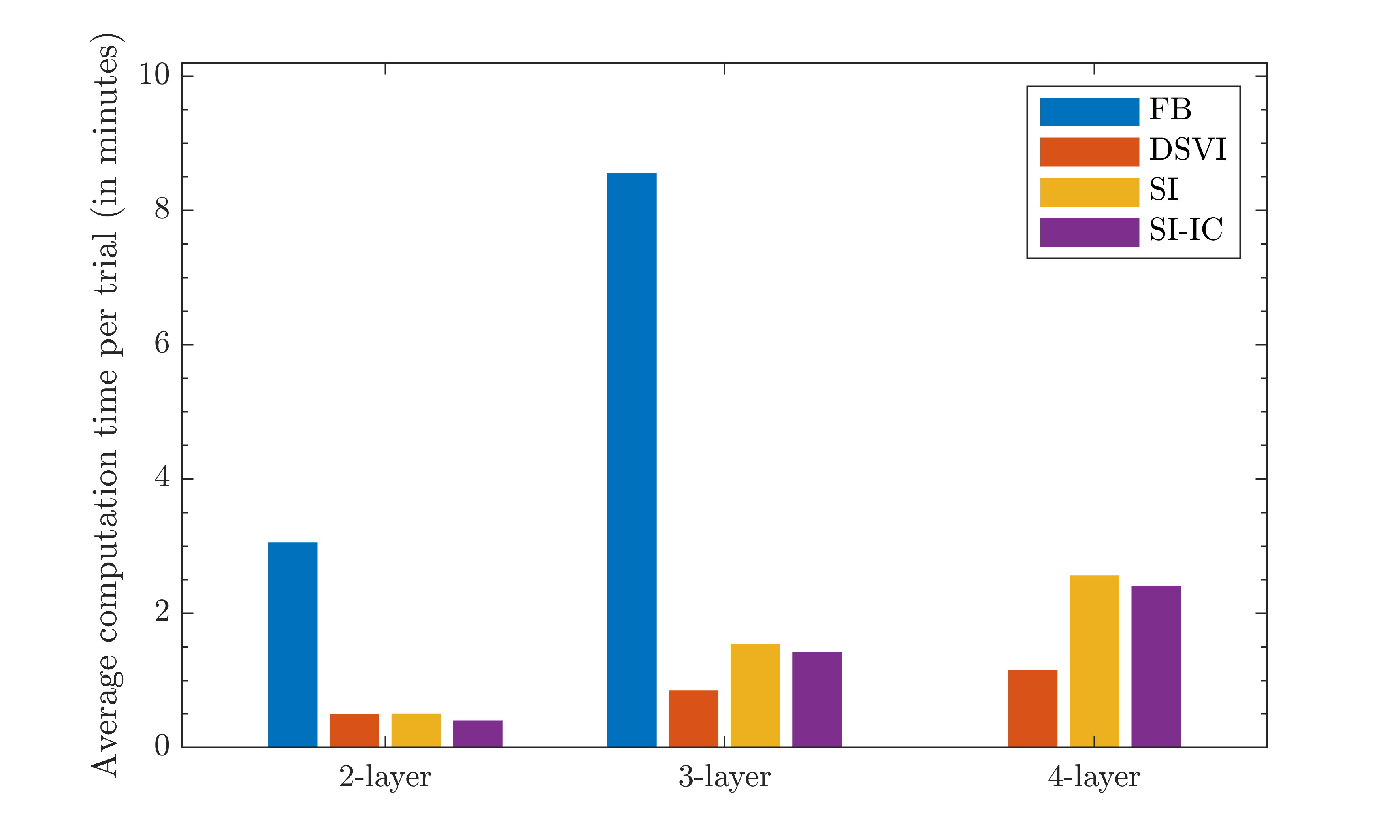}}
\caption{Comparison of FB, DSVI, SI, and SI-IC for $40$ repeatedly trained DGP emulators (i.e., $40$ inference trials) of Delta ($\Delta_t$) from the Heston model. The dash-dot line represents the NRMSEP of a trained conventional GP emulator.}
\label{fig:delta_rmse_time}
\end{figure}

\begin{figure}[!ht]
\centering 
\subfloat[2-layer (FB)]{\label{fig:delta_fb2}\includegraphics[width=0.25\linewidth]{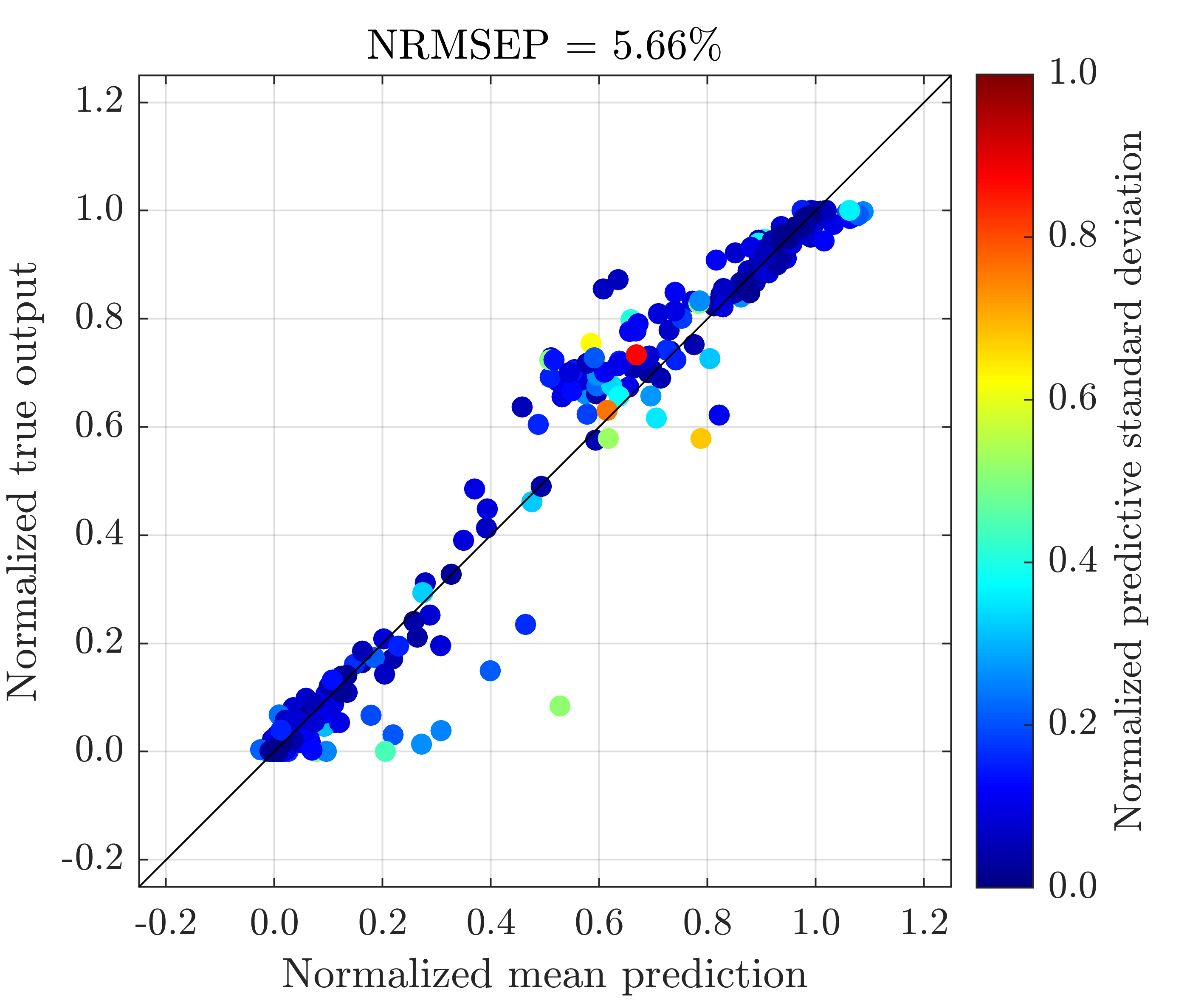}}
\subfloat[2-layer (DSVI)]{\label{fig:delta_vi2}\includegraphics[width=0.25\linewidth]{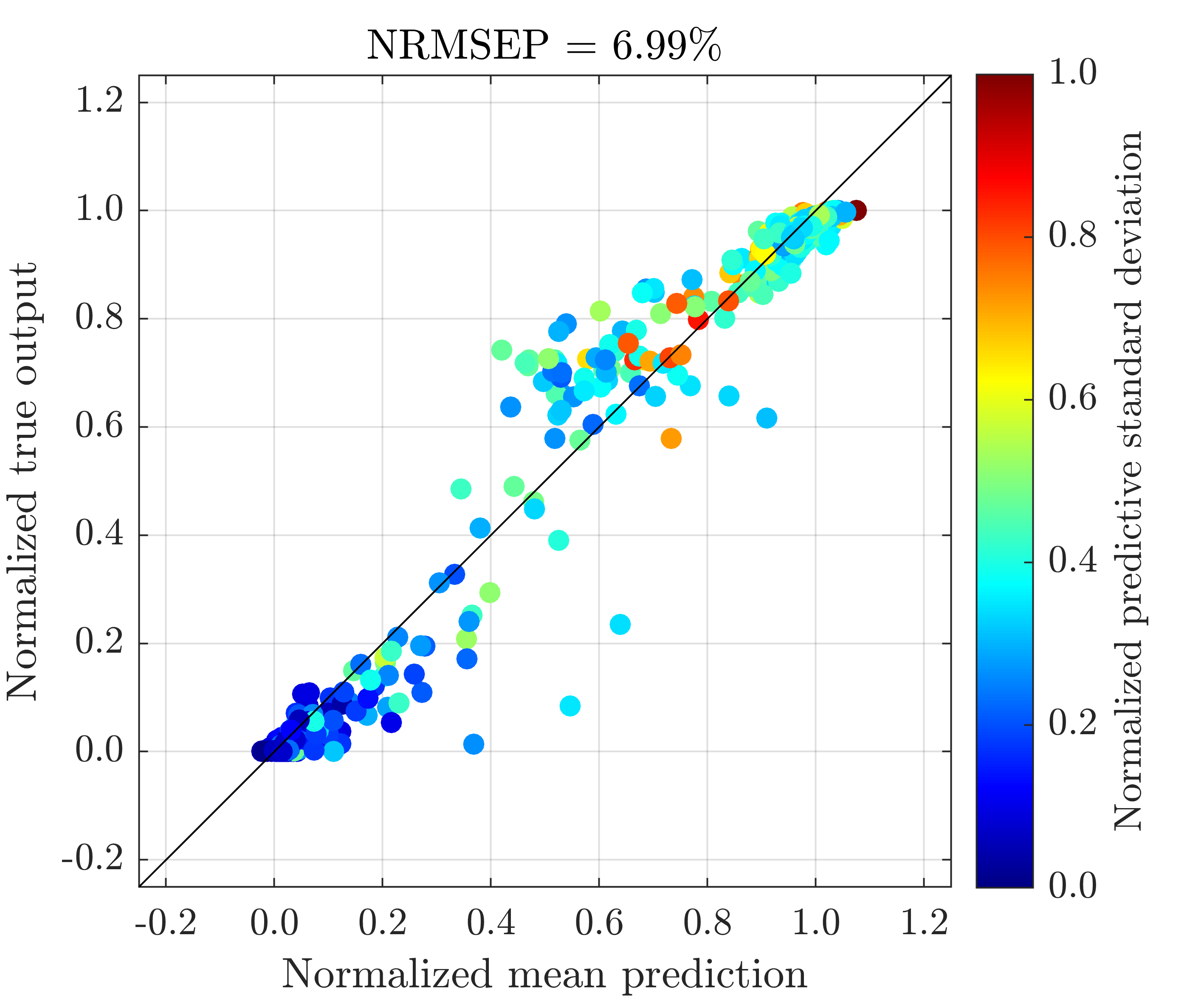}}
\subfloat[2-layer (SI)]{\label{fig:delta_si2}\includegraphics[width=0.25\linewidth]{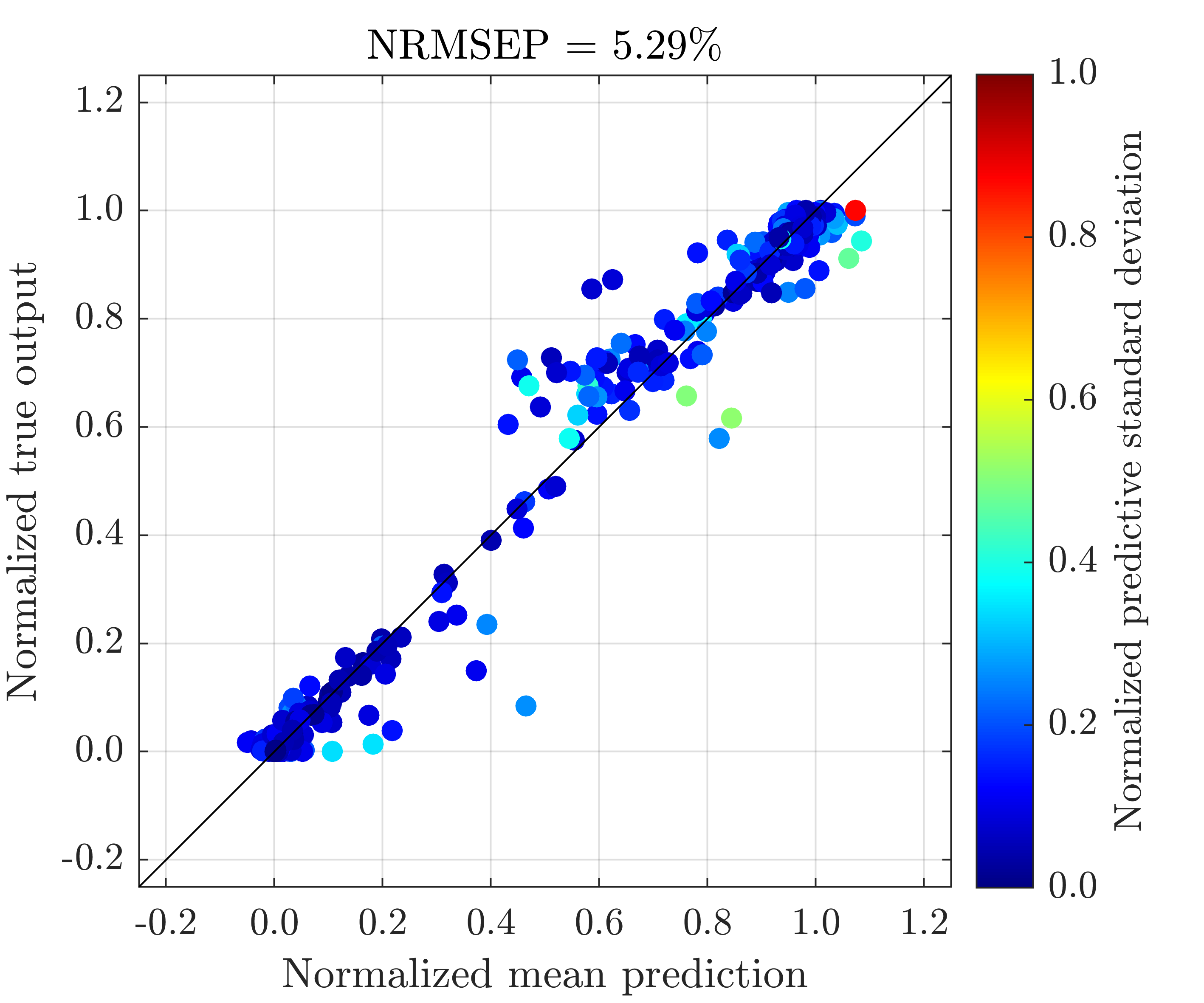}}
\subfloat[2-layer (SI-IC)]{\label{fig:delta_si2_con}\includegraphics[width=0.25\linewidth]{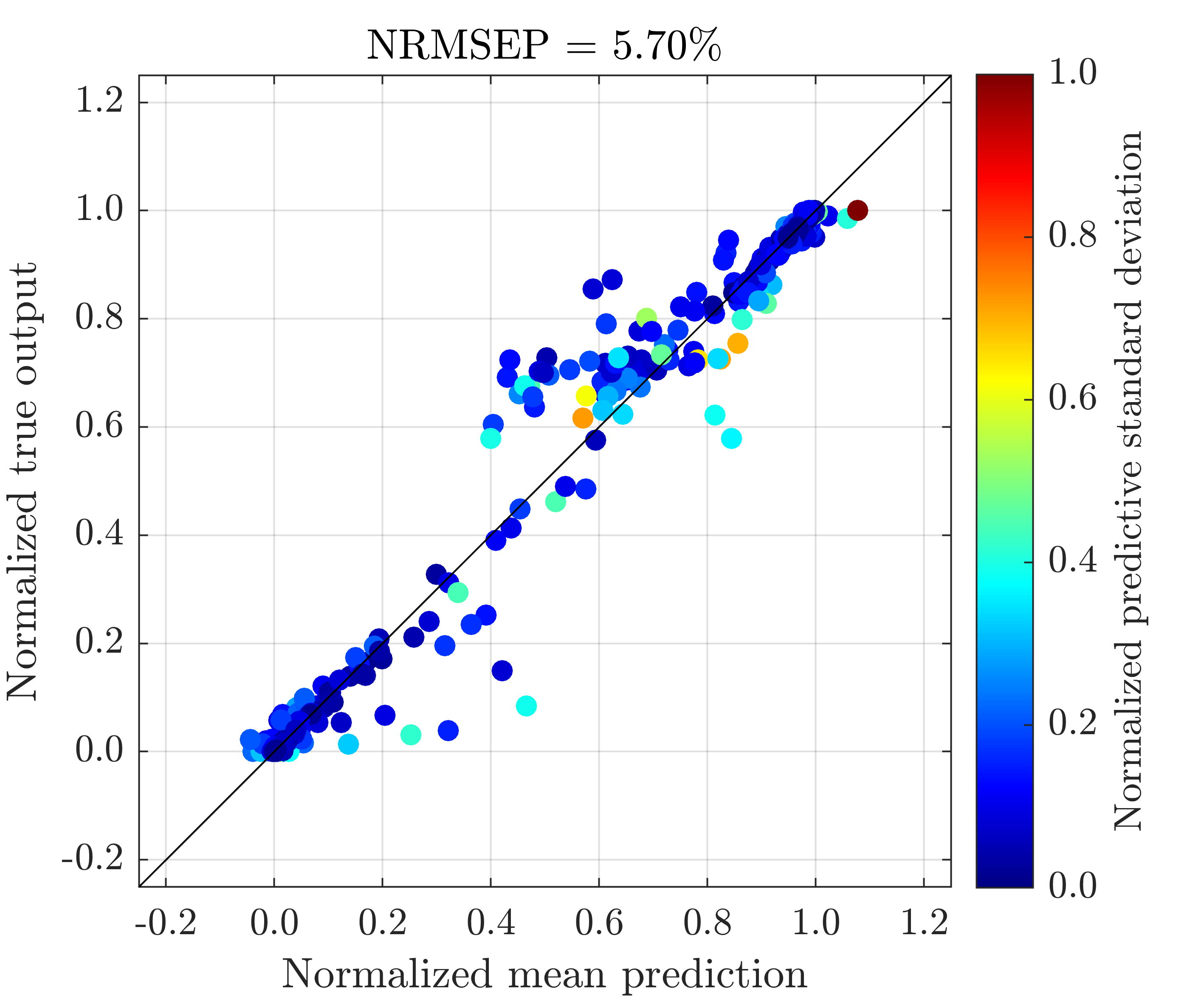}}\\
\subfloat[3-layer (FB)]{\label{fig:delta_fb3}\includegraphics[width=0.25\linewidth]{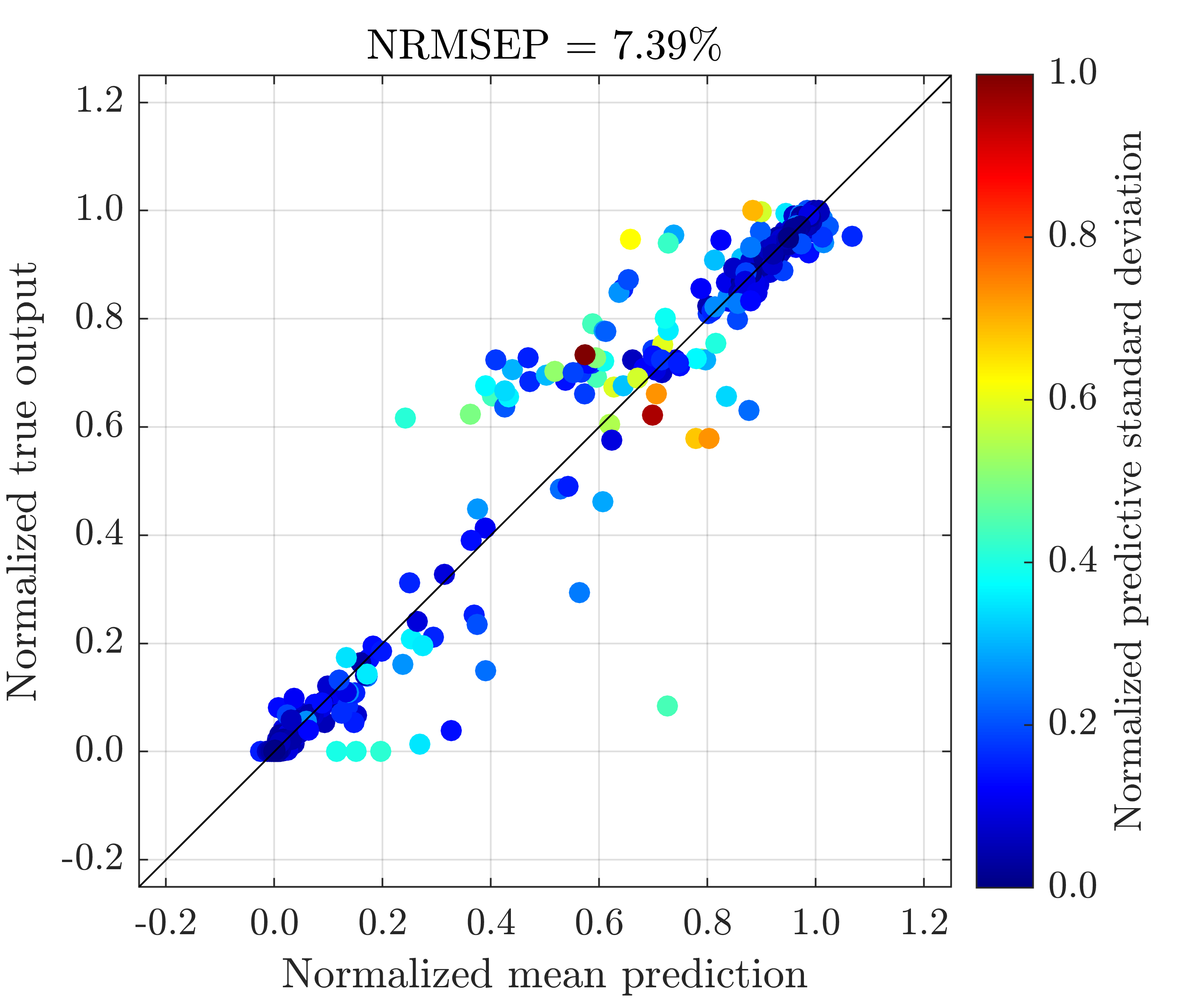}}
\subfloat[3-layer (DSVI)]{\label{fig:delta_vi3}\includegraphics[width=0.25\linewidth]{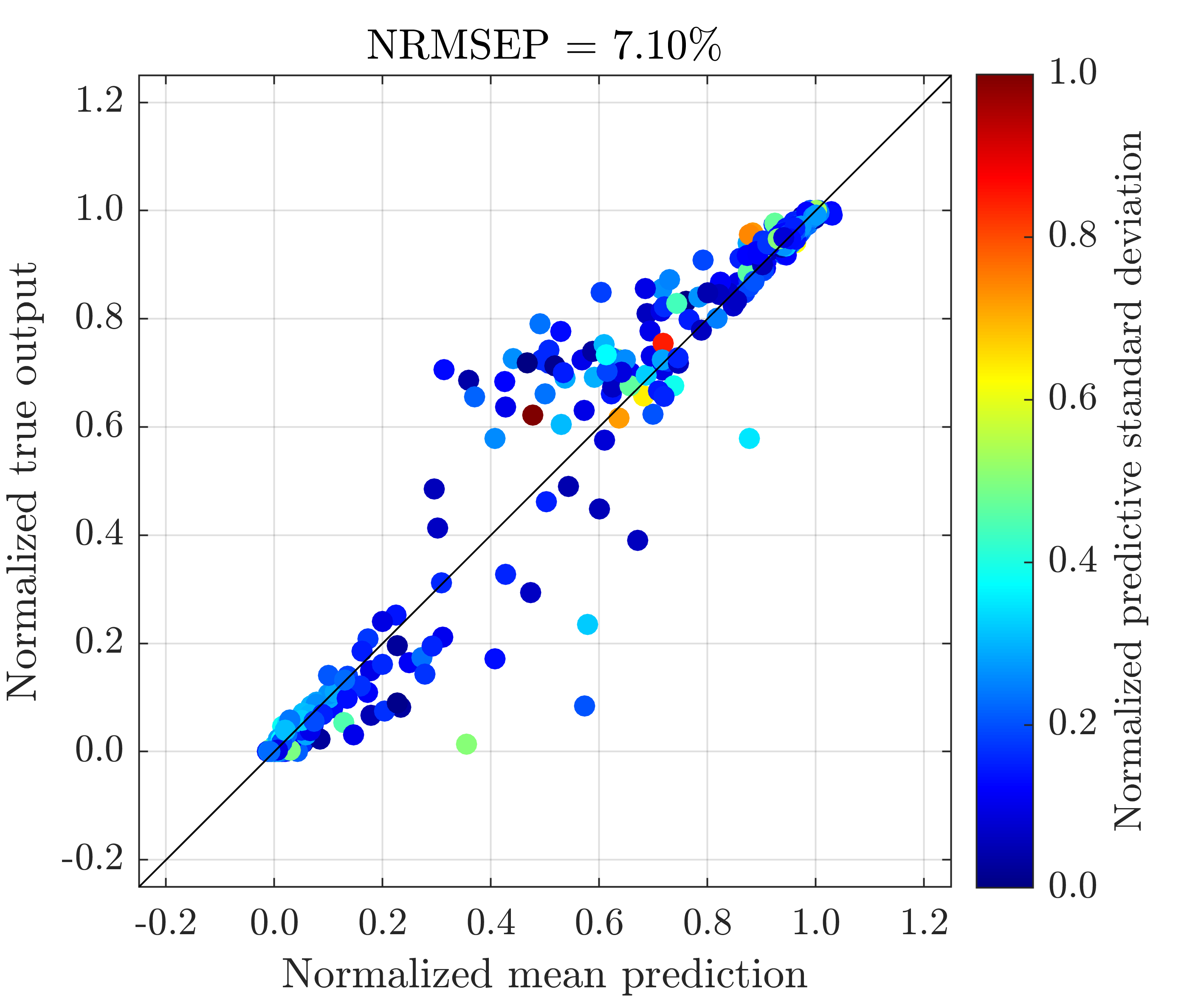}}
\subfloat[3-layer (SI)]{\label{fig:delta_si3}\includegraphics[width=0.25\linewidth]{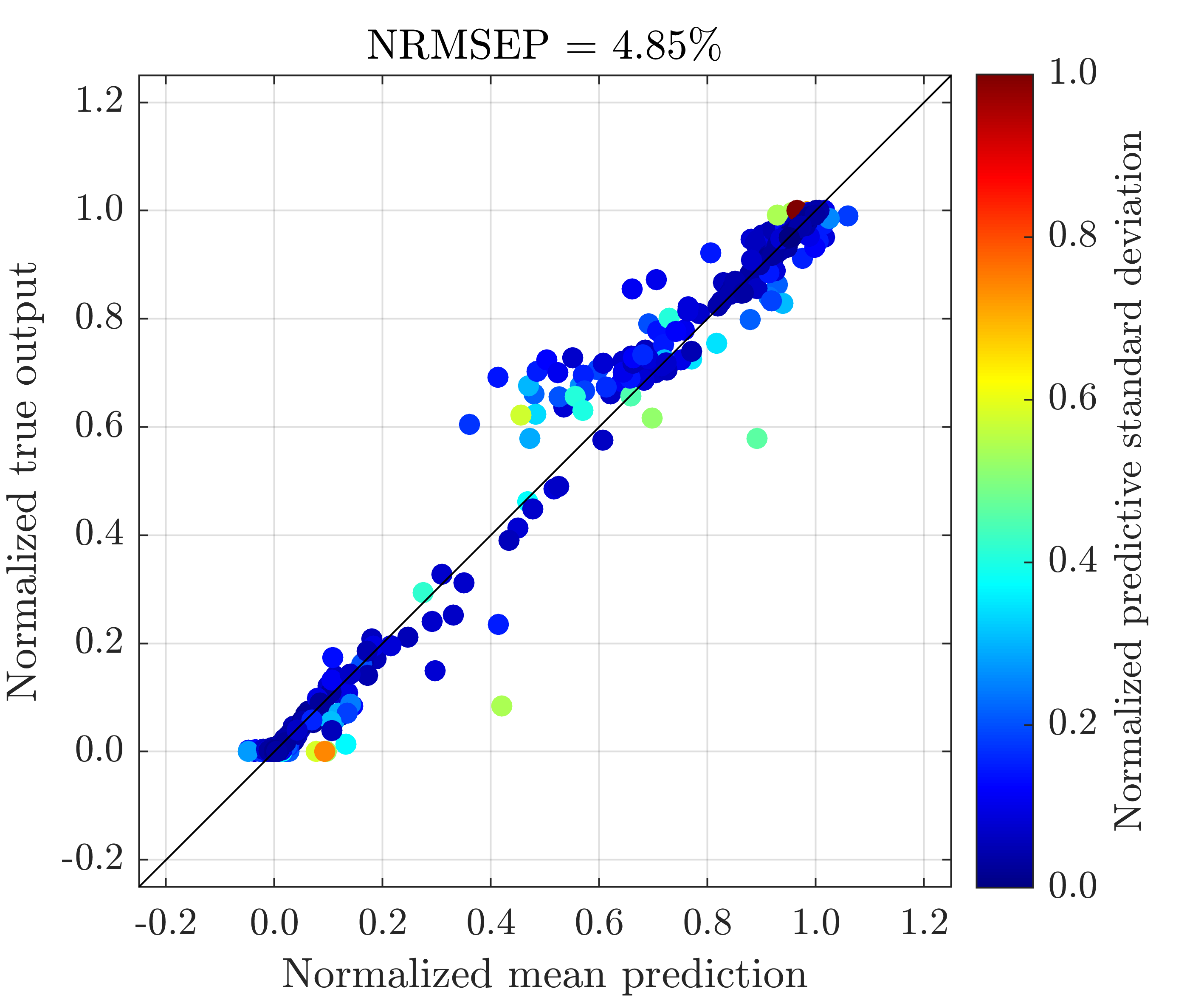}}
\subfloat[3-layer (SI-IC)]{\label{fig:delta_si3_con}\includegraphics[width=0.25\linewidth]{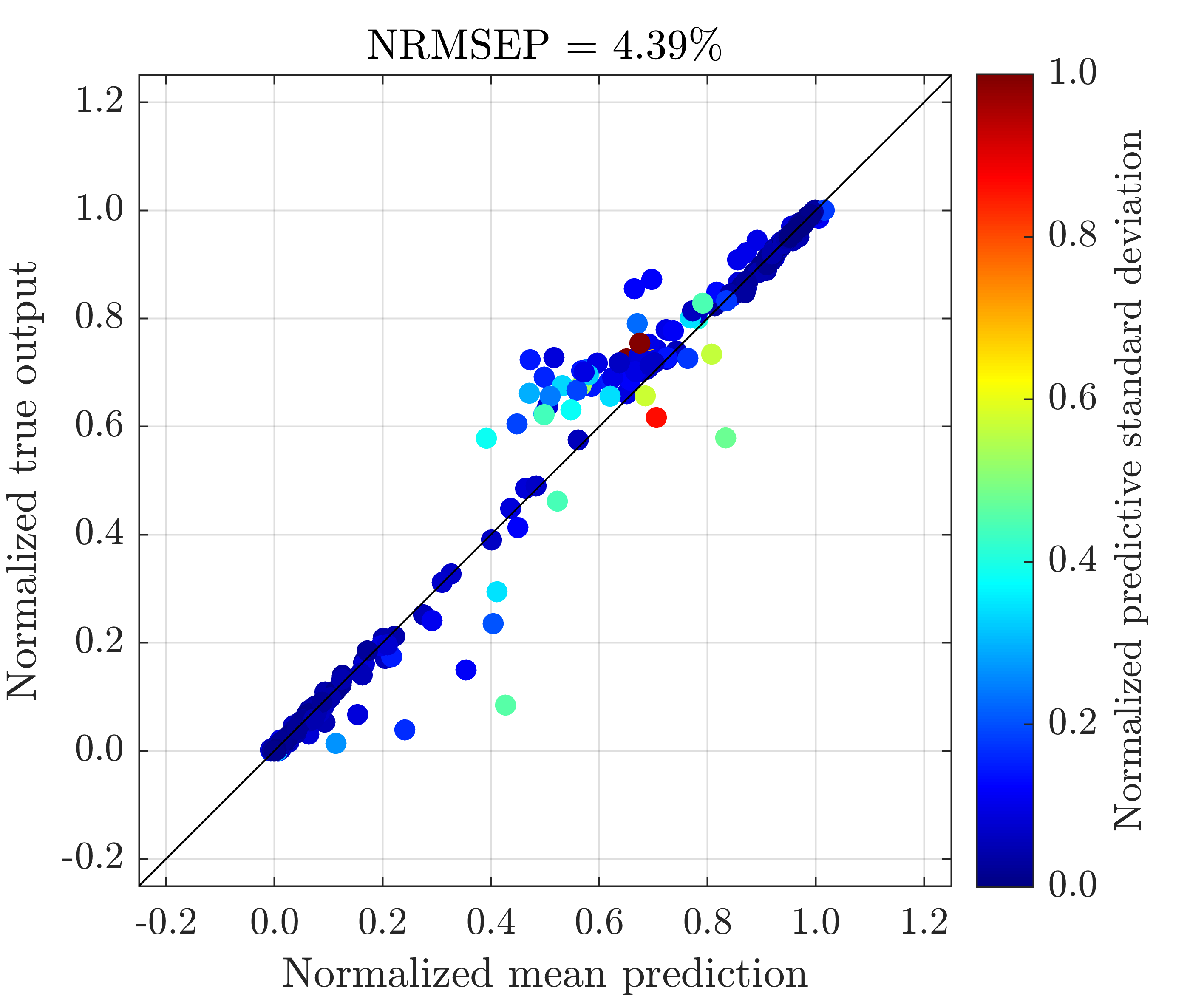}}\\
\subfloat[GP]{\label{fig:delta_gp}\includegraphics[width=0.25\linewidth]{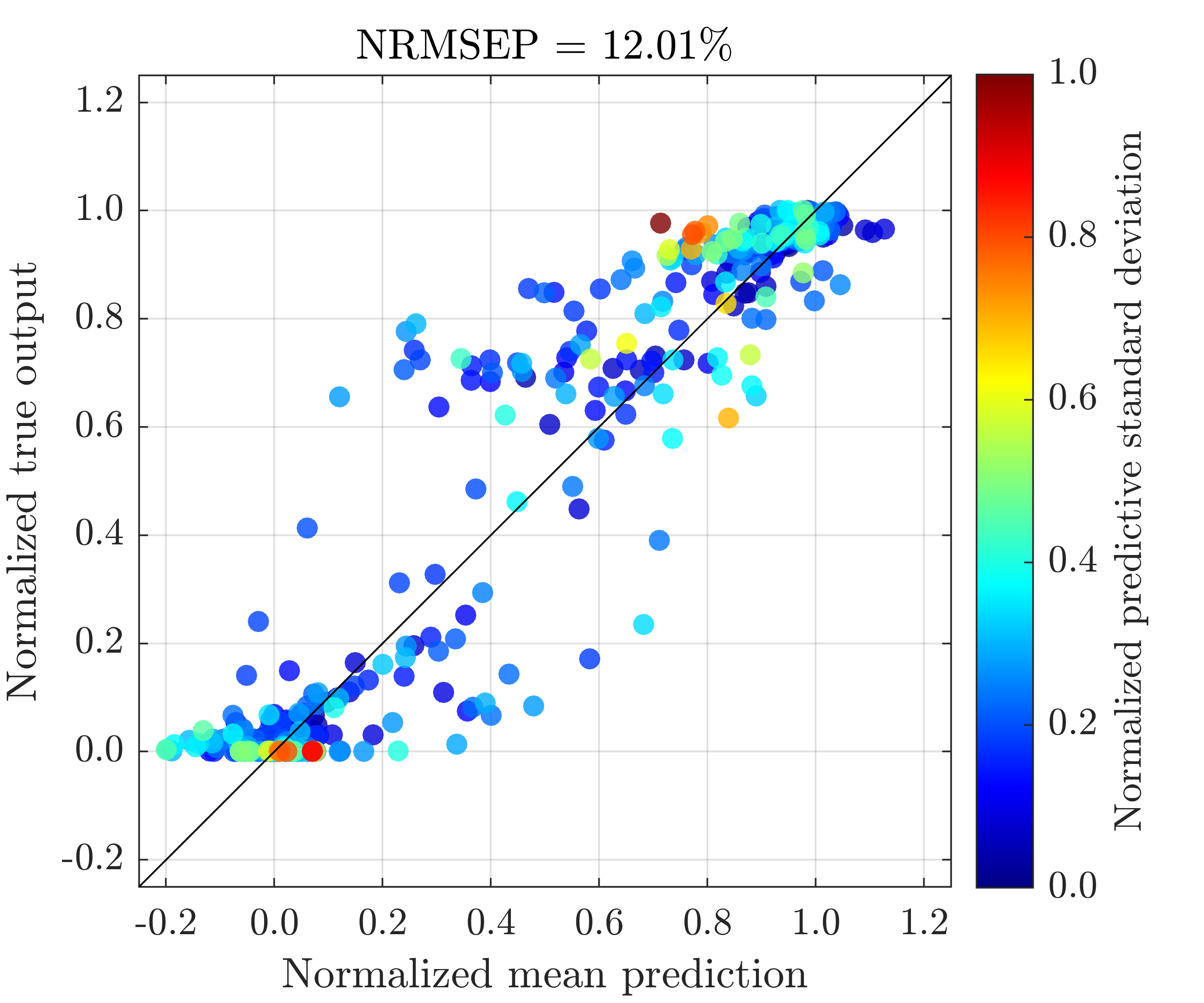}} 
\subfloat[4-layer (DSVI)]{\label{fig:delta_v4}\includegraphics[width=0.25\linewidth]{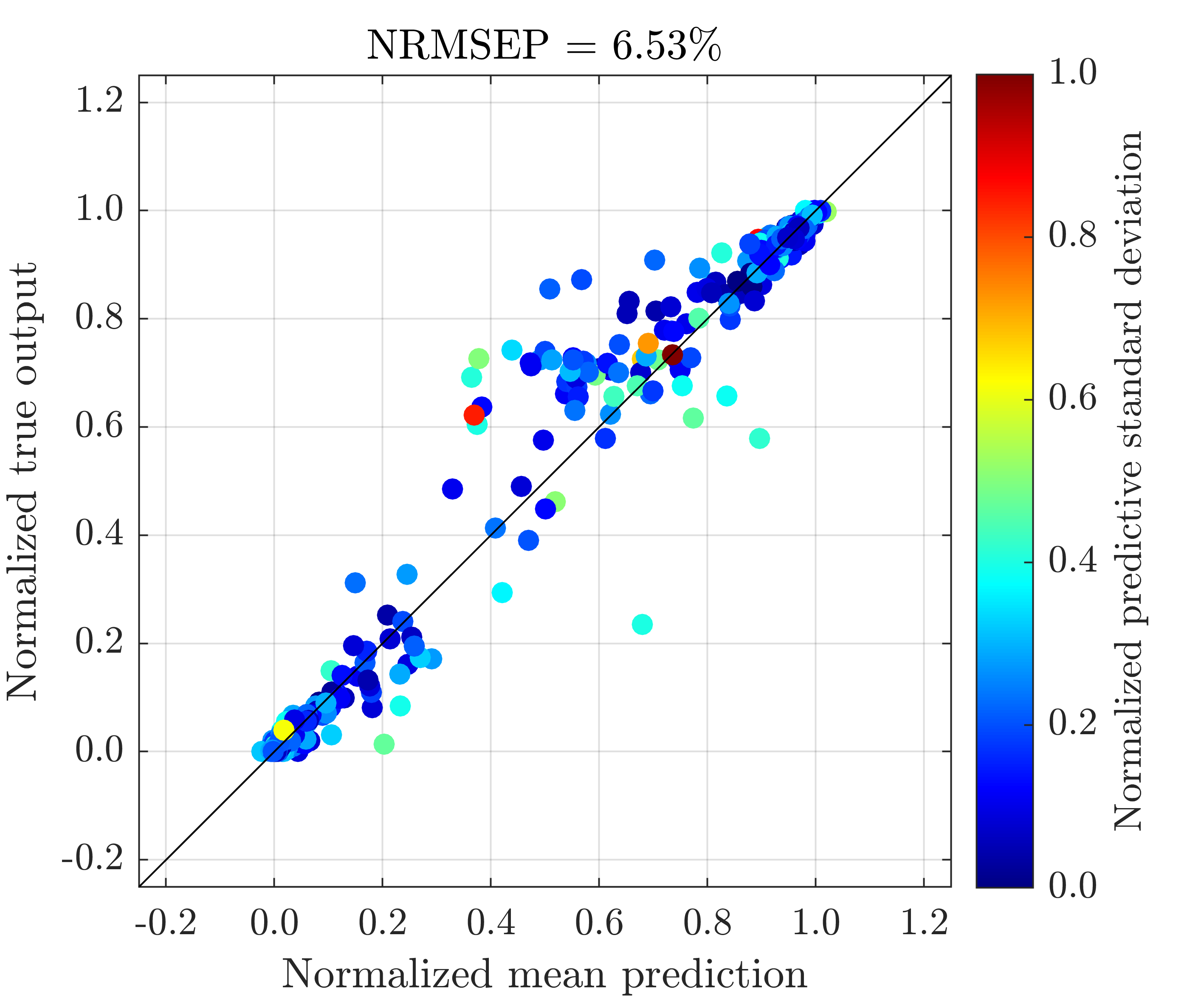}}
\subfloat[4-layer (SI)]{\label{fig:delta_si4}\includegraphics[width=0.25\linewidth]{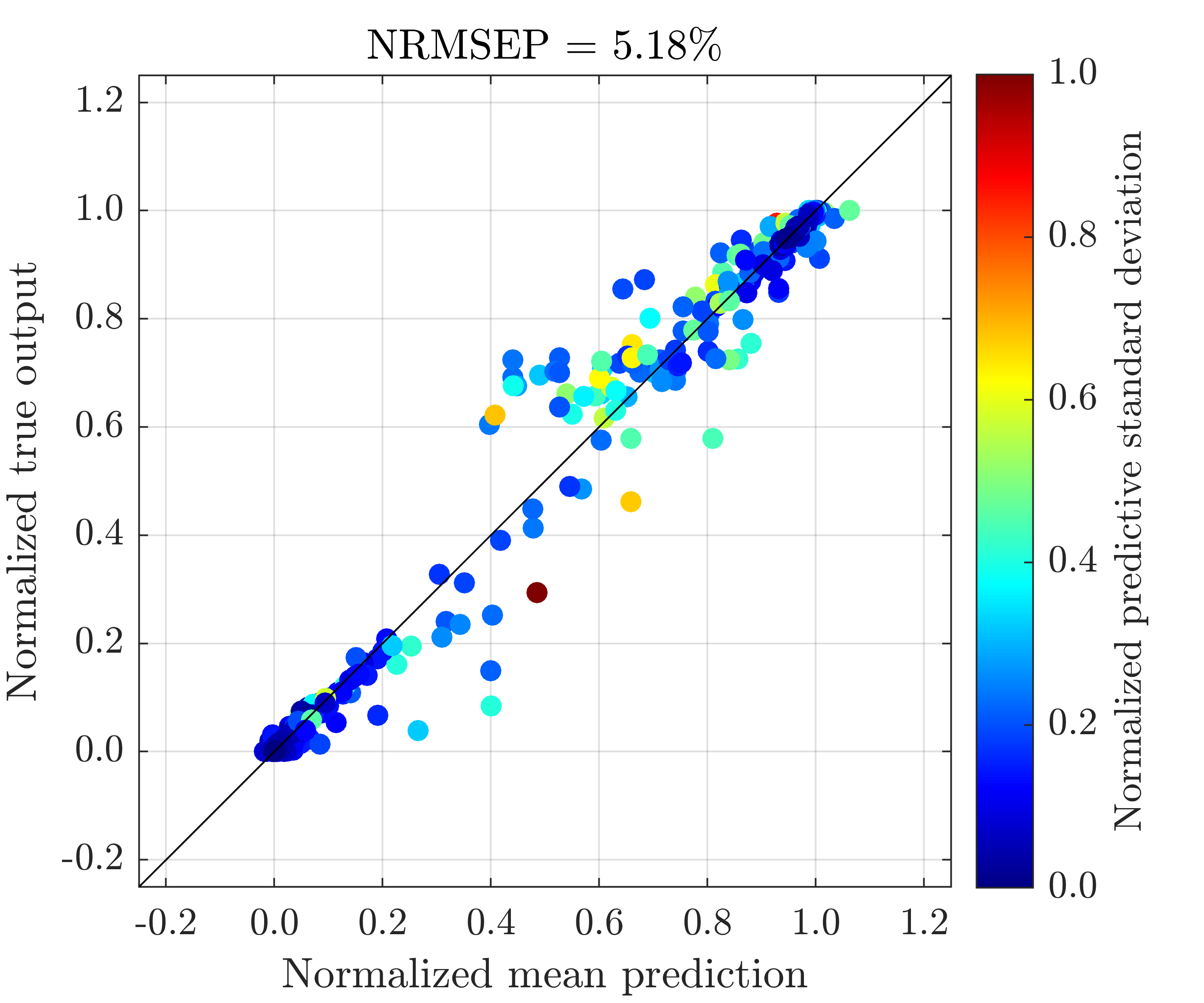}}
\subfloat[4-layer (SI-IC)]{\label{fig:delta_si4_con}\includegraphics[width=0.25\linewidth]{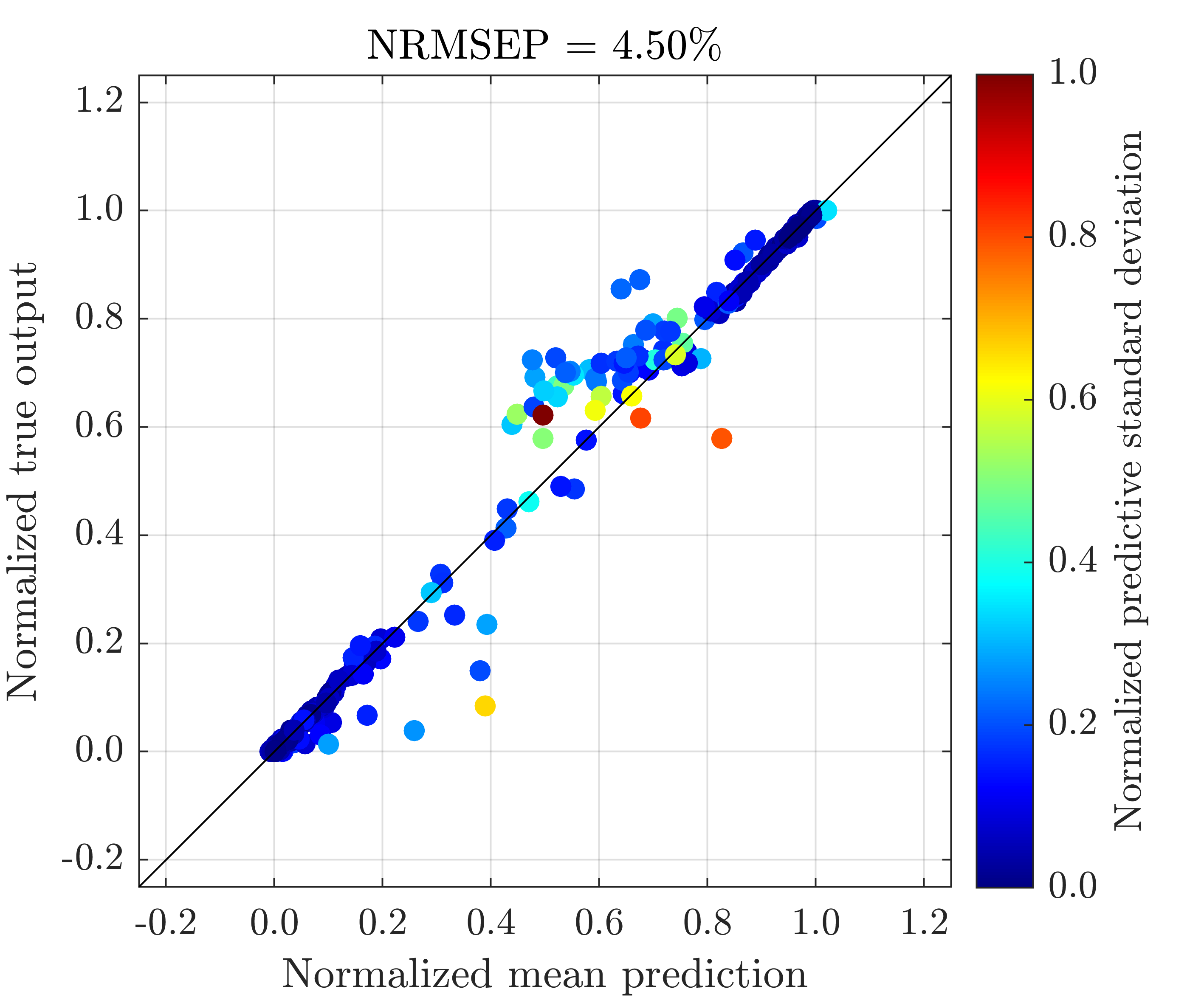}}
\caption{Plots of numerical solutions of Delta ($\Delta_t$) (normalized by their max and min values) from the Heston model at $500$ testing positions \emph{vs} the mean predictions (normalized by the max and min values of numerical solutions of Delta), along with predictive standard deviations (normalized by their max and min values), made by the best emulator (with the lowest NRMSEP out of $40$ inference trials) produced by FB, DSVI, SI and SI-IC. \emph{GP} represents a conventional GP emulator.}
\label{fig:delta_diag}
\end{figure}

From Figure~\subref*{fig:delta_rmse}, we can draw similar conclusions to those for Vega in Section~\ref{sec:heston} of the manuscript. SI-IC produces DGP emulators with the overall best performance in terms of mean predictions and inference stability.    
The uncertainty profiles in Figure~\ref{fig:delta_diag} clearly demonstrate that uncertainties quantified by the conventional GP emulator fail to distinguish input regions where Delta has little variations (i.e., input locations that correspond to OTM options with Delta values close to zeros and ITM options with Delta values close to ones) from areas where Delta has large variations (i.e., input locations that correspond to ATM options with Delta values swing around $0.5$). DGP emulators in general address this issue by producing constantly small predictive standard deviations at test positions that correspond to OTM and ITM options. DGP emulators trained by SI appear to be over-confident at input positions corresponding to ATM options that have varying Delta values. The issue seems to be alleviated to some extent by using SI-IC since we observe more testing positions that correspond to ATM options have higher predictive standard deviations.

\section{Results for Gamma}
\label{sec:gamma}
\begin{figure}[!ht]
\centering 
\subfloat[NRMSEP]{\label{fig:gamma_rmse}\includegraphics[width=0.45\linewidth]{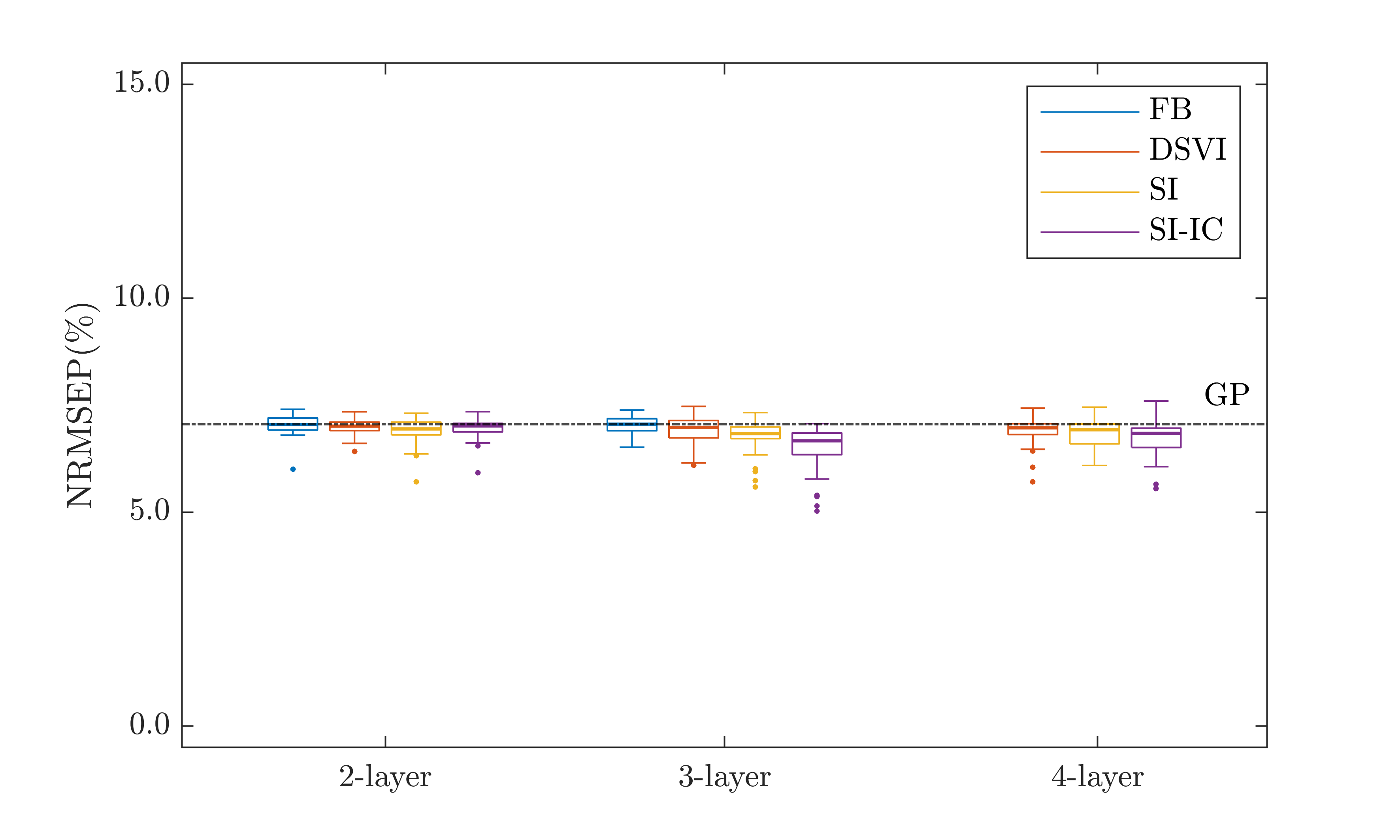}}
\subfloat[Computation time]{\label{fig:gamma_time}\includegraphics[width=0.45\linewidth]{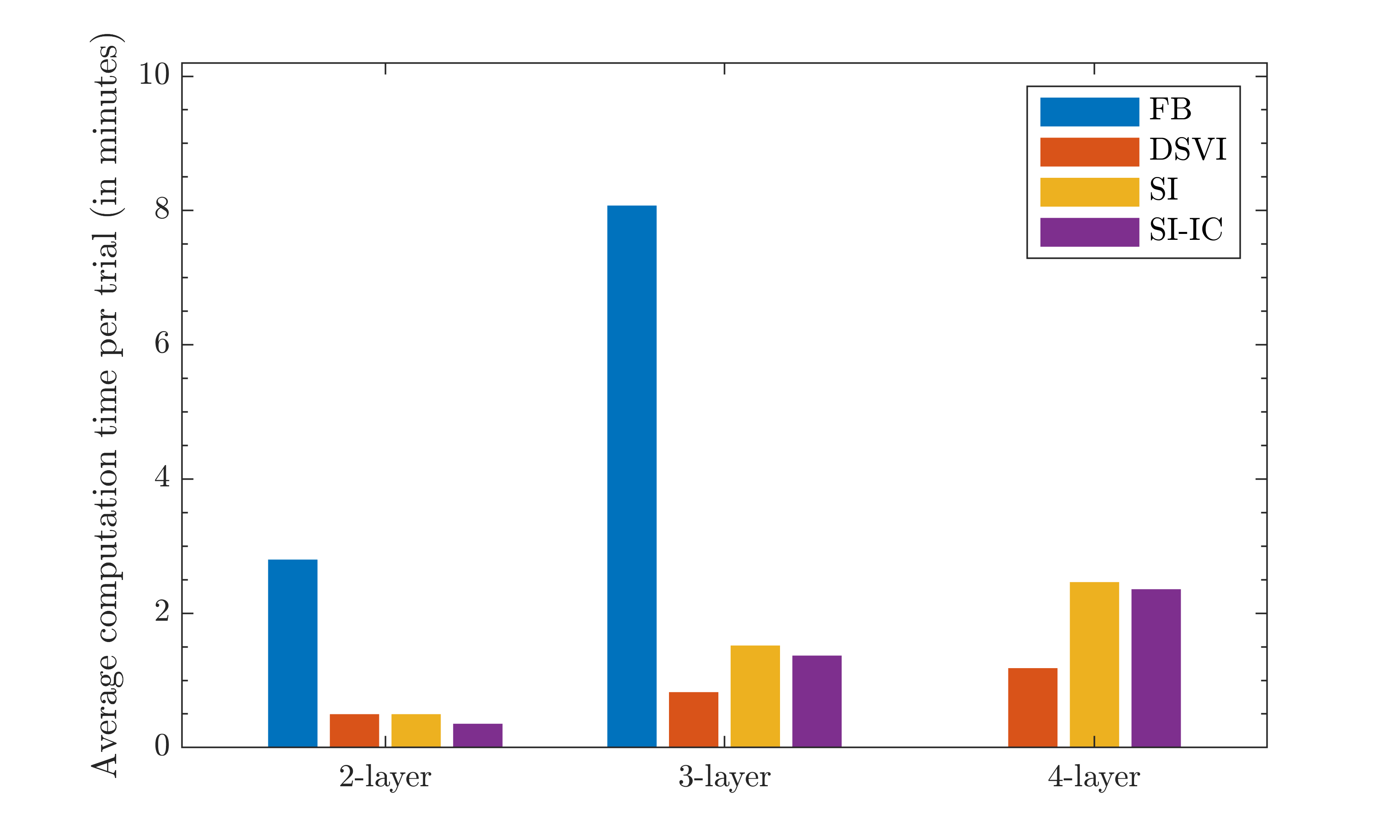}} 
\caption{Comparison of FB, DSVI, SI, and SI-IC for $40$ repeatedly trained DGP emulators (i.e., $40$ inference trials) of Gamma ($\Gamma_t$) from the Heston model. The dash-dot line represents the NRMSEP of a trained conventional GP emulator.}
\label{fig:gamma_rmse_time}
\end{figure}

From Figure~\subref*{fig:gamma_rmse}, we see that no DGP emulators, regardless of the inference approaches, present significantly better accuracy on mean predictions than the conventional GP. This result is not surprising since Gammas of OTM and ITM options are near zero while Gammas of ATM options exhibit spikes (see Figure~\subref*{fig:gamma} of the manuscript) that are difficult to be captured adequately by the training data generated by the LHS, which is a static space-filling design strategy. Even so, we observe better overall performances of three-layered DGP emulators trained by SI-IC than emulators trained by other approaches.  

\begin{figure}[!ht]
\centering 
\subfloat[2-layer (FB)]{\label{fig:gamma_fb2}\includegraphics[width=0.25\linewidth]{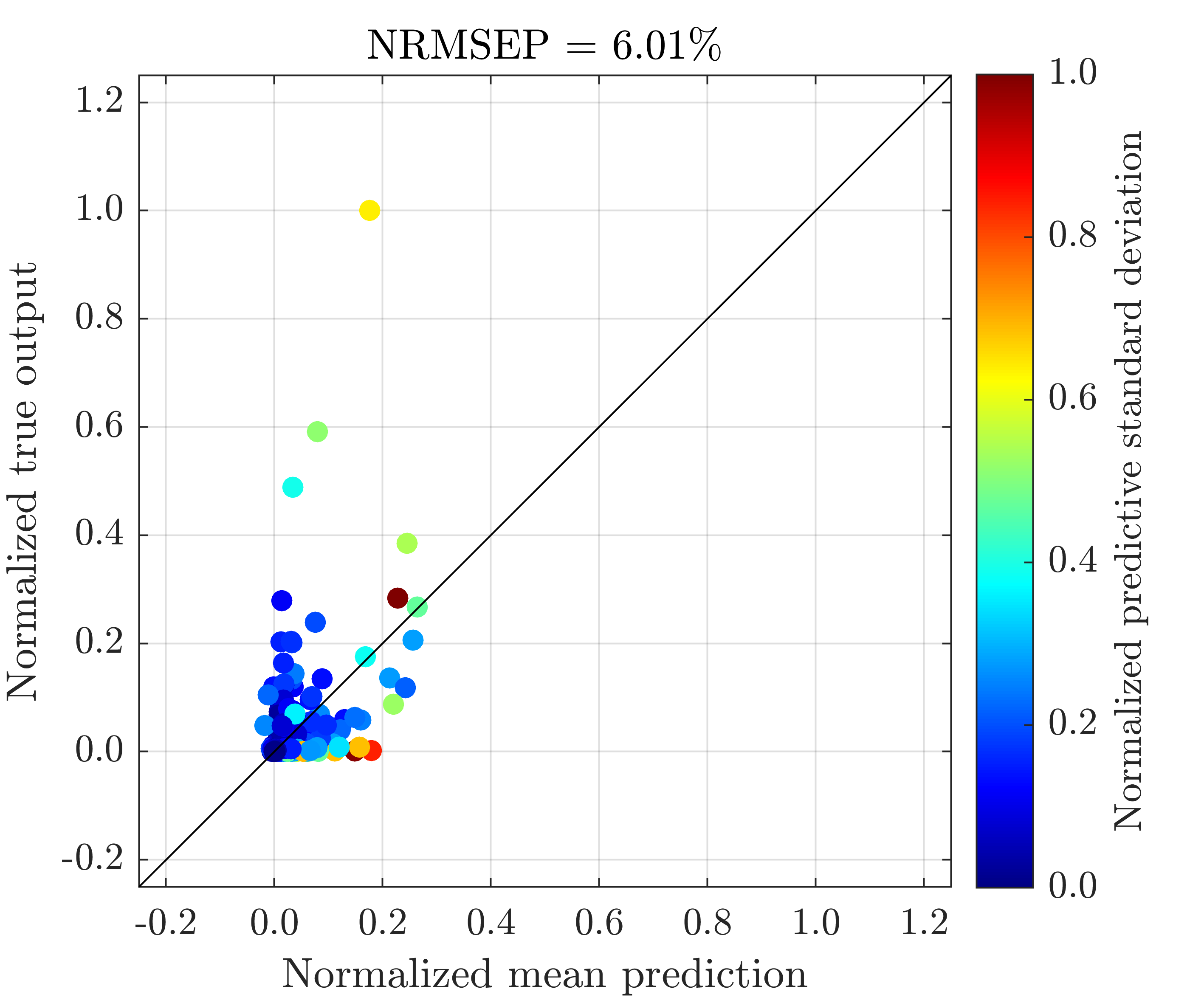}}
\subfloat[2-layer (DSVI)]{\label{fig:gamma_vi2}\includegraphics[width=0.25\linewidth]{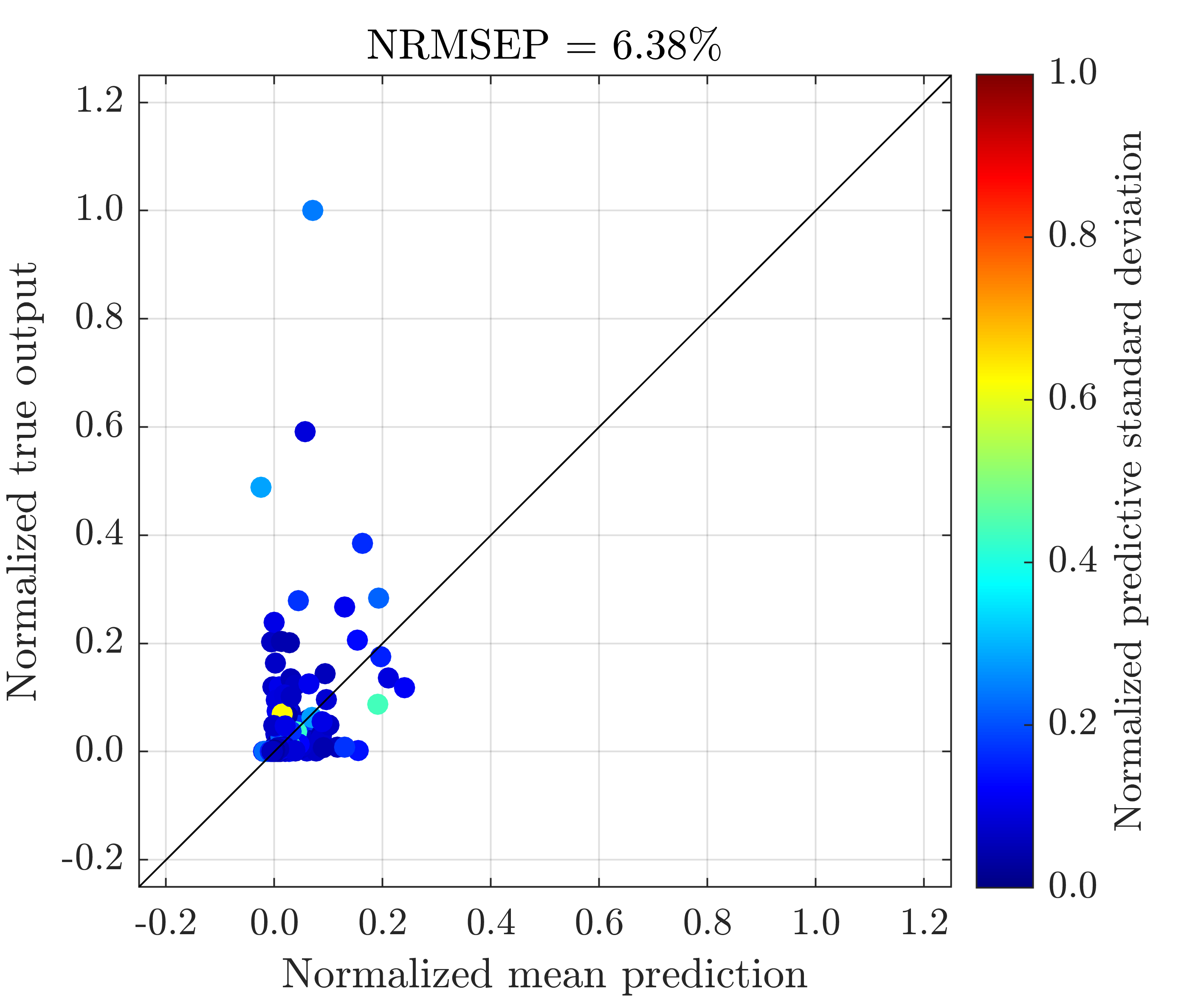}}
\subfloat[2-layer (SI)]{\label{fig:gamma_si2}\includegraphics[width=0.25\linewidth]{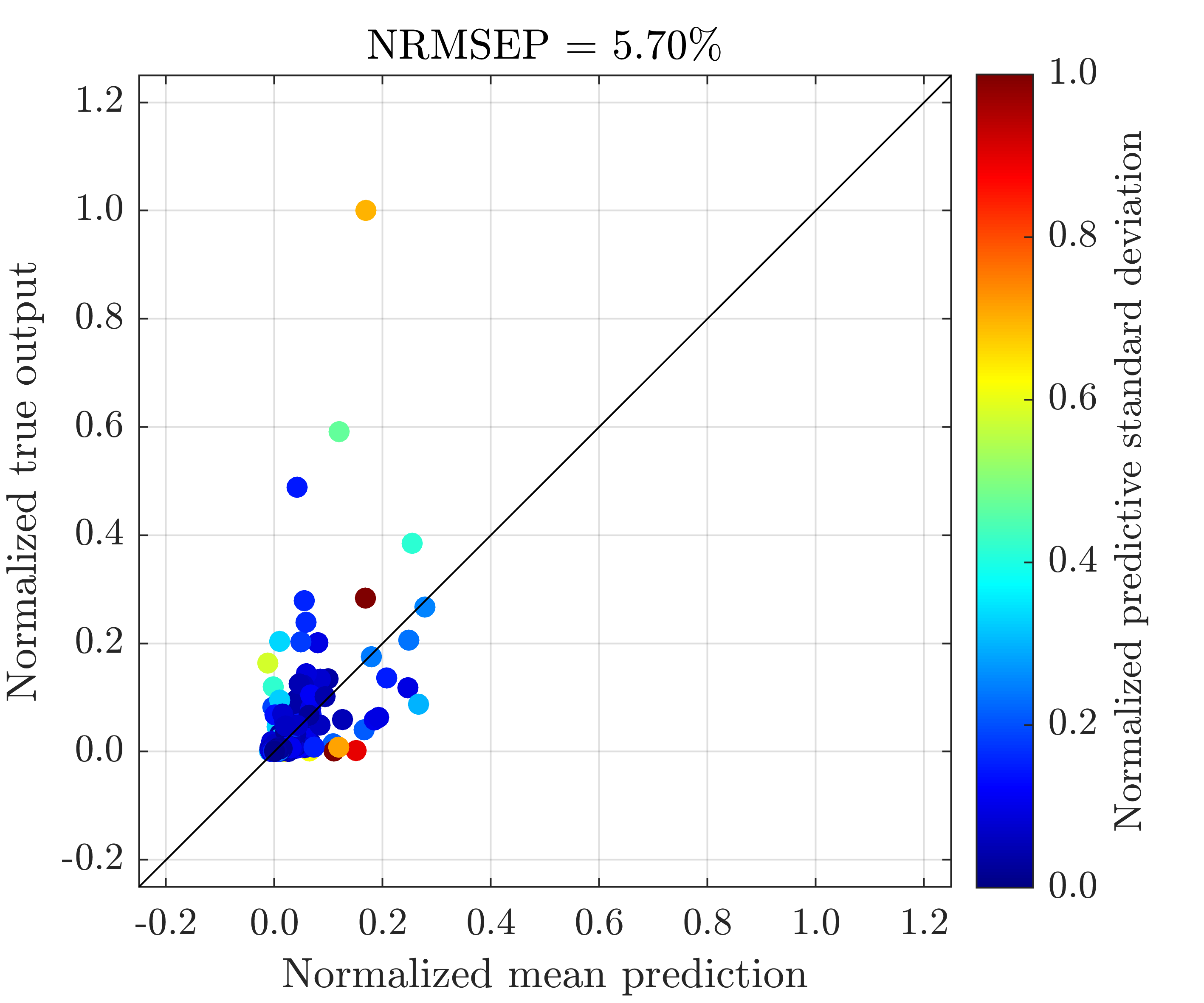}}
\subfloat[2-layer (SI-IC)]{\label{fig:gamma_si2_con}\includegraphics[width=0.25\linewidth]{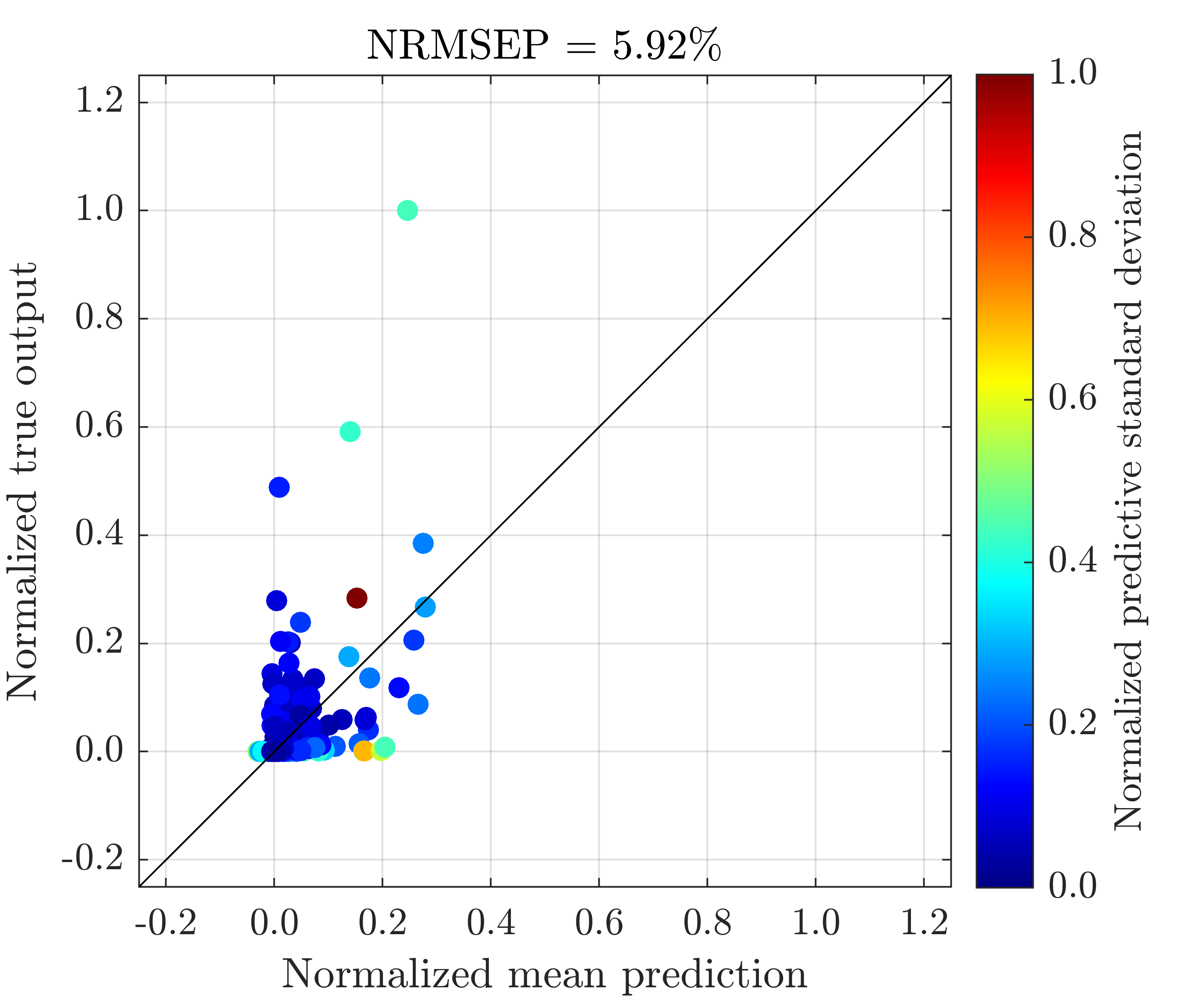}}\\
\subfloat[3-layer (FB)]{\label{fig:gamma_fb3}\includegraphics[width=0.25\linewidth]{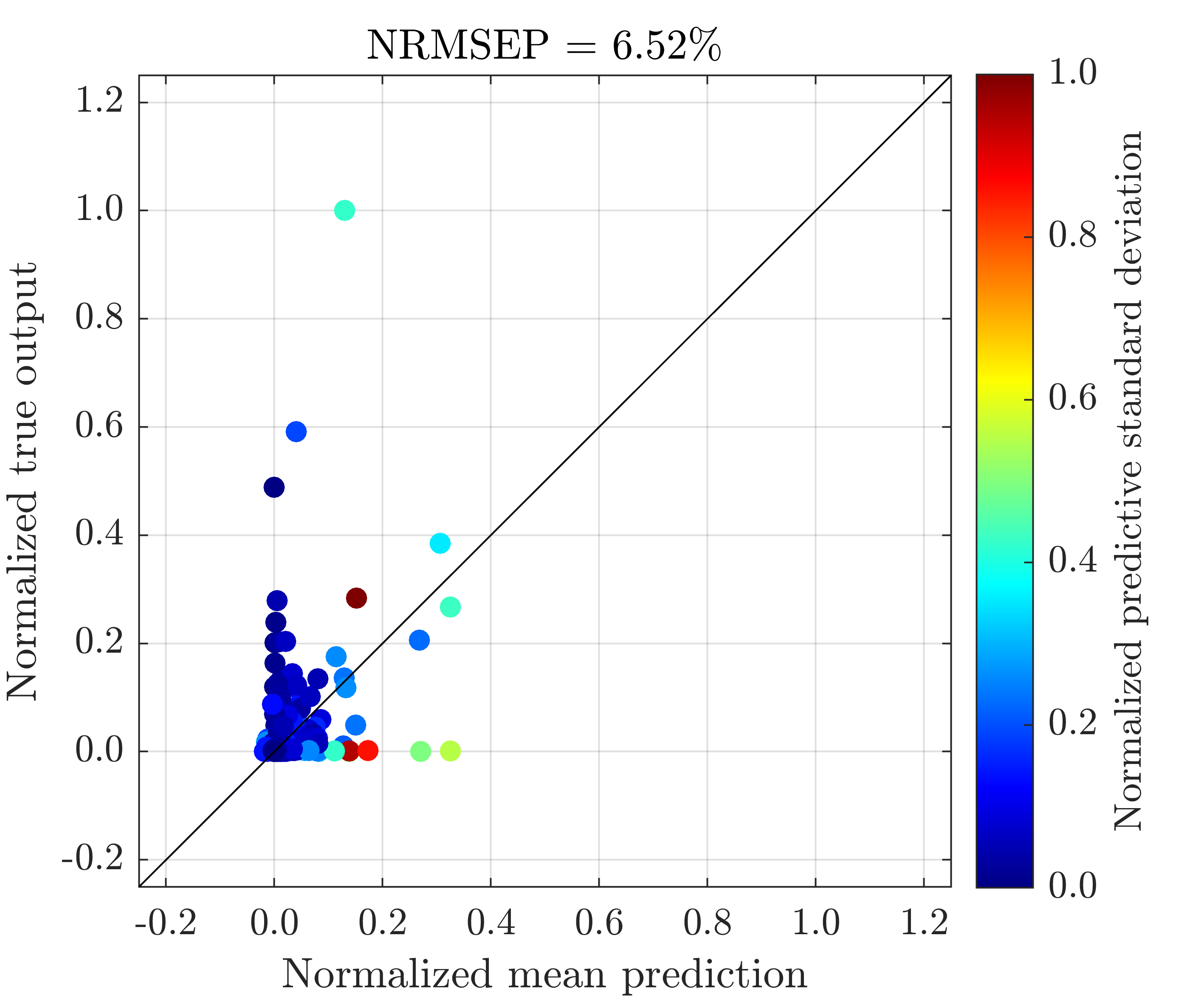}}
\subfloat[3-layer (DSVI)]{\label{fig:gamma_vi3}\includegraphics[width=0.25\linewidth]{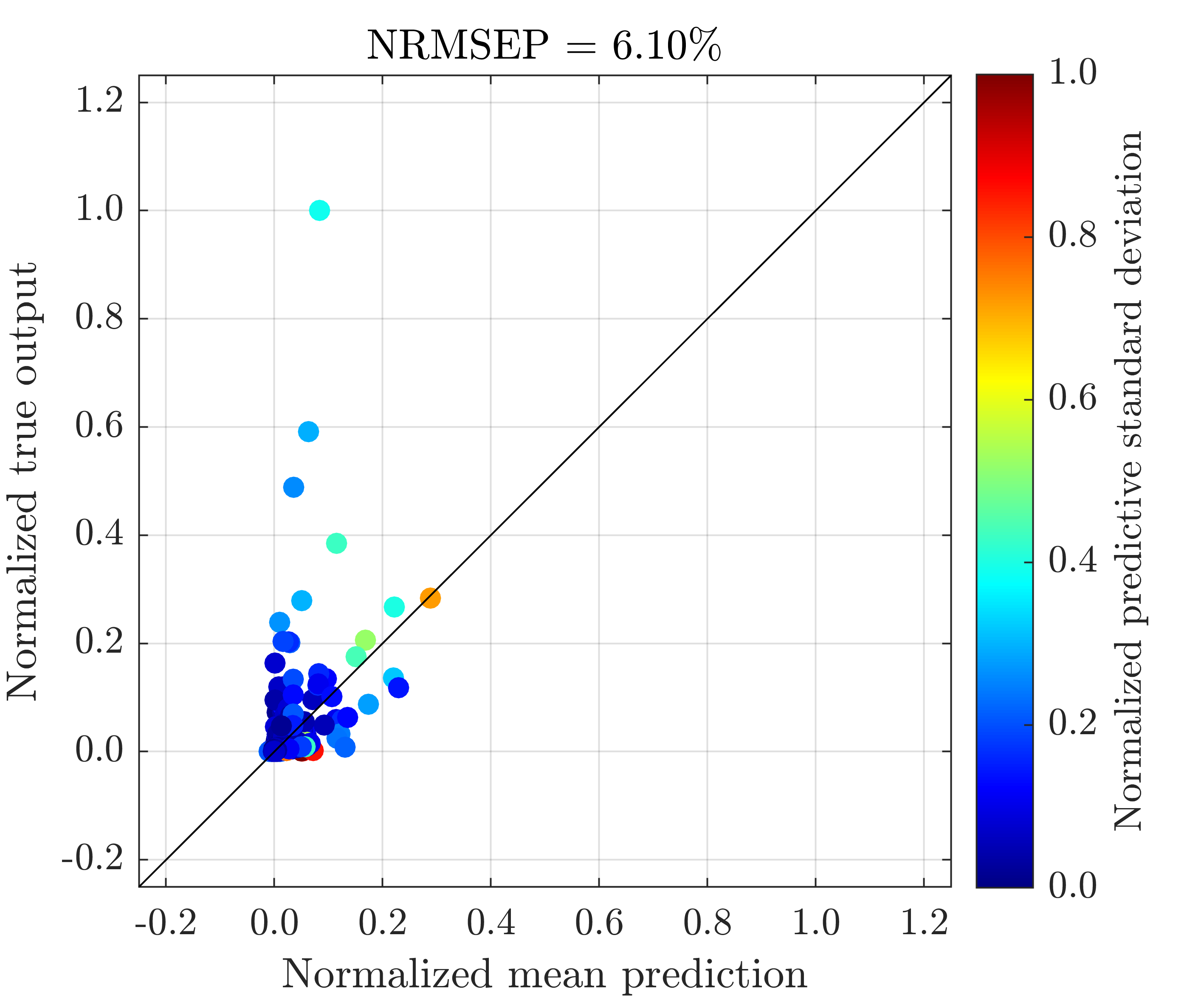}}
\subfloat[3-layer (SI)]{\label{fig:gamma_si3}\includegraphics[width=0.25\linewidth]{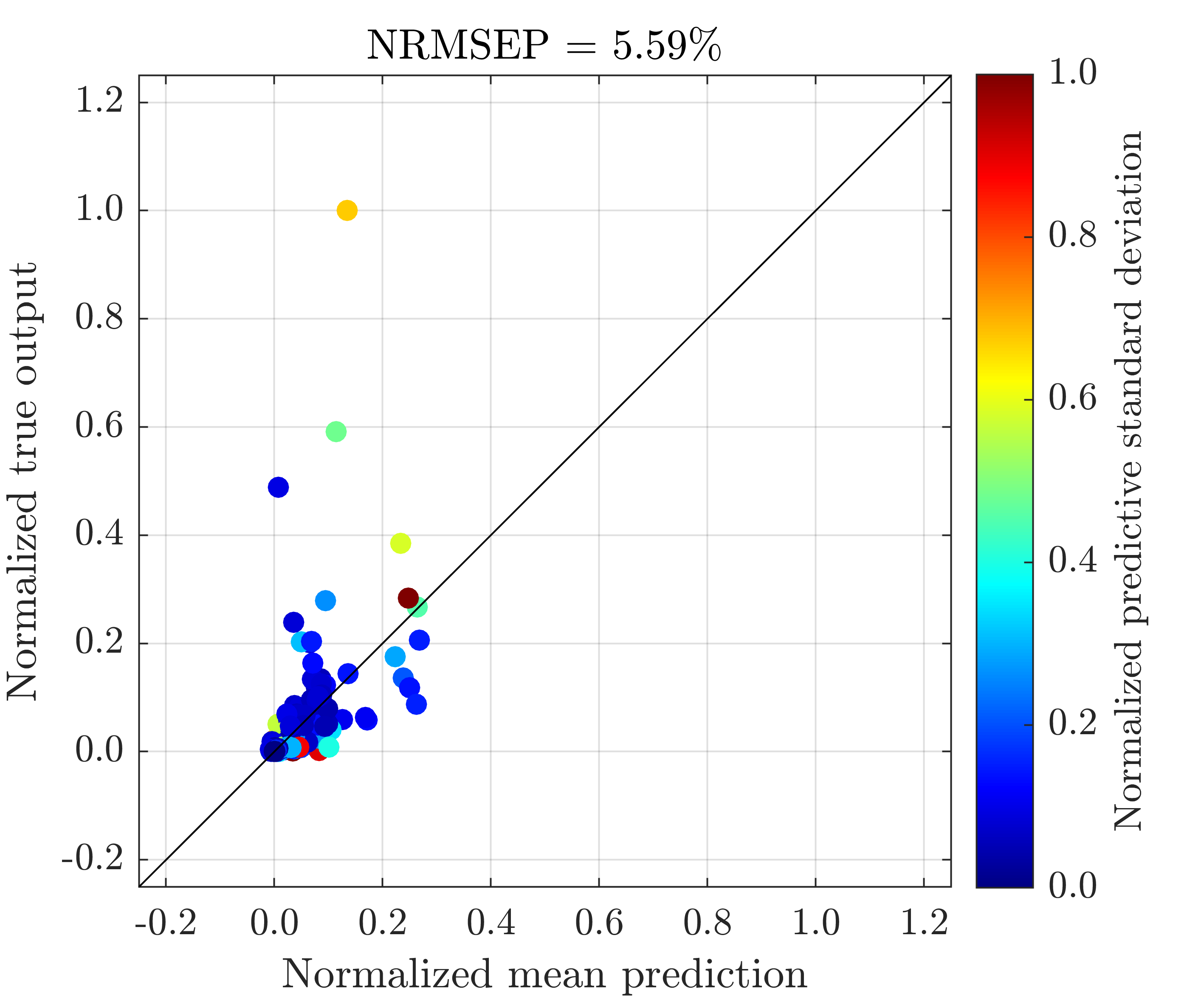}}
\subfloat[3-layer (SI-IC)]{\label{fig:gamma_si3_con}\includegraphics[width=0.25\linewidth]{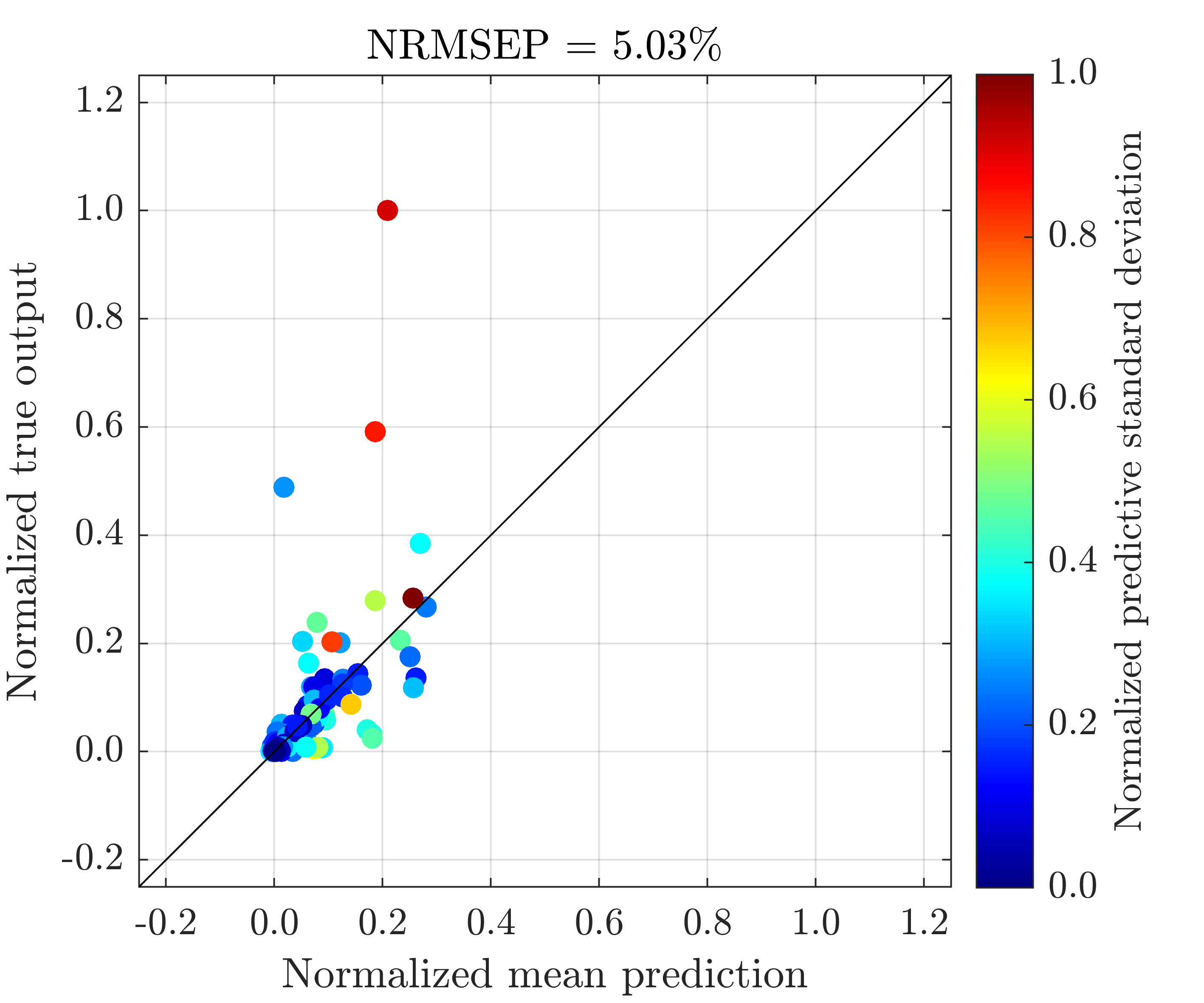}}\\
\subfloat[GP]{\label{fig:gamma_gp}\includegraphics[width=0.25\linewidth]{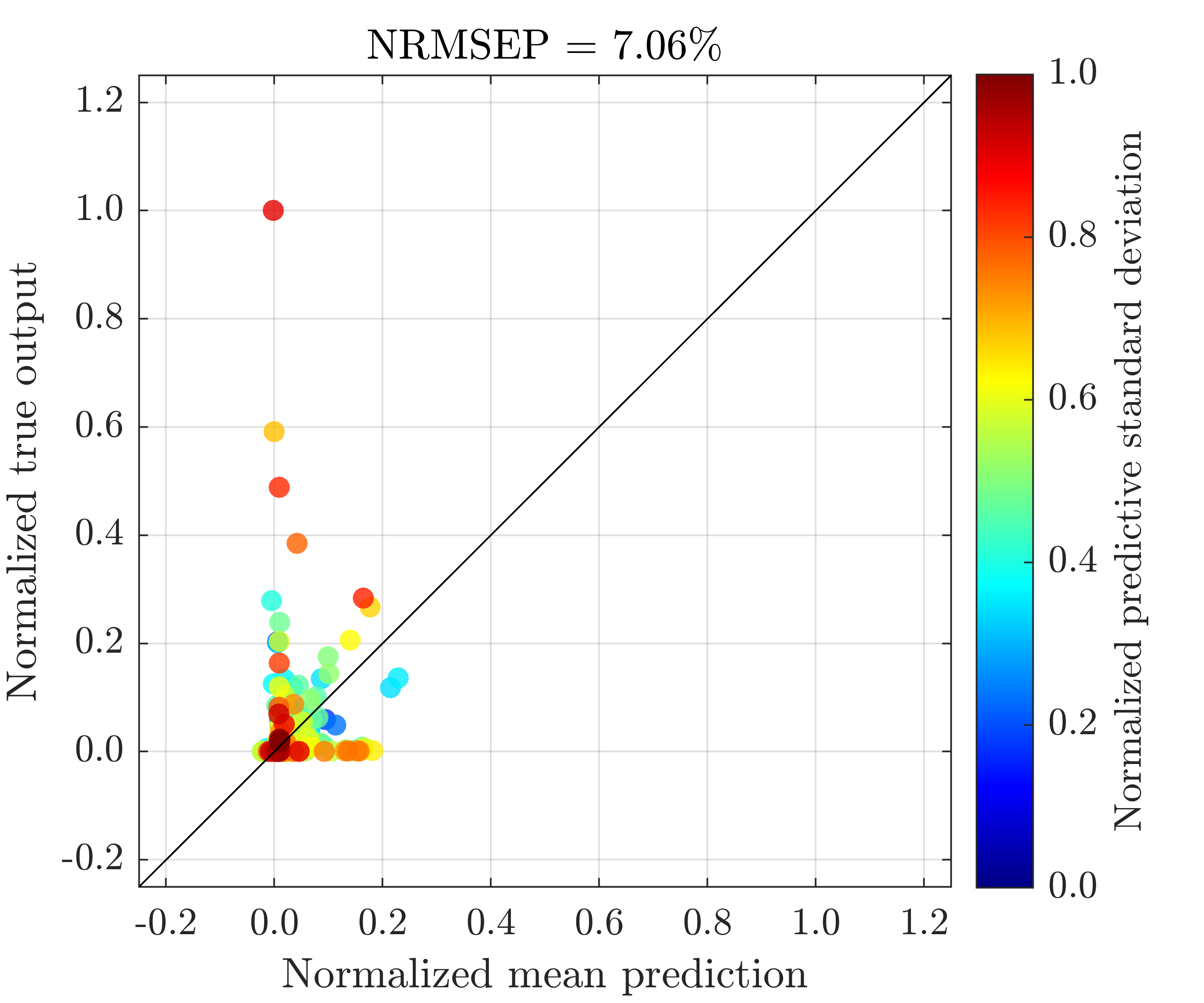}} 
\subfloat[4-layer (DSVI)]{\label{fig:gamma_v4}\includegraphics[width=0.25\linewidth]{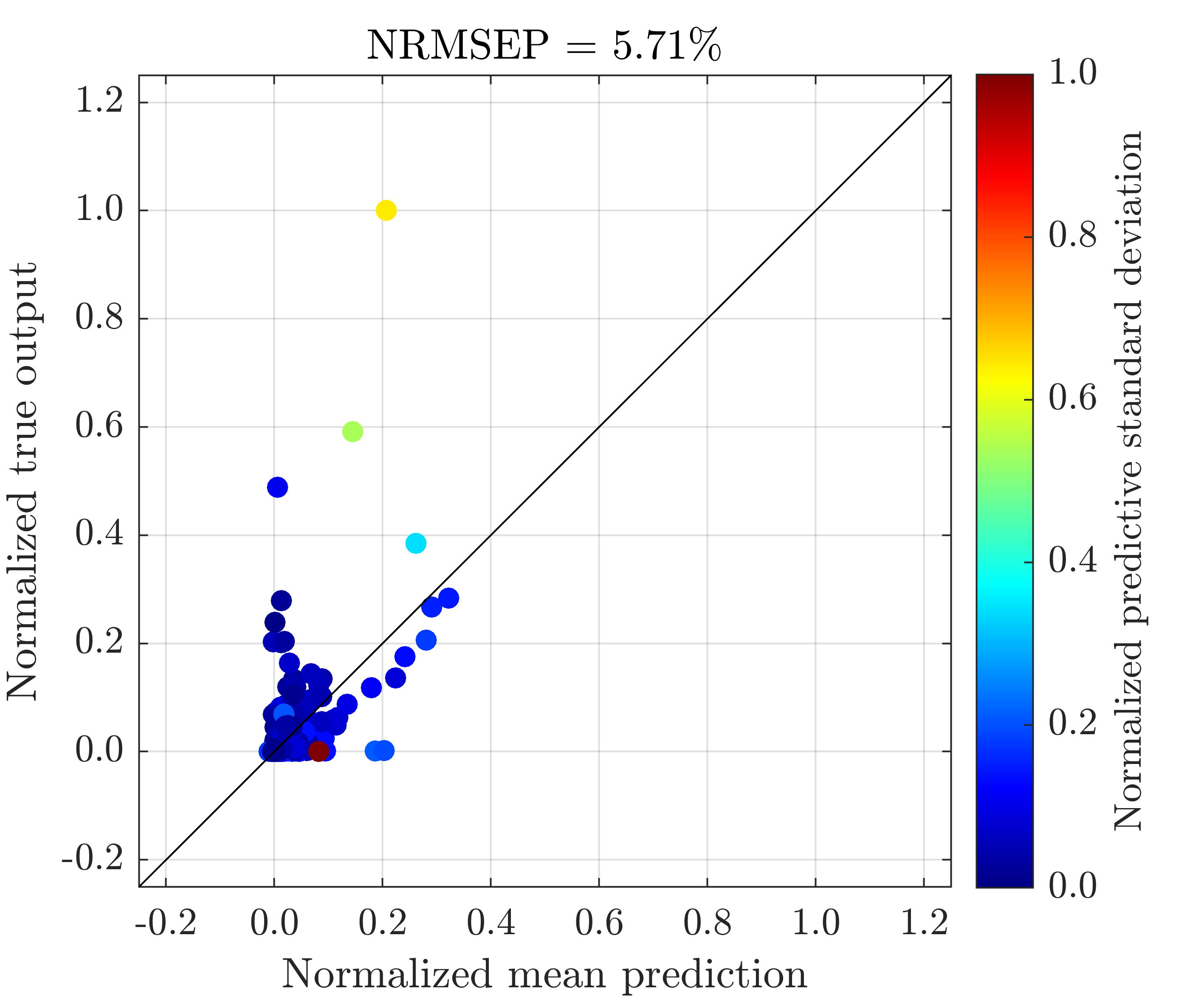}}
\subfloat[4-layer (SI)]{\label{fig:gamma_si4}\includegraphics[width=0.25\linewidth]{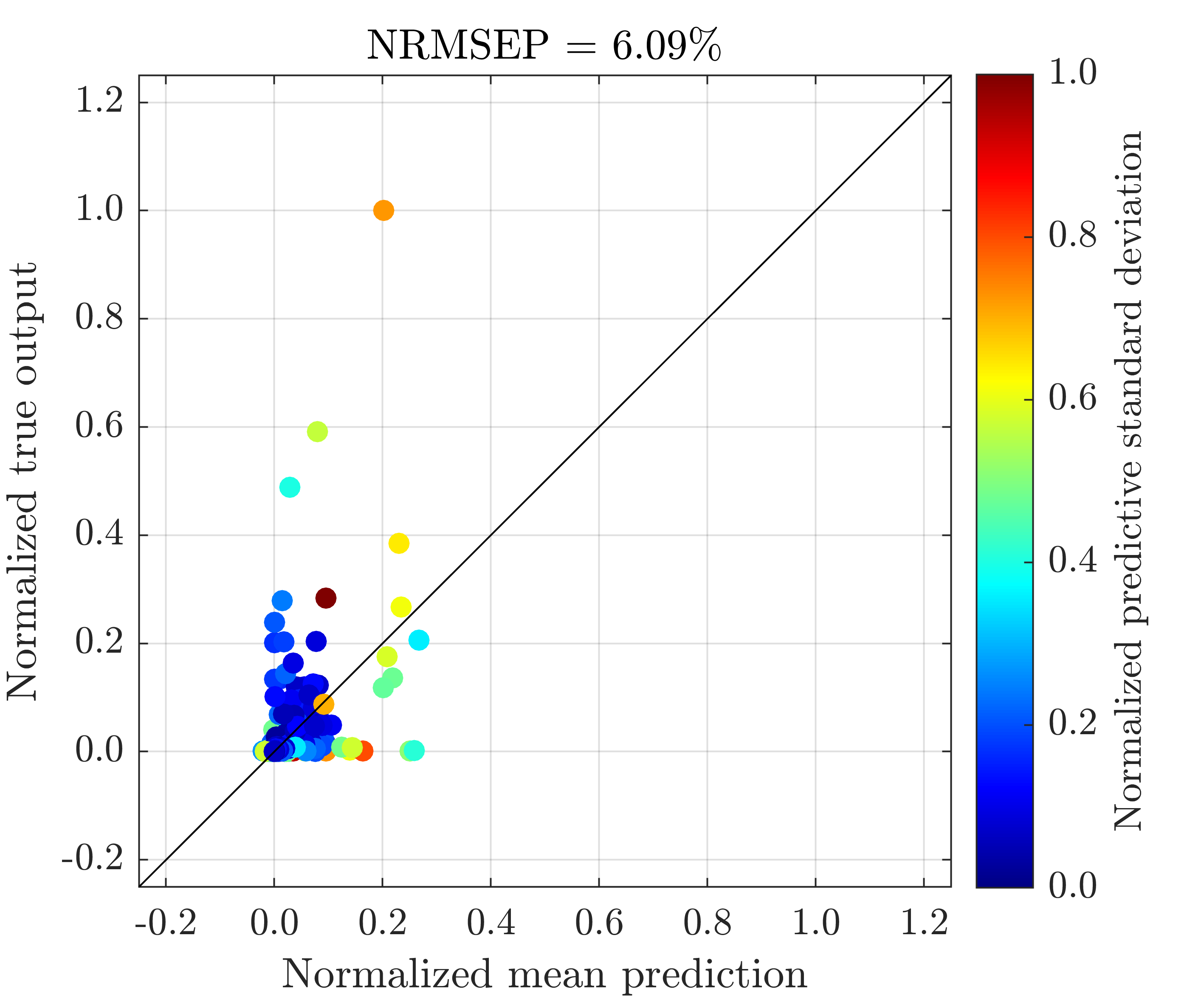}}
\subfloat[4-layer (SI-IC)]{\label{fig:gamma_si4_con}\includegraphics[width=0.25\linewidth]{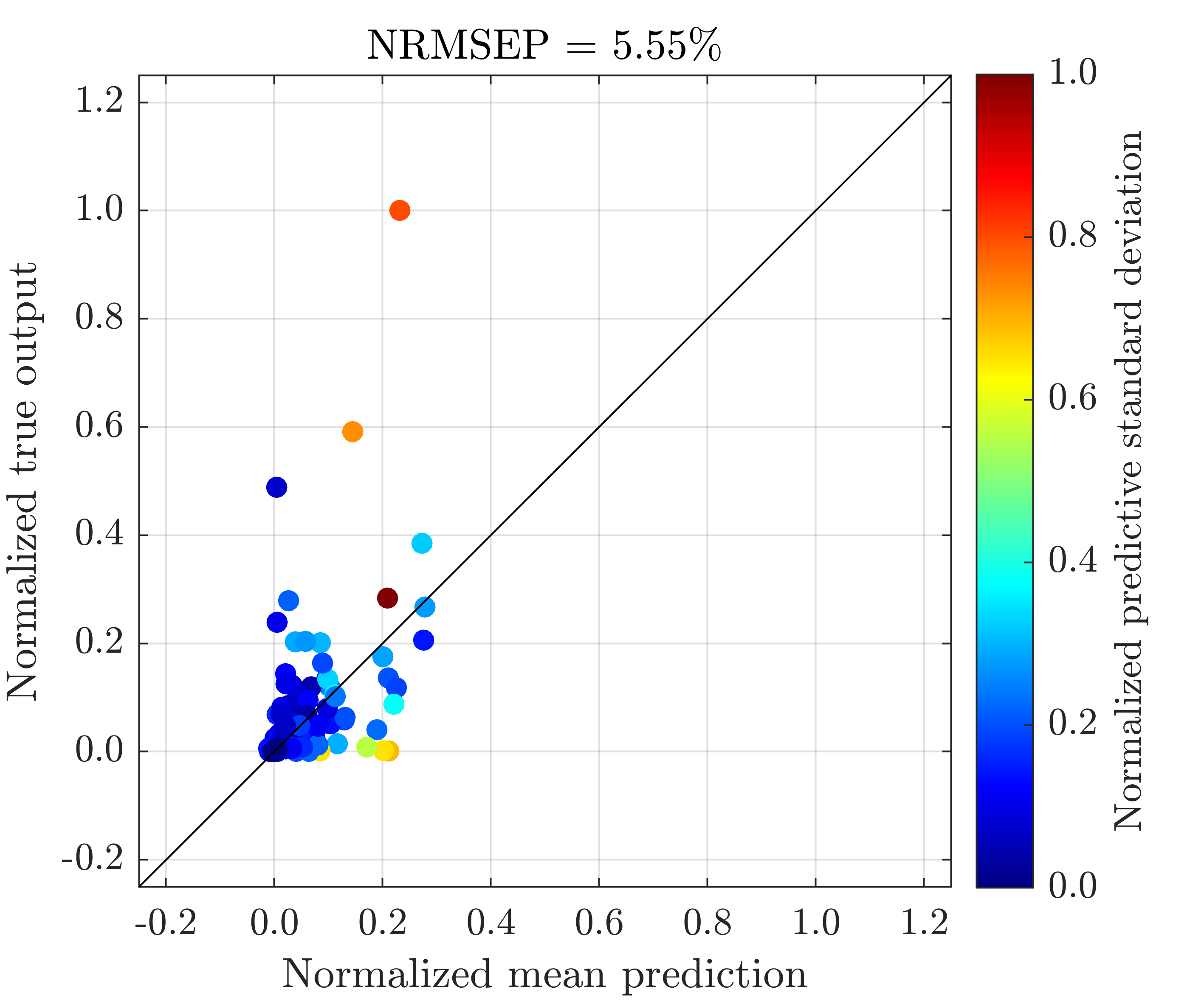}}
\caption{Plots of numerical solutions of Gamma ($\Gamma_t$) (normalized by their max and min values) from the Heston model at $500$ testing positions \emph{vs} the mean predictions (normalized by the max and min values of numerical solutions of Gamma), along with predictive standard deviations (normalized by their max and min values), made by the best emulator (with the lowest NRMSEP out of $40$ inference trials) produced by FB, DSVI, SI and SI-IC. \emph{GP} represents a conventional GP emulator.}
\label{fig:gamma_diag}
\end{figure}

Since a Latin-hypercube design (LHD) is insufficient to emulate adequately Gamma, it is natural to attempt sequential designs that would utilize uncertainty quantified by emulators (that are initially trained with a static LHD) to decide next design points, at which to compute corresponding Gamma numerically from the Heston model, and enrich the exiting training data set. However, as can be seen from Figure~\subref*{fig:gamma_gp}, conventional GP emulator assigns large predictive standard deviations to both locations away from spikes (i.e., points with corresponding true Gamma values close to zero) and near spikes (i.e., points with corresponding true Gamma values being away from zero). This can be problematic because the sequential design may select input locations outside the region of spikes. On the contrary, most of DGP emulators, particularly the three-layered one produced by SI-IC, capture the near-spike positions with notably higher and distinguishable predictive standard deviations. This behavior of uncertainty quantified by the DGP emulators is beneficiary since it would be easier and more efficient for sequential designs to pick design points that correspond to spikes embedded in Gamma. 

\end{document}